%% file: main.tex
\documentclass[10pt,twocolumn,letterpaper]{article}

\usepackage[pagenumbers]{cvpr} 

\input{preamble}

\definecolor{cvprblue}{rgb}{0.21,0.49,0.74}
\usepackage[pagebackref,breaklinks,colorlinks,citecolor=cvprblue]{hyperref}

\usepackage{graphicx}
\usepackage{animate}
\usepackage{subcaption}
\usepackage{makecell}
\usepackage{multirow}
\usepackage{color}
\usepackage{xcolor}
\usepackage{colortbl}
\usepackage[title]{appendix}
\usepackage[accsupp]{axessibility}

\def\paperID{618} 
\def\confName{CVPR}
\def\confYear{2024}

\title{AnimateZero: Video Diffusion Models are Zero-Shot Image Animators}

\author{Jiwen Yu$^1$\footnotemark[1]\qquad\quad Xiaodong Cun$^{2}$\footnotemark[2]\qquad\quad Chenyang Qi$^{3}$ \qquad\quad Yong Zhang$^2$ \\ \qquad\quad Xintao Wang$^2$\qquad\quad Ying Shan$^2$ \qquad\quad  Jian Zhang$^{1}$\footnotemark[2]\\
$^1$ Peking University \qquad $^2$ Tencent AI Lab \qquad $^3$ HKUST\\
\url{https://github.com/vvictoryuki/AnimateZero}
}

\begin{document}
\input{all_subtex/teaser}
\renewcommand{\thefootnote}{\fnsymbol{footnote}}
\footnotetext[1]{~Work done during an internship at Tencent AI Lab.}
\footnotetext[2]{~Corresponding Authors.}

\input{tex/0_abs}
\input{tex/1_intro}
\input{tex/2_related}

\input{tex/3_method}

\input{tex/4_exp}
\input{tex/5_conclusion}

{
    \small
    \bibliographystyle{ieeenat_fullname}
    \bibliography{main}
}

\appendix
\onecolumn{
\begin{appendices}
    This appendix includes our supplementary materials as follows:
\begin{itemize}
    \item Section~\ref{sec:append-implemtation}: Implementation details of utilized T2I checkpoints and the proposed position-enhanced window attention.
    \item Section~\ref{sec:append_tt}: Introduce the effect of time-travel sampling strategy. 
    \item Section~\ref{sec:append_applications}: Introduce the extensive applications of AnimateZero.
    \item Section~\ref{sec:append_moreresults}: Provide more visual results. Dynamic videos can be found in our project page: \url{https://vvictoryuki.github.io/animatezero.github.io/}
\end{itemize}

\input{appendix_tex/sec1}

\input{appendix_tex/sec2}

\input{appendix_tex/sec3}
\input{appendix_tex/sec4}

\end{appendices}
}

\end{document}

%% file: preamble.tex
\usepackage[dvipsnames]{xcolor}

\definecolor{ao}{rgb}{0.0, 0.5, 0.0}

\newcommand{\animation}{1}

\definecolor{color3}{rgb}{0.95,0.95,0.95}

%% file: all_subtex/teaser.tex
\if \animation 1
\twocolumn[{
\maketitle
\begin{center}
    \captionsetup{type=figure}
    \vspace{-2em}
    \begin{tabular}{c@{\hspace{0.3em}}c@{\hspace{0.8em}}c@{\hspace{0.3em}}c@{\hspace{0.8em}}c@{\hspace{0.3em}}c}
  Generated Image & Output Video & Generated Image & Output Video & Generated Image & Output Video \vspace{0.3em} \\
  \includegraphics[width=0.15\linewidth]{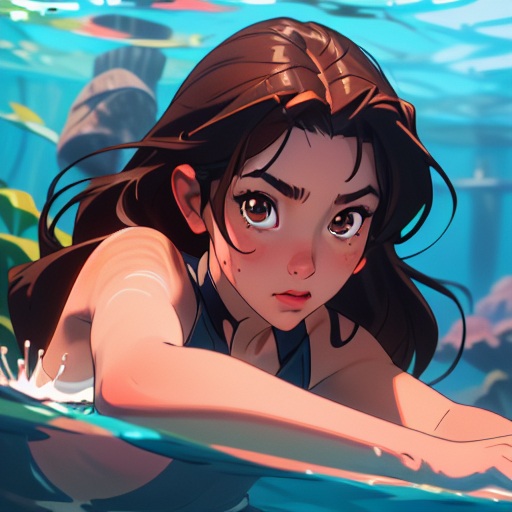}&
  \animategraphics[width=0.15\linewidth]{8}{gif/teaser/1/082_frame_}{0}{15} &
  \includegraphics[width=0.15\linewidth]{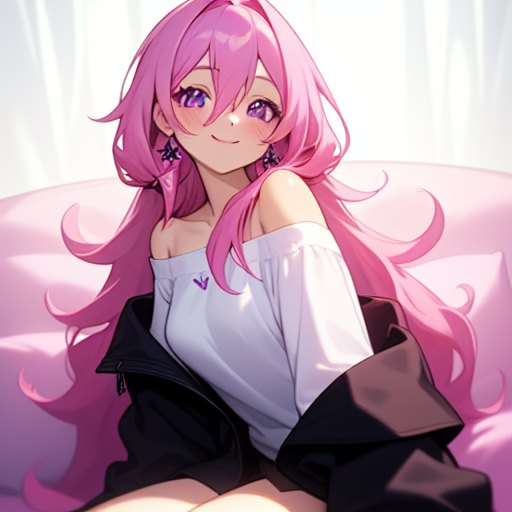}&
  \animategraphics[width=0.15\linewidth]{8}{gif/teaser/3/012_frame_}{0}{15}&
  \includegraphics[width=0.15\linewidth]{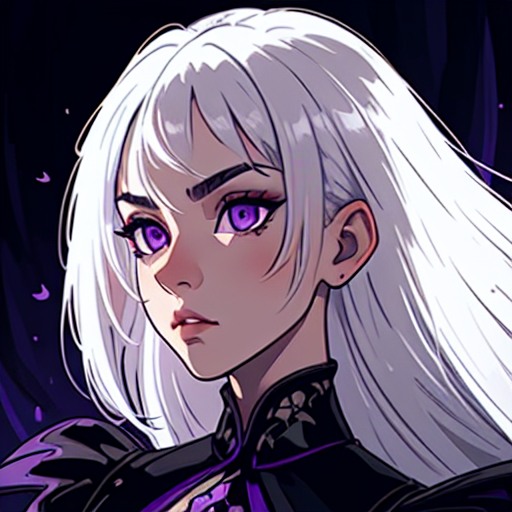}&
  \animategraphics[width=0.15\linewidth]{8}{gif/teaser/4/020_frame_}{0}{15}\\
  \multicolumn{2}{c}{\textit{``1girl, underwater, swimsuit, ...''}} & \multicolumn{2}{c}{\textit{``1girl, black jacket, long sleeves,  ...''}} & \multicolumn{2}{c}{\textit{``dark fantasy, purple eyes ...''}}\vspace{0.5em}\\
  \includegraphics[width=0.15\linewidth]{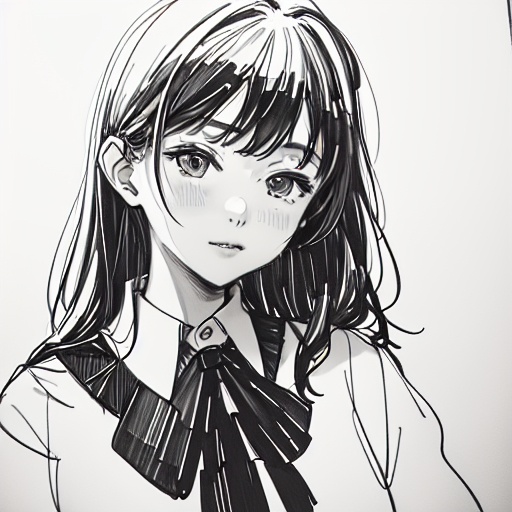}&
  \animategraphics[width=0.15\linewidth]{8}{gif/teaser/5/034_frame_}{0}{15}&
  \includegraphics[width=0.15\linewidth]{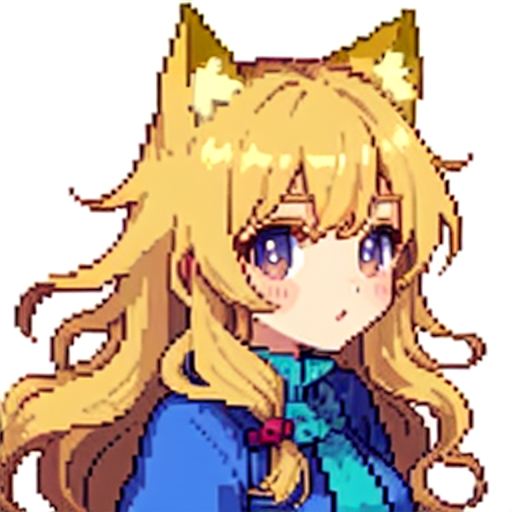}&
  \animategraphics[width=0.15\linewidth]{8}{gif/teaser/6/051_frame_}{0}{15}&
  \includegraphics[width=0.15\linewidth]{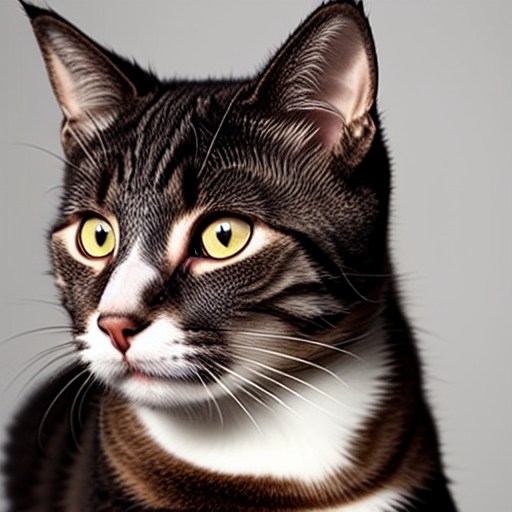}&
  \animategraphics[width=0.15\linewidth]{8}{gif/teaser/10/141_frame_}{0}{15}\\
  \multicolumn{2}{c}{\textit{``school uniform, JK, sketch ...''}} & \multicolumn{2}{c}{\textit{``pixel art, cat ears, blonde hair  ...''}}\vspace{0.5em} & \multicolumn{2}{c}{\textit{``a cat head, look to one side''}} \\
  \includegraphics[width=0.15\linewidth]{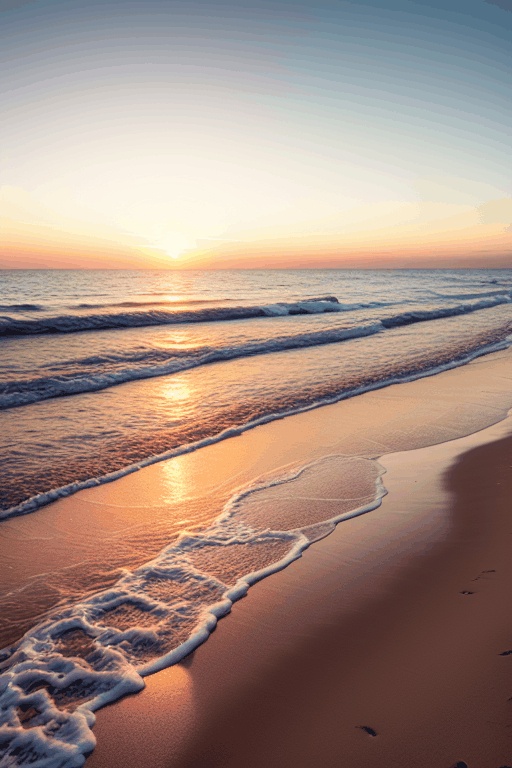}&
  \animategraphics[width=0.15\linewidth]{8}{gif/teaser/7/100-}{0}{15}&
  \includegraphics[width=0.15\linewidth]{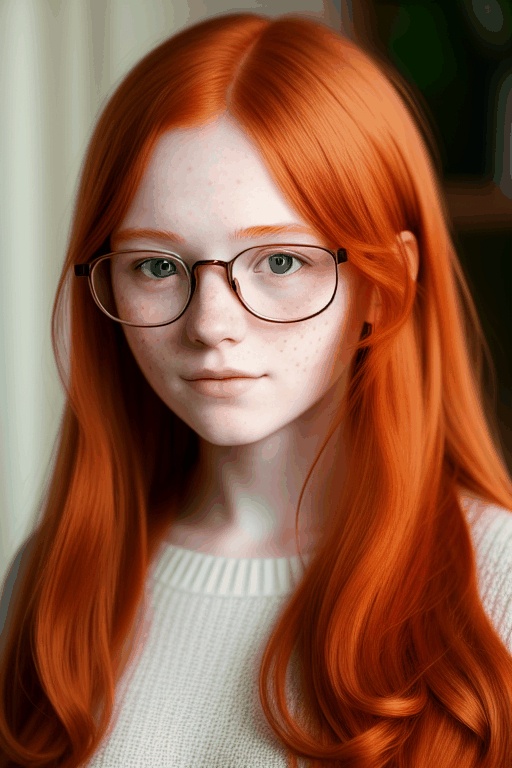}&
  \animategraphics[width=0.15\linewidth]{8}{gif/teaser/8/580-}{0}{15}&
  \includegraphics[width=0.15\linewidth]{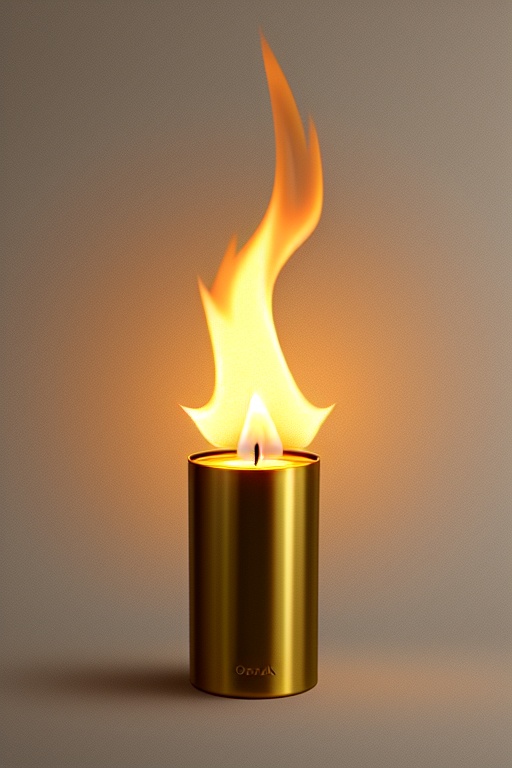}&
  \animategraphics[width=0.15\linewidth]{8}{gif/teaser/12/153_frame_}{0}{15}\\
  \multicolumn{2}{c}{\textit{``waves hit the beach''}} & \multicolumn{2}{c}{\textit{``freckles, orange hair, glasses, ...''}} & \multicolumn{2}{c}{\textit{``lighter, flame, candle''}}

    \end{tabular}
    
    \vspace{-0.5em}
    \captionof{figure}{ 
    Our proposed AnimateZero modifies the architecture of the text-to-video diffusion model, AnimateDiff~\cite{guo2023animatediff}, to achieve more controllable~(\eg, control the appearance using images generated by pre-trained text-to-image models) video generation without further training. The results above demonstrate the effectiveness of AnimateZero in generating animated videos from the exactly same domains of the generated images. These personalized image domains include anime style, sketch style, pixel-art style, and realistic style. \textit{Best viewed with Acrobat Reader. Click the video to play the animation clips. \textbf{Static frames are provided in supplementary materials.}} 
    }
    \label{fig:teaser}
\end{center}
}]
\else
\twocolumn[{
\maketitle
\begin{center}
    \captionsetup{type=figure}
    \vspace{-2em}
    \begin{tabular}{c@{\hspace{0.3em}}c@{\hspace{0.8em}}c@{\hspace{0.3em}}c@{\hspace{0.8em}}c@{\hspace{0.3em}}c}
  Generated Image & Output Video & Generated Image & Output Video & Generated Image & Output Video \vspace{0.3em} \\
  \includegraphics[width=0.15\linewidth]{gif/teaser/1/082_frame_0.jpg}&
  \includegraphics[width=0.15\linewidth]{gif/teaser/1/082_frame_0.jpg}&
  \includegraphics[width=0.15\linewidth]{gif/teaser/3/012_frame_0.jpg}&
  \includegraphics[width=0.15\linewidth]{gif/teaser/3/012_frame_0.jpg}&
  \includegraphics[width=0.15\linewidth]{gif/teaser/4/020_frame_0.jpg}&
  \includegraphics[width=0.15\linewidth]{gif/teaser/4/020_frame_0.jpg}\\
  \multicolumn{2}{c}{\textit{``1girl, underwater, swimsuit, ...''}} & \multicolumn{2}{c}{\textit{``1girl, black jacket, long sleeves,  ...''}} & \multicolumn{2}{c}{\textit{``dark fantasy, purple eyes ...''}}\vspace{0.5em}\\
  \includegraphics[width=0.15\linewidth]{gif/teaser/5/034_frame_0.jpg}&
  \includegraphics[width=0.15\linewidth]{gif/teaser/5/034_frame_0.jpg}&
  \includegraphics[width=0.15\linewidth]{gif/teaser/6/051_frame_0.jpg}&
  \includegraphics[width=0.15\linewidth]{gif/teaser/6/051_frame_0.jpg}&
  \includegraphics[width=0.15\linewidth]{gif/teaser/10/141_frame_0.jpg}&
  \includegraphics[width=0.15\linewidth]{gif/teaser/10/141_frame_0.jpg}\\
   \multicolumn{2}{c}{\textit{``school uniform, JK, sketch ...''}} & \multicolumn{2}{c}{\textit{``pixel art, cat ears, blonde hair  ...''}}\vspace{0.5em} & \multicolumn{2}{c}{\textit{``a cat head, look to one side''}} \\
  \includegraphics[width=0.15\linewidth]{gif/teaser/7/100-0.jpg}&
  \includegraphics[width=0.15\linewidth]{gif/teaser/7/100-0.jpg}&
  \includegraphics[width=0.15\linewidth]{gif/teaser/8/580-0.jpg}&
  \includegraphics[width=0.15\linewidth]{gif/teaser/8/580-0.jpg}&
  \includegraphics[width=0.15\linewidth]{gif/teaser/12/153_frame_0.jpg}&
  \includegraphics[width=0.15\linewidth]{gif/teaser/12/153_frame_0.jpg}\\
  \multicolumn{2}{c}{\textit{``waves hit the beach''}} & \multicolumn{2}{c}{\textit{``freckles, orange hair, glasses, ...''}} & \multicolumn{2}{c}{\textit{``lighter, flame, candle''}}

    \end{tabular}
    
    \vspace{-0.5em}
    \captionof{figure}{ 
    Our proposed AnimateZero modifies the architecture of the text-to-video diffusion model, AnimateDiff~\cite{guo2023animatediff}, to achieve more controllable~(\eg, control the appearance using images generated by pre-trained text-to-image models) video generation without further training. The results above demonstrate the effectiveness of AnimateZero in generating animated videos from the exactly same domains of the generated images. These personalized image domains include anime style, sketch style, pixel-art style, and realistic style. \textit{Best viewed with Acrobat Reader. Click the video to play the animation clips. \textbf{Static frames are provided in supplementary materials.}} 
    }
    \label{fig:teaser}
\end{center}
}]

\fi

%% file: tex/0_abs.tex
\begin{abstract}
Large-scale text-to-video (T2V) diffusion models have great progress in recent years in terms of visual quality, motion and temporal consistency. However, the generation process is still a black box, where all attributes~(\eg, appearance, motion) are learned and generated jointly without precise control ability other than rough text descriptions.
Inspired by image animation which decouples the video as one specific appearance with the corresponding motion, we propose AnimateZero to
unveil the pre-trained text-to-video diffusion model, \ie, AnimateDiff, and provide more precise appearance and motion control abilities for it.
For appearance control, we borrow intermediate latents and their features from the text-to-image (T2I) generation for ensuring the generated first frame is equal to the given generated image. 
For temporal control, we replace the global temporal attention of the original T2V model with our proposed positional-corrected window attention to ensure other frames align with the first frame well. 
Empowered by the proposed methods, 
AnimateZero can successfully control the generating progress without further training.
As a zero-shot image animator for given images, AnimateZero also enables multiple new applications, including interactive video generation and real image animation.
The detailed experiments demonstrate the effectiveness of the proposed method in both T2V and related applications.
\end{abstract}

%% file: tex/1_intro.tex
\section{Introduction}

Empowered by the recent development of generative priors in large-scale text-to-image (T2I) diffusion models, the video diffusion models (VDMs), especially text-to-video (T2V) diffusion models, have experienced rapid developments in terms of the resolutions~\cite{chen2023videocrafter1,he2023scalecrafter}, network structures~\cite{guo2023animatediff, he2022lvdm,chen2023seine}, and commercial applications~\cite{pikalabs,genmo}, \etc.

Although VDMs are easy to use, the whole generation process is still a black box without precise control capabilities, where the users need to wait for a relatively long time to know the generated results if they have limited GPUs. 
Moreover, because most VDMs are trained jointly in terms of appearance and temporal aspects, it is not easy to control these two parts separately.
These problems can be natively handled by generating videos by a chain of T2I and I2V~(Image-to-Video).
However, these two different networks, T2I and I2V model, might not be in the same domain, \eg, the T2I produces a comic image, whereas the I2V diffusion models are only trained on real-world clips. Thus, the generated results might exhibit domain bias.
To this end, we are curious about the detailed generation process in the T2V generation so that we can decouple and control appearance and motion respectively and generate better videos step by step.

To achieve this goal, we are inspired by the image animation methods to consider the video as a single keyframe appearance and its corresponding movement. 
The keyframe can be described by the text prompt, which is a constant in the generation, and other frames utilize the knowledge of this frame for animation through the temporal modules. 

Based on the above observations, we propose AnimateZero, a zero-shot method modifying the architecture of pre-trained VDMs to unveil the generation process of the pre-trained VDMs so that the appearance and motion control can be easily separated. Specifically, we have designed spatial appearance control and temporal consistency control for these two parts. Spatial appearance control involves modifying the spatial modules to insert the generated images into the first frame of generated videos. Temporal consistency control involves modifying the motion modules to make other frames aligned with the first frame.  Finally, we have achieved step-by-step video generation from T2I to I2V in a \textbf{zero-shot} manner. 
It is worth emphasizing that leveraging the well-established Stable Diffusion~\cite{rombach2022high} community, our approach supports various personalized image domains, including but not limited to realistic style, anime style, pixel art style, and more.

Our contributions can be summarized as follows: 
\begin{itemize}
    \item We propose a novel controllable video generation method called AnimateZero, which decouples the generation progress of pre-trained VDMs, thus achieving step-by-step video generation from T2I to I2V.

    \item We propose spatial appearance control and temporal consistency control for AnimateZero to animate generated images in a zero-shot way. Our approach is the first to prove that the pre-trained VDMs have the potential to be zero-shot image animators.

    \item Experimental results highlight AnimateZero's effectiveness in various personalized data domains. In video generation, AnimateZero surpasses AnimateDiff in similarity to the text and the T2I domain. It excels in multiple metrics compared to current I2V methods and is on par with the best method in other metrics.

\end{itemize}

%% file: tex/2_related.tex
\section{Related Work}
\subsection{Text-to-Video Diffusion Models}
Video Diffusion Models (VDMs)~\cite{ho2022video,luo2023videofusion}, especially Text-to-Video Diffusion Models (T2Vs)~\cite{he2022lvdm,singer2023makeavideo,ho2022imagen,zhou2022magicvideo,He_2023_CVPR,wang2023modelscope,guo2023animatediff,wang2023lavie,zhang2023show,chen2023videocrafter1}, have experienced rapid development recent years, making significant progress in the quality, diversity, and resolution of generated videos. Many works within these VDMs are based on tuning text-to-image diffusion models (T2Is)~\cite{rombach2022high} with the addition of temporal modules. These approaches reduce the training costs of VDMs and leverage prior knowledge from the image domain. 
However, the tuning efforts in these works do not decouple the T2Is from the added temporal modules.  Instead, they train them together, making it difficult to separate the appearance and motion control. Additionally, these methods inevitably disrupt the original T2I domain, resulting in a domain gap.

Recently, a category of VDMs that decouples T2Is and the temporal modules has emerged~\cite{guo2023animatediff, hotshotxl}.  
While they provide the potential to control appearance and motion separately, they still face the challenge of disrupting the original T2I domain (demonstrated in Fig.~\ref{fig:ADvsAZ}).
Our proposed AnimateZero is based on AnimateDiff~\cite{guo2023animatediff}.

\subsection{Zero-shot Modification for Diffusion Models}
Diffusion models~\cite{ho2020denoising, song2019generative, song2021scorebased}, as representatives of large-scale vision models, have attracted considerable research attention on how to utilize them in zero-shot or training-free manners for various downstream tasks~\cite{qi2023fatezero, wang2023zeroshot, wang2023unlimited, yu2023freedom, yu2023cross}. Among these efforts, many works attempt to directly modify the model architecture to achieve new capabilities, for instance: Prompt-to-Prompt~\cite{hertz2022prompt} modifies the cross-attention of Stable Diffusion~\cite{rombach2022high} for convenient image editing; ScaleCrafter~\cite{he2023scalecrafter} modifies the convolutional kernels in the UNet of diffusion models to achieve high-quality generation at higher resolutions; MasaCtrl~\cite{cao_2023_masactrl} achieves personalized image generation by sharing keys and values of the source images from the self-attention in Stable Diffusion. 

Our proposed AnimateZero is also a method modifying the architecture of diffusion models, achieving zero-shot step-by-step video generation from generated images.

\subsection{Image-to-Video Diffusion Models}
In the realm of downstream tasks utilizing VDMs for video-related applications, there exists a category of work known as Image-to-Video Diffusion Models (I2Vs)~\cite{chen2023videocrafter1,xing2023dynamicrafter,i2vgenxl}. The goals of these models are similar to Image Animation, but they differ in some aspects. The primary difference is that most of these methods employ an image encoder to extract semantic features from a reference image to guide video generation, without requiring the generated video to precisely include the given image as the first frame.

Recently, there have been some attempts to move towards Image Animation: publicly available tools include Gen-2~\cite{gen2}, Genmo~\cite{genmo}, and Pika Labs~\cite{pikalabs}. Among them, Gen-2, as a commercial large-scale model, delivers impressive results in the realistic image domain in its November 2023 update. However, its performance in other domains, which might not have been covered in training, is still not entirely satisfactory. Genmo and Pika Labs also face the same challenge.
Related research papers include SEINE~\cite{chen2023seine} and LAMP~\cite{wu2023lamp}, which are currently under submission. However, their I2V models require training and are still dependent on specific training data domains.

In comparison, our approach holds unique advantages due to its characteristic of being training-free and supporting various personalized image domains.


%% file: tex/3_method.tex
\section{Preliminaries: AnimateDiff~\cite{guo2023animatediff}}
To simplify the experiments and hypotheses, we choose one specific video diffusion model, \ie, AnimateDiff~\cite{guo2023animatediff}, as the base video model, since it only trains additional temporal layers based on a fixed text-to-image diffusion model for text-to-video generation, as shown in Fig.~\ref{fig:ad_achitecture}.
Below, we give the details of the whole network structure of AnimateDiff and its motion modules in Section~\ref{subsec:ad_architecture} and Section~\ref{subsec:ad_mm}.

\subsection{Architecture Overview}
\label{subsec:ad_architecture}

 \begin{figure}[!tbp]
  \centering
  \includegraphics[width=1\linewidth]{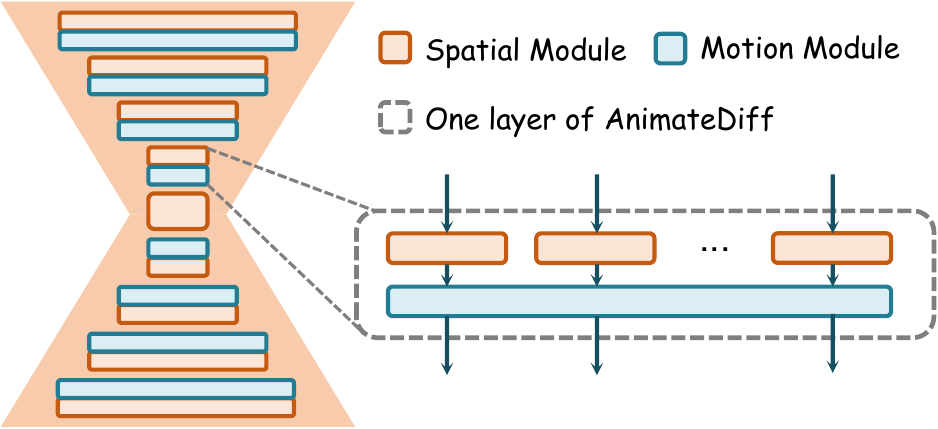}
  \caption{The architecture of the diffusion UNet in AnimateDiff~\cite{guo2023animatediff}. It decouples the video diffusion model into two kinds of modules: the spatial module is responsible for generating appearance, and the motion module is responsible for generating motion.}
\label{fig:ad_achitecture} 
\end{figure}

AnimateDiff~\cite{guo2023animatediff} aims to learn additional temporal information on top of the pretrained large-scale text-to-image model, \ie, stable diffusion~\cite{rombach2022high}, for video generation. 
To achieve this, AnimateDiff decouples the video generation models into the spatial modules and motion modules individually (shown in Fig.~\ref{fig:ad_achitecture}). Respectively, AnimateDiff fixes the parameters of the spatial modules from the original weights of Stable Diffusion and only trains the motion modules inserted into spatial modules to generate several frames all at once. 
For the training dataset, the authors train motion modules of AnimateDiff on the large-scale WebVid dataset~\cite{webvid} with real-world videos to learn the motion prior. 
Interestingly, during inference, we can replace the weights of the original spatial modules~(\ie, the weights of the original Stable Diffusion) with various personalized checkpoints provided by the community, resulting in high visual quality videos in personalized image domains.

\input{all_subtex/method}

\subsection{Details of Motion Module}
\label{subsec:ad_mm}
The magic of AnimateDiff lies in the temporal motion modules for temporally consistent video generation. In detail,
a motion module consists sequentially of a project-in linear layer, two self-attention blocks, and a project-out linear layer, respectively as shown in the middle of Fig.~\ref{fig:architecture}. The self-attention operates in the frame dimension, facilitating interactions between frames.  Because frame-level self-attention is independent across different batches, heights, and widths, for the sake of simplicity, we omit the batch size, height, and width dimensions in our notation. We represent the input of a self-attention as $Z_{in}=\{z_1, z_2, ... , z_f; z_i\in\mathbb{R}^{c\times 1}\}$ where $f$ and $c$ are numbers of frames and channels. The self-attention block first adds position embeddings $P=\{p_1,p_2,...,p_f;p_i\in\mathbb{R}^{c\times 1}\}$ to each input token and then projects them to queries, keys, and values, which can be described by:
\begin{align}
    Q&=\{q_{i}^{i};q_{i}^{i}=W_q(z_i+p_i),1\leq i\leq f\}, \notag \\
    K&=\{k_{i}^{i};k_{i}^{i}=W_k(z_i+p_i),1\leq i\leq f\},  \\
    V&=\{v_{i}^{i};v_{i}^{i}=W_v(z_i+p_i),1\leq i\leq f\}, \notag 
    \label{eq:compute_qkv}
\end{align}
where $W_q$, $W_k$ and $W_v$ are the linear projection parameters. $Q$, $K$ and $V$ represent queries, keys and values. 
The subscript  ``$i$'' and superscript ``$j$'' in $q_{i}^{j}$ indicates the addition of $i$-th input token $z_i$ and $j$-th position embedding $p_j$. Here, we distinguish the serial numbers of tokens and position embeddings for the convenience of the following explanations.
Finally, the calculation of output $Z_{out}$ is:
\begin{align}
    Z_{out}=V\cdot\mathbf{Softmax}(Q^{\top}K/\sqrt{c})^{\top}.
\end{align}
It can be observed that the temporal consistency in AnimateDiff is achieved through weighted operations of self-attention, which average all frames to get smooth results.

\section{Method} 
Using the pre-trained AnimateDiff, our objective is to adapt it for step-by-step video generation with better visual and controllable quality. Specifically, we first generate one satisfactory image, and then utilize the intermediate latents and features of its generation process to guide the video generation.
Our method consists of two parts: the spatial appearance control, discussed in Sec.~\ref{subsec:method-spatial}, modifies the spatial modules to guarantee that the generated first frame is equal to the given image, while the temporal control, described in Sec.~\ref{subsec:method-temporal}, modifies the motion modules to ensure temporal consistency throughout the entire generated video. 

\subsection{Spatial Appearance Control}
\label{subsec:method-spatial}



We first generate an image using the same personalized T2I model in AnimateDiff, so that we can get the generated image $I_1$ and the intermediate latents $\{\mathbf{z}_T^1, ..., \mathbf{z}_{t}^1, ..., \mathbf{z}_0^1\}$ responsible for generating this image. Then, we can use these latents and features for further animation. The goal of spatial appearance control is to ensure that the first frame of the generated video is identical to $I_1$. 
The left part of Fig.~\ref{fig:architecture} illustrates the control mechanism.

\paragraph{Inserting Intermediate Latents.}
To exactly mock the generation process of image animation, for video generation, we discard the originally generated latents of the first frame in each step. 
Instead, we insert the intermediate latents from T2I as replacements. Notice that those intermediate latents of previous steps have not been involved in the temporal modules.  
This approach not only ensures that the final sampled first frame closely resembles $I_1$, but also allows contents of $I_1$ to participate in the computation of temporal attention with other frames at each intermediate step.

\paragraph{Sharing K\&V in Spatial Self-Attention.}
Relying solely on temporal attention within the motion module makes it challenging to align the semantic and style information of other frames with the first frame. Inspired by studies in personalized image generation and editing~\cite{cao_2023_masactrl, wei2023elite}, 
we make spatial modules of all frames share the same keys and values from the spatial self-attention of the first frame. The underlying implication is that each frame draws values from the same sets, implicitly ensuring similar semantic and style across frames.

\subsection{Temporal Consistency Control}
\label{subsec:method-temporal}

While we have made the first frame identical to $I_1$ using spatial appearance control, the motion module introduced in Sec.~\ref{subsec:ad_mm} does not guarantee temporal consistency. 
This is because the weighted operations in self-attention of motion modules are based on the computed similarity between different frames and can not automatically align other frames to a specific frame.
In order to align other frames with the first frame explicitly, we propose the Positional-Corrected Window Attention to modify the original global attention (shown in the right part of Fig.~\ref{fig:architecture}), which will be introduced in detail below.

\paragraph{From Global Attention to Window Attention.}
First, we need to provide the formula for the self-attention calculation in the motion module, where query, key, value, and output are denoted as $Q$, $K$, $V$, and $Z_{out}$, respectively. The specific form is as follows:
\begin{align}
    Q&=\{q_{1}^{1},q_{2}^{2},...,q_{f}^{f};q_{i}^{i}\in\mathbb{R}^{c\times 1}\},& Q\in\mathbb{R}^{c\times f}, \notag \\
    K&=\{k_{1}^{1},k_{2}^{2},...,k_{f}^{f};k_{i}^{i}\in\mathbb{R}^{c\times 1}\},& K\in\mathbb{R}^{c\times f}, \notag \\
    V&=\{v_{1}^{1},v_{2}^{2},...,v_{f}^{f};v_{i}^{i}\in\mathbb{R}^{c\times 1}\},& V\in\mathbb{R}^{c\times f}, \notag \\
    Z_{out}&=\{\hat{z}_{1},\hat{z}_{2},...,\hat{z}_{f};\hat{z}_{i}\in\mathbb{R}^{c\times 1}\},&Z_{out}\in\mathbb{R}^{c\times f}, \notag 
\end{align}
where $c$ and $f$ represent the numbers of channels and frames.
The output $\hat{z}_i$ for the $i$-th frame can be written as:
\begin{equation}
    \hat{z}_i=V\cdot\mathbf{Softmax}((q_{i}^{i})^{\top}K/\sqrt{c})^{\top}.
    \label{eq:one_frame_attention}
\end{equation}
From Eq.~\ref{eq:one_frame_attention}, it can be observed that the attention calculation range for each frame is global, meaning $K$ and $V$ include keys and values from all frames (shown in Fig.~\ref{fig:architecture} (a)). Although this global design helps in averaging all frames to achieve a smooth result, it hinders the ability to align with the first frame.
Therefore, our proposed improvement is the introduction of window attention (shown in Fig.~\ref{fig:architecture} (b)), where the sources of keys and values for the calculation of the $i$-th output are limited to the preceding $i$ frames. 
The specific formula can be written as:
\begin{equation}
    \hat{z}_i=\Tilde{V}_i\cdot\mathbf{Softmax}((q_{i}^{i})^{\top}\Tilde{K}_i/\sqrt{c})^{\top},
    \label{eq:one_frame_attention_modified}
\end{equation}
where $\Tilde{K}_i,\Tilde{V}_i\in\mathbb{R}^{c\times f}$ can be written as:
\begin{equation}
\Tilde{K}_i=\{
    \underbrace{k_{1}^{1},...,k_{1}^{1}}_{(f-i+1)},...,k_{i}^{i}\},
\Tilde{V}_i=\{\underbrace{v_{1}^{1},...,v_{1}^{1}}_{(f-i+1)},...,v_{i}^{i}\}.
\label{eq:modify_kv}
\end{equation}
As described in Eq.~\ref{eq:modify_kv},
we duplicate tokens from the first frame to ensure that the number of tokens in both $\Tilde{K}_i$ and $\Tilde{V}_i$ remains equal to $f$, emphasizing its importance during the attention computation, which further promotes alignment of other frames with the first frame.

\paragraph{Correct Position Embedding Makes Better Results.}
Our design philosophy for the zero-shot module modification aims to ensure the operations remain unchanged from the original AnimateDiff. The local attention introduced above still has some limitations.
The issue lies in the positional embeddings. 
Ideally, a set of keys and values should include all possible positional embeddings from $p_1$ to $p_f$. However, because the position embeddings are added before attention calculation, 
the $i$-th token only carry $i$-th position embedding. 
Therefore, $\Tilde{K}_i$ and $\Tilde{V}_i$ described in Eq.~\ref{eq:modify_kv} include only the first $i$ positions. 

Based on this observation, we modified the mechanism for adding positional embeddings (details can be found in supplementary materials) for queries, keys, and values, so that the $i$-th token is added with the $j$-th positional embedding ($i$ may not be equal to $j$). 
In the end, we achieved that the $f$ tokens in $\Tilde{K}_i$ and $\Tilde{V}_i$ could carry positional embeddings from the $1$-st to the $f$-th position, illustrated in Fig.~\ref{fig:architecture} (b) and written as:
\begin{align}
    \Tilde{K}_i=\{
    k_{1}^{1},k_{1}^{2},...,k_{1}^{f-i+1},k_{2}^{f-i+2}...,k_{i}^{f}\},\notag\\
\Tilde{V}_i=\{v_{1}^{1},v_{1}^{2},...,v_{1}^{f-i+1},v_{2}^{f-i+2}...,v_{i}^{f}\}.
\label{eq:final_kv}
\end{align}

Although proposed window attention has shown significant advantages over global attention in aligning other frames with the first frame, global attention tends to produce smoother results, enhancing the visual quality of the output. As we still need to increase the overall consistency via the global solution, our final solution integrates the strengths of both attentions into a Diffusion UNet. Specifically, we use a motion module with local attention in the encoder part of the UNet to align each frame with the first frame. In the decoder, we utilize a motion module with global attention to smooth all frames. We also find the time-travel sampling strategies will produce smoother results as discussed in \cite{wang2023zeroshot, yu2023freedom}, which we give more additional experiments in the supplementary.


\subsection{Discussion}
From the proposed method, we can successfully give more control handles to the T2V generation. Also, since we find that the video diffusion model is an image animator, our method can also be considered as an image animation method for the generated image. Given the real image, we can also perform DDIM inversion~\cite{ddim, mokady2023null} to get the intermediate latents. Moreover, our approach, particularly the aspect related to temporal consistency control, has the potential to inspire the training of video foundation models, leading to improved training-based image-to-video models.









%% file: all_subtex/method.tex
\begin{figure*}[t]
  \centering
  \includegraphics[width=1\linewidth]{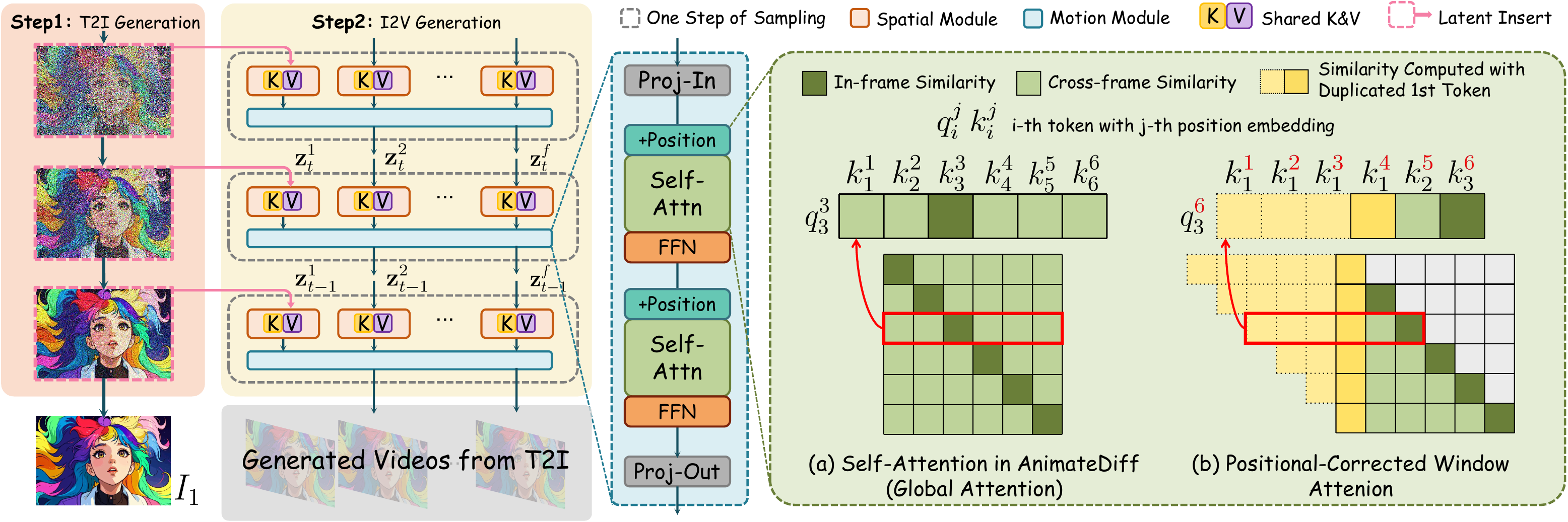}
  \vspace{-0.6cm}
  \caption{The overall pipeline of our proposed AnimateZero. Given spatial modules from a pre-trained T2I model~\cite{rombach2022high} and its corresponding motion modules~\cite{guo2023animatediff}, we first generate a single image $I_1$ using the T2I model (\textbf{step1}) and then generate a video animated from this image (\textbf{step2}). 
  The \textbf{Left} part shows the image generation process with the intermediate latents $\{\mathbf{z}_T^1, ..., \mathbf{z}_{0}^1\}$ and our proposed Spatial Appearance Control (Sec.~\ref{subsec:method-spatial}). Spatial Appearance Control makes modifications to the spatial modules, including the latent insertion for ensuring the first frame equal to $I_1$ and sharing keys and values from spatial self-attention of the first frame across other frames to align both semantics and styles. 
  \textbf{Right} part is the Temporal Consistency Control (Sec.~\ref{subsec:method-temporal}). We propose modifications to the original self-attention in AnimateDiff~\cite{guo2023animatediff}, which is a global attention and illustrated in (a).
  Our modifications include three key points (illustrated in (b)): (1) we replace global attention to window attention, which computes the $i$-th output token only using preceding $i$ frames; (2) we duplicate the similarity computed with the first token to emphasize the importance of the first frame $I_1$; (3) we correct the position embeddings (marked as \textcolor{red}{red} in the superscripts of $q$ and $k$, and the calculation of qkv is described by Eq.~\ref{eq:compute_qkv}) added to input tokens to get better results.
  }
  \vspace{-0.5cm}
\label{fig:architecture} 
\end{figure*}

%% file: tex/4_exp.tex
\input{all_subtex/compare_with_animate_diff}
\input{all_subtex/ADvsAZ}
\section{Experiments}
\subsection{Implementation and Setting Details}
In our experiments, spatial modules are based on Stable Diffusion V1.5~\cite{rombach2022high}, and motion modules use the corresponding AnimateDiff~\cite{guo2023animatediff} checkpoint V2. We experiment with various personalized T2I checkpoints downloaded from Civitai~\cite{civitai}, and detailed information about these checkpoints can be found in the supplementary materials.
For AnimateZero, utilizing both spatial appearance control and temporal consistency control is sufficient to achieve satisfactory results in most cases, without involving any hyper-parameters to be chosen. 
The length for our generated videos is $16$ frames, and the video resolution is unrestricted, with a standard resolution of $512 \times 512$.

\subsection{Comparison Results}
We construct a benchmark for quantitative comparison, which includes 20 prompts and 20 corresponding generated images. To achieve a comprehensive evaluation, these prompts and images include different styles (realistic and cartoon styles) and contents (characters, animals, and landscapes).
Regarding evaluation metrics in Tab.~\ref{table:compare_all} and Tab.~\ref{table:ad-az}, we design: 
(1) \textbf{`$I_1$-MSE'} uses MSE to measure whether the generated first frame matches the given image $I_1$; 
(2) \textbf{`Warping Error'}~\cite{Lai-ECCV-2018} evaluates the temporal consistency of the generated videos; 
(3) \textbf{`Text-Sim'} evaluates the similarity between the prompt and each generated frame using their features extracted by CLIP~\cite{radford2021learning} Text and Image Encoders; 
(4) \textbf{`Domain-Sim'} assesses the similarity between the T2I domain and the generated videos. We first use the T2I model to generate 16 images and then calculate and average the CLIP feature similarity between each of these images and each frame of the generated video; 
(5) \textbf{`Style-Dist'} evaluates the style matching degree between the each generated frame and the given image $I_1$, by calculating the distance between their style information which is represented by the gram matrix of the third layer features of the CLIP Image Encoder; 
(6) \textbf{`User Study'}, which is divided into three aspects: \textit{Motion} evaluates the quality of the generated motion, \textit{Appearance} assesses whether the generated appearance matches the given image $I_1$, and \textit{Subjective} evaluates the subjective quality of the generated videos. We ask 20 subjects to rank different methods in these three aspects and use the average rank number to evaluate each method.

\input{all_subtex/AZvsOthers}
\input{all_subtex/comparison_all}

\paragraph{Compared with AnimateDiff.}
While AnimateDiff~\cite{guo2023animatediff} demonstrates good generalization ability on many personalized T2I models, it occasionally produces low-quality videos (shown in Fig.~\ref{fig:ADvsAZ}), especially on anime-style T2I models. These low-quality videos mainly manifest in two aspects: (1) the generated videos are not within the same domain as the original T2I models; (2) a decrease in text-frame alignment in the generated videos. Surprisingly, in our experiments, we find that AnimateZero excels in both of these aspects compared to AnimateDiff, which has been demonstrated in Fig.~\ref{fig:ADvsAZ}. In Tab.~\ref{table:ad-az}, we also quantitatively evaluate AnimateDiff and AnimateZero on our benchmark at four metrics. Our proposed AnimateZero outperforms AnimateDiff in all four metrics in terms of text-frame alignment and matching degree between the generated videos and original T2I domains.

\paragraph{Compared with Publicly Available I2V Tools.}
Existing I2V methods claim to be versatile but still struggle with domain gap issues. In our experiments, we use the generated image as a condition for video creation, ensuring alignment with the T2I domain. This aims to explore AnimateZero's advantages over existing I2V methods and highlight their limitations. We compare AnimateZero with several publicly available image-to-video tools, both closed-source (Gen-2~\cite{gen2}, Genmo~\cite{genmo}, Pika Labs~\cite{pikalabs}) and open-source (VideoCrafter~\cite{chen2023videocrafter1}, I2VGen-XL~\cite{i2vgenxl}), using benchmark images and their corresponding prompts.
In terms of subjective quality, as shown in Fig.~\ref{fig:ADvsOthers}, our proposed AnimateZero achieves performance comparable to, or even better than, the current state-of-the-art Gen-2 and Pika Labs, standing out as the best among open-source tools. 
In contrast, Genmo, VideoCrafter and I2VGen-XL can only leverage the semantic information of the given generated images, failing to ensure the first frame matches the given image.
Gen-2, Genmo, VideoCrafter and I2VGen-XL suffer from domain gap issues, particularly noticeable in anime-style images, whereas AnimateZero does not encounter this problem.
We also conduct a comprehensive evaluation of AnimateZero and these I2V methods across all metrics in Tab.~\ref{table:compare_all}. It can be observed that our proposed AnimateZero achieves the best performance in certain metrics and is comparable to the best methods in other metrics. Considering that AnimateZero is a method that does not require additional training specifically for image animation, achieving the mentioned performance is highly remarkable.

\subsection{Ablation Study}

\input{all_subtex/ablation_study}
We conduct ablation experiments on the spatial appearance control (introduced in Sec.~\ref{subsec:method-spatial}) and temporal consistency control (introduced in Sec.~\ref{subsec:method-temporal}). The experimental results are shown in Fig.~\ref{fig:ablation} to illustrate the role of each component in our proposed method. 
Firstly, Fig.~\ref{fig:ablation} (a) shows the results generated by AnimateDiff with the provided text, which serves as the baseline for our ablation experiments. We will demonstrate the step-by-step process of incorporating our proposed techniques to achieve animation of the generated image. 
In Fig.~\ref{fig:ablation} (b), we insert the intermediate latents, making the first frame almost identical to the generated image. This also implicitly controls the content and style of the other frames. However, notable differences persist in terms of style and colors when compared to the generated image. In Fig.~\ref{fig:ablation} (c), we employ the strategy of sharing keys and values, further aligning the style and semantic information between the first frame and other frames.
However, the spatial appearance control mentioned above cannot guarantee a seamless connection between the first frame and the rest frames.  This is where our temporal consistency control (TCC) comes into play. We first attempt TCC without position correction (TCC w/o PC) in Fig.~\ref{fig:ablation} (d), which ensures the temporal connection of the first several frames. However, the quality of frames towards the end of the video significantly deteriorates. This is addressed by employing TCC with position correction (TCC w/ PC) in Fig.~\ref{fig:ablation} (e).

\subsection{Limitations}
Although our method enables the possibility of both controllable video generation and image animation, there are still some limitations. 
These limitations mainly stem from the constraints in motion prior within AnimateDiff~\cite{guo2023animatediff}. AnimateDiff struggles to generate complex motions, such as sports movements or the motion of uncommon objects (demonstrated in Fig.~\ref{fig:limitation}). In theory, since the generated motion of AnimateZero relies on motion prior of AnimateDiff, AnimateZero is also less proficient in creating videos in the mentioned scenarios. 
However, we believe these limitations can be solved with a better video foundation model with more powerful motion prior.

\input{all_subtex/limitation}

%% file: all_subtex/compare_with_animate_diff.tex

\if \animation 1
\begin{figure*}[t]
  \centering
  \begin{tabular}{c@{\hspace{0.3em}}c@{\hspace{0.3em}}c@{\hspace{0.8em}}c@{\hspace{0.3em}}c@{\hspace{0.3em}}c}
  Generated Images & AnimateDiff~\cite{guo2023animatediff} & AnimateZero(ours)& Generated Images & AnimateDiff~\cite{guo2023animatediff} & AnimateZero(ours)\vspace{0.3em} \\
    \includegraphics[width=0.16\linewidth]{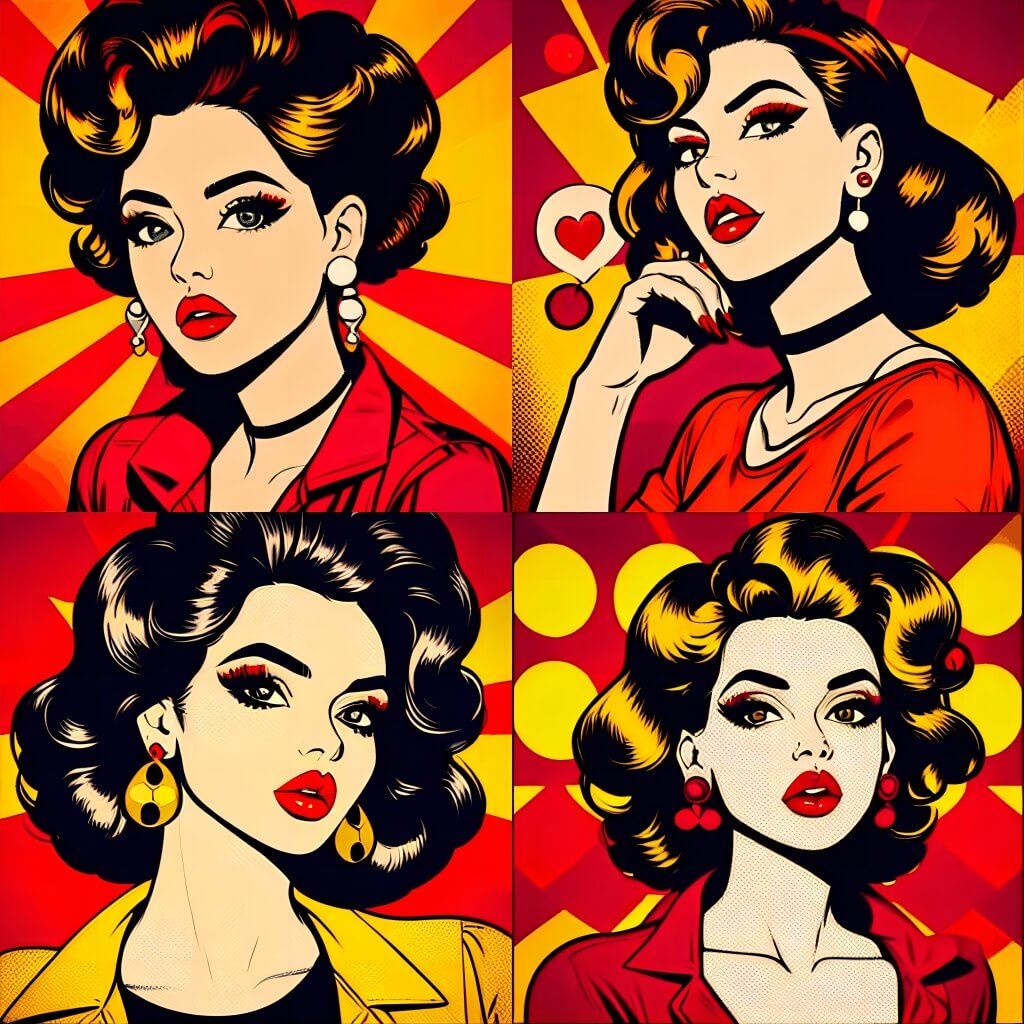}&
    \animategraphics[width=0.16\linewidth]{8}{gif/ADvsAZ/526/526-}{0}{15}&
     \animategraphics[width=0.16\linewidth]{8}{gif/ADvsAZ/525/525-}{0}{15}&
     \includegraphics[width=0.16\linewidth]{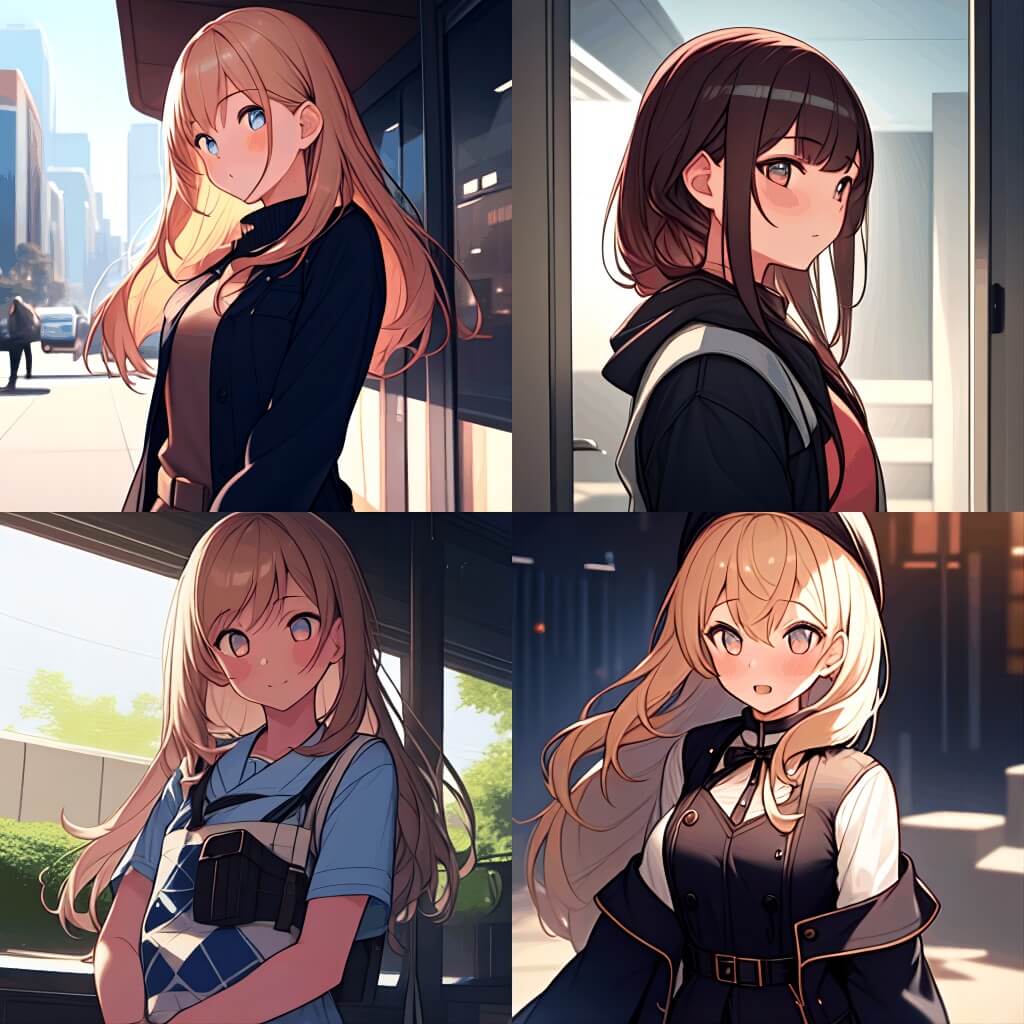}&
      \animategraphics[width=0.16\linewidth]{8}{gif/ADvsAZ/528/528-}{0}{15}&
       \animategraphics[width=0.16\linewidth]{8}{gif/ADvsAZ/531/531-}{0}{15}
       \\
       \multicolumn{3}{l}{(a) \textit{``1girl, jewelry, \textcolor{red}{upper body}, earrings, pop art, ...''}} & \multicolumn{3}{l}{(b) \textit{``1girl, long hair, looking at the camera, ...''}}\vspace{0.5em}\\
       \includegraphics[width=0.16\linewidth]{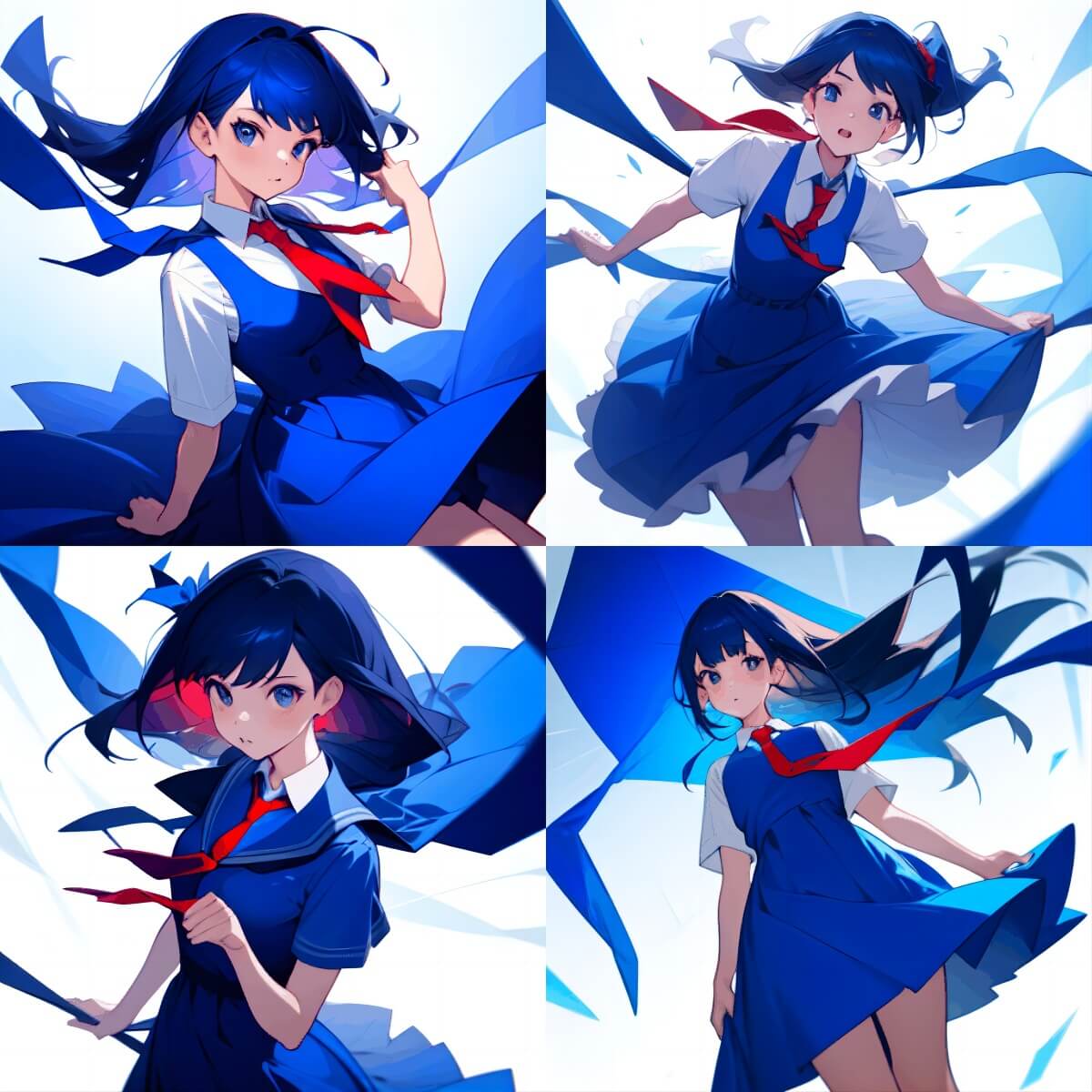}&
       \animategraphics[width=0.16\linewidth]{8}{gif/ADvsAZ/532/532-}{0}{15}&
     \animategraphics[width=0.16\linewidth]{8}{gif/ADvsAZ/533/533-}{0}{15}&
     \includegraphics[width=0.16\linewidth]{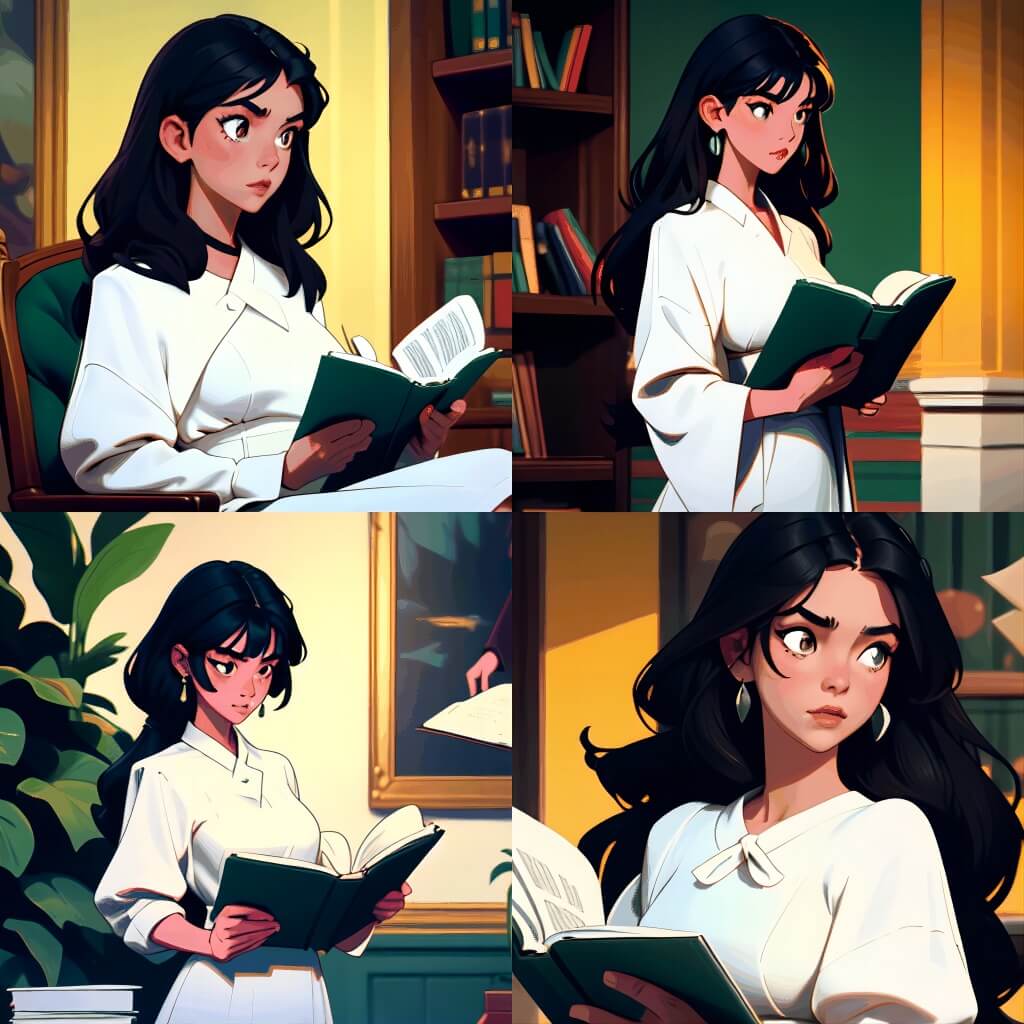}&
      \animategraphics[width=0.16\linewidth]{8}{gif/ADvsAZ/545/545-}{0}{15}&
       \animategraphics[width=0.16\linewidth]{8}{gif/ADvsAZ/549/549-}{0}{15}\\
       \multicolumn{3}{l}{(c) \textit{``1girl, blue dress, \textcolor{red}{red tie}, floating blue, ...''}} & \multicolumn{3}{l}{(d) \textit{``1girl wearing white \textcolor{red}{dress} is reading \textcolor{red}{green} book, ...''}}
    \end{tabular}
\vspace{-0.5em}
  \caption{Qualitative comparison results between AnimateDiff~\cite{guo2023animatediff} and our proposed AnimateZero. As shown in (a), (b) and (c), the videos generated by AnimateDiff are not in the same domain as the generated images. In contrast, AnimateZero is capable of maintaining consistency with the original T2I domains; In (a), (c) and (d), it is demonstrated that AnimateDiff may encounter inconsistencies between the provided text and the generated frames (highlighted in \textcolor{red}{red}). AnimateZero, on the other hand, performs better in this regard. \textit{Best viewed with Acrobat Reader. Click the video to play the animation clips. \textbf{Static frames are provided in supplementary materials.}}}
  \vspace{-1em}
\label{fig:ADvsAZ} 
\end{figure*}

\else
\begin{figure*}[t]
  \centering
  \begin{tabular}{c@{\hspace{0.3em}}c@{\hspace{0.3em}}c@{\hspace{0.8em}}c@{\hspace{0.3em}}c@{\hspace{0.3em}}c}
  Generated Images & AnimateDiff~\cite{guo2023animatediff} & AnimateZero(ours)& Generated Images & AnimateDiff~\cite{guo2023animatediff} & AnimateZero(ours)\vspace{0.3em} \\
    \includegraphics[width=0.16\linewidth]{gif/ADvsAZ/mix-001.jpg}&
    \includegraphics[width=0.16\linewidth]{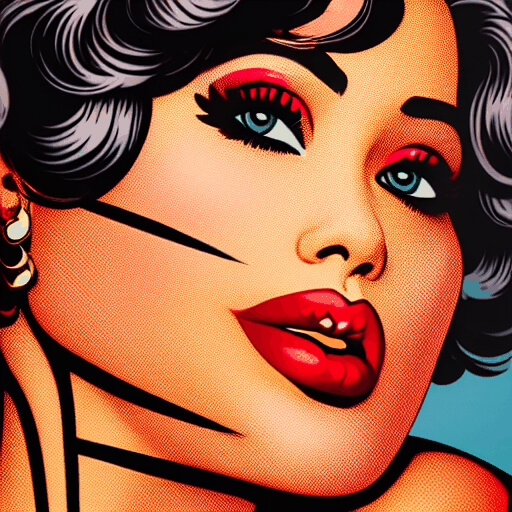}&
     \includegraphics[width=0.16\linewidth]{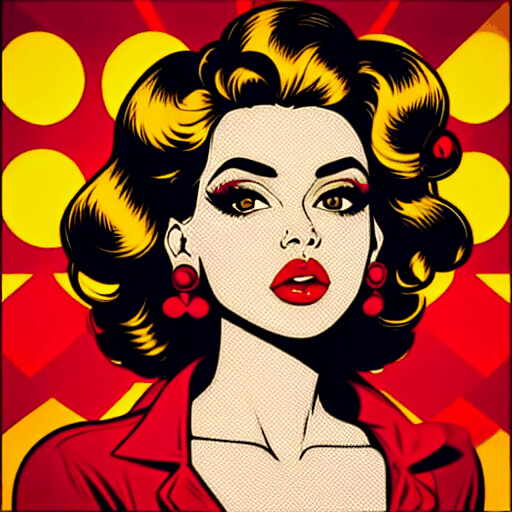}&
     \includegraphics[width=0.16\linewidth]{gif/ADvsAZ/mix-002.jpg}&
      \includegraphics[width=0.16\linewidth]{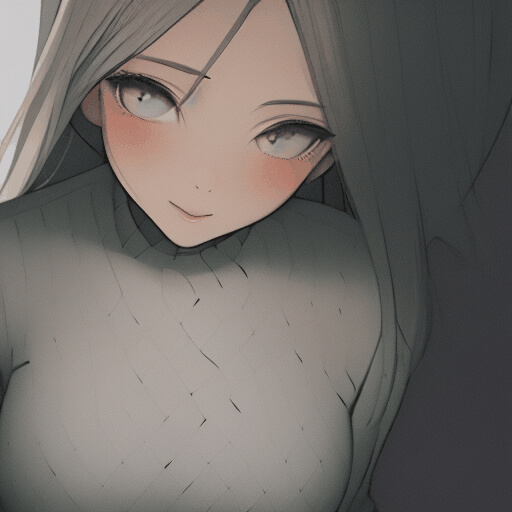}&
       \includegraphics[width=0.16\linewidth]{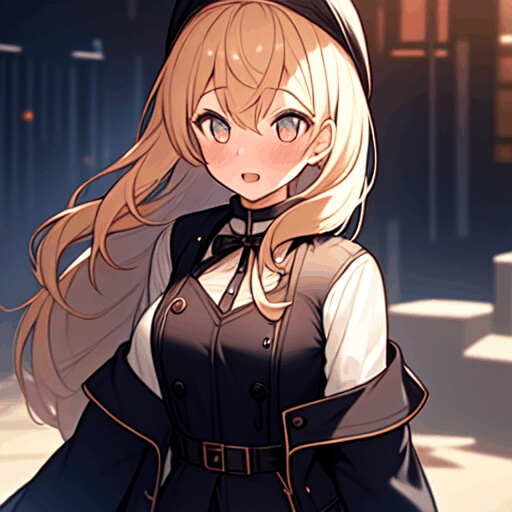}
       \\
       \multicolumn{3}{l}{(a) \textit{``1girl, jewelry, \textcolor{red}{upper body}, earrings, pop art, ...''}} & \multicolumn{3}{l}{(b) \textit{``1girl, long hair, looking at the camera, ...''}}\vspace{0.5em}\\
       \includegraphics[width=0.16\linewidth]{gif/ADvsAZ/mix-003.jpg}&
       \includegraphics[width=0.16\linewidth]{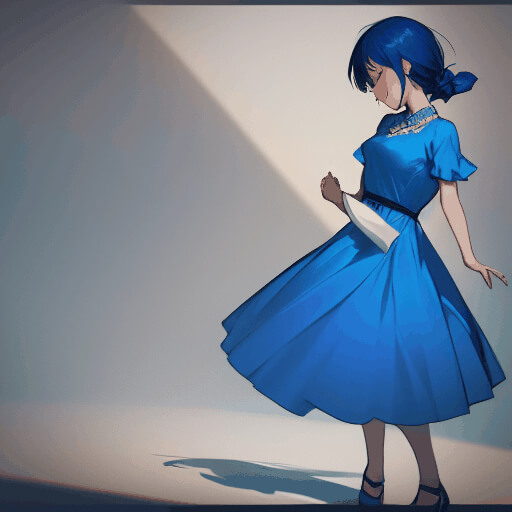}&
     \includegraphics[width=0.16\linewidth]{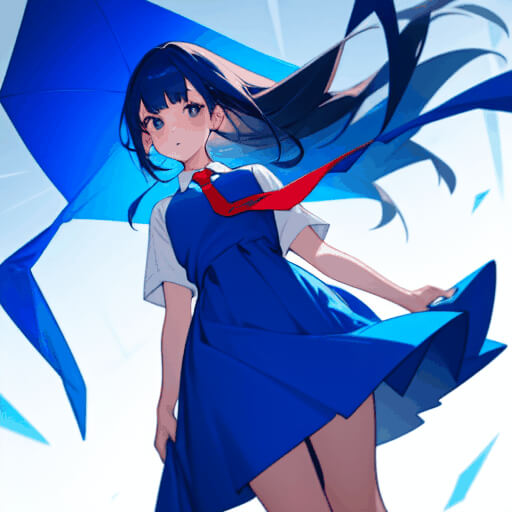}&
     \includegraphics[width=0.16\linewidth]{gif/ADvsAZ/mix-004.jpg}&
      \includegraphics[width=0.16\linewidth]{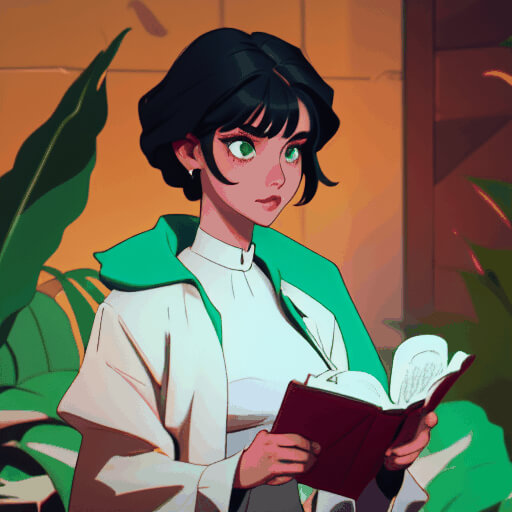}&
       \includegraphics[width=0.16\linewidth]{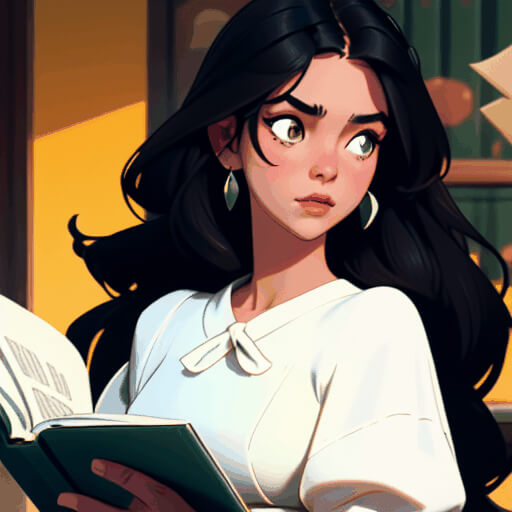}\\
       \multicolumn{3}{l}{(c) \textit{``1girl, blue dress, \textcolor{red}{red tie}, floating blue, ...''}} & \multicolumn{3}{l}{(d) \textit{``1girl wearing white \textcolor{red}{dress} is reading \textcolor{red}{green} book, ...''}}
    \end{tabular}
\vspace{-0.5em}
  \caption{Qualitative comparison results between AnimateDiff~\cite{guo2023animatediff} and our proposed AnimateZero. As shown in (a), (b) and (c), the videos generated by AnimateDiff are not in the same domain as the generated images. In contrast, AnimateZero is capable of maintaining consistency with the original T2I domains; In (a), (c) and (d), it is demonstrated that AnimateDiff may encounter inconsistencies between the provided text and the generated frames (highlighted in \textcolor{red}{red}). AnimateZero, on the other hand, performs better in this regard. \textit{Best viewed with Acrobat Reader. Click the video to play the animation clips. \textbf{Static frames are provided in supplementary materials.}}}
  \vspace{-1em}
\label{fig:ADvsAZ} 
\end{figure*}

\fi

%% file: all_subtex/ADvsAZ.tex
\begin{table}[t]
    \centering
    \scriptsize
    \tabcolsep=0.09cm
    \begin{tabular}{l | c c c c}
        \toprule
        \rowcolor{color3}Method & Warping Error $\downarrow$ & Text-Sim $\uparrow$ & Domain-Sim $\uparrow$ & Style-Dist $\downarrow$\\
        \hline \hline
        AnimateDiff~\cite{guo2023animatediff} & $0.6719$ & $0.3254$ &  $0.8081$ & $0.3809$ \\
        AnimateZero (ours) & $\mathbf{0.6562}$ & $\mathbf{0.3314}$ & $\mathbf{0.8671}$ & $\mathbf{0.1666}$\\
        \bottomrule
    \end{tabular}
    \vspace{-1em}
    \caption{Quantitative comparison results between AnimateDiff~\cite{guo2023animatediff} and our proposed AnimateZero. AnimateZero exhibits a higher similarity to the text and the original T2I domain.
    }
    \vspace{-1em}
    \label{table:ad-az}
\end{table}

%% file: all_subtex/AZvsOthers.tex
\if \animation 1

\begin{figure*}[t]
  \centering
  \begin{tabular}{c@{\hspace{0.1em}}c@{\hspace{0.1em}}c@{\hspace{0.1em}}c@{\hspace{0.1em}}c@{\hspace{0.1em}}c@{\hspace{0.1em}}c}
  Generated Image & Gen-2~\cite{gen2} & Genmo~\cite{genmo} & Pika Labs~\cite{pikalabs} & {\small VideoCrafter1}~\cite{chen2023videocrafter1} & {\small I2VGen-XL}~\cite{i2vgenxl} & {\small AnimateZero(ours)}\\
    \includegraphics[width=0.14\linewidth]{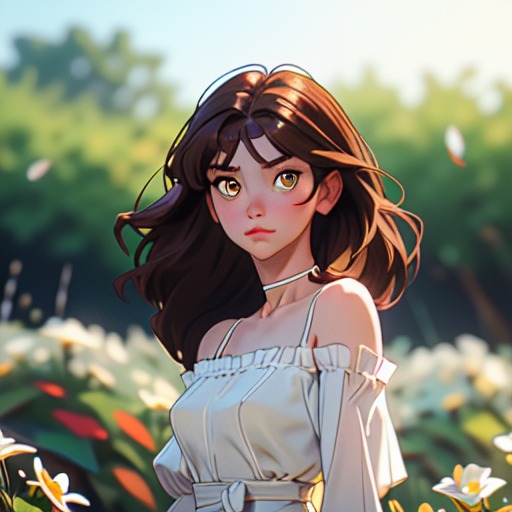}&
    \animategraphics[width=0.14\linewidth]{8}{gif/AZvsOthers/line3/gen2/011_}{0}{15}&
     \animategraphics[width=0.14\linewidth]{8}{gif/AZvsOthers/line3/genmo/011_}{0}{15}&
     \animategraphics[width=0.14\linewidth]{8}{gif/AZvsOthers/line3/pika/011_}{0}{15}&
     \animategraphics[width=0.14\linewidth]{8}{gif/AZvsOthers/line3/videocrafter/011_}{0}{15}&
      \animategraphics[width=0.14\linewidth]{8}{gif/AZvsOthers/line3/i2vgenxl/011_}{0}{15}&
       \animategraphics[width=0.14\linewidth]{8}{gif/AZvsOthers/line3/az/011_}{0}{15}
       \\
       \multicolumn{7}{c}{\textit{``1girl, brown hair, a lot of white flowers, leaf, blurry foreground, ...''}} \vspace{0.5em}\\
       \includegraphics[width=0.14\linewidth]{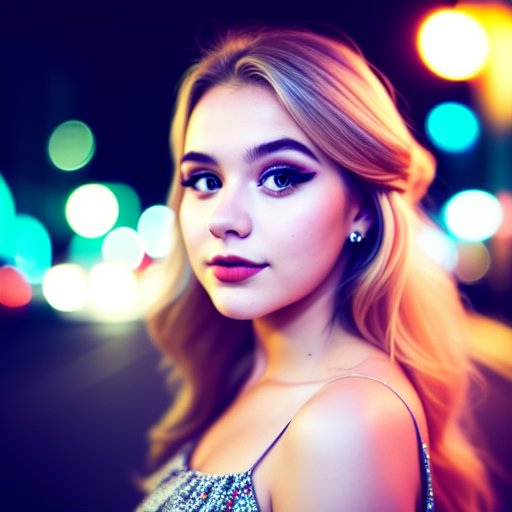}&
       \animategraphics[width=0.14\linewidth]{8}{gif/AZvsOthers/line2/gen2/007_}{0}{15}&
     \animategraphics[width=0.14\linewidth]{8}{gif/AZvsOthers/line2/genmo/007_}{0}{15}&
     \animategraphics[width=0.14\linewidth]{8}{gif/AZvsOthers/line2/pika/007_}{0}{15}&
     \animategraphics[width=0.14\linewidth]{8}{gif/AZvsOthers/line2/videocrafter/007_}{0}{15}&
      \animategraphics[width=0.14\linewidth]{8}{gif/AZvsOthers/line2/i2vgenxl/007_}{0}{15}&
       \animategraphics[width=0.14\linewidth]{8}{gif/AZvsOthers/line2/az/015_frame_}{0}{15}
       \\
       \multicolumn{7}{c}{\textit{``closeup face photo of 18 y.o swedish woman in dress, makeup, night city street, motion blur, ...''}}
    \end{tabular}
\vspace{-0.7em}
  \caption{Qualitative comparison results between publicly available image-to-video tools and our proposed AnimateZero. \textit{Best viewed with Acrobat Reader. Click the video to play the animation clips. \textbf{Static frames are provided in supplementary materials.}}}
\label{fig:ADvsOthers} 
\end{figure*}

\else

\begin{figure*}[t]
  \centering
  \begin{tabular}{c@{\hspace{0.1em}}c@{\hspace{0.1em}}c@{\hspace{0.1em}}c@{\hspace{0.1em}}c@{\hspace{0.1em}}c@{\hspace{0.1em}}c}
  Generated Image & Gen-2~\cite{gen2} & Genmo~\cite{genmo} & Pika Labs~\cite{pikalabs} & {\small VideoCrafter1}~\cite{chen2023videocrafter1} & {\small I2VGen-XL}~\cite{i2vgenxl} & {\small AnimateZero(ours)}\\
    \includegraphics[width=0.14\linewidth]{gif/AZvsOthers/line3/011.jpg}&
    \includegraphics[width=0.14\linewidth]{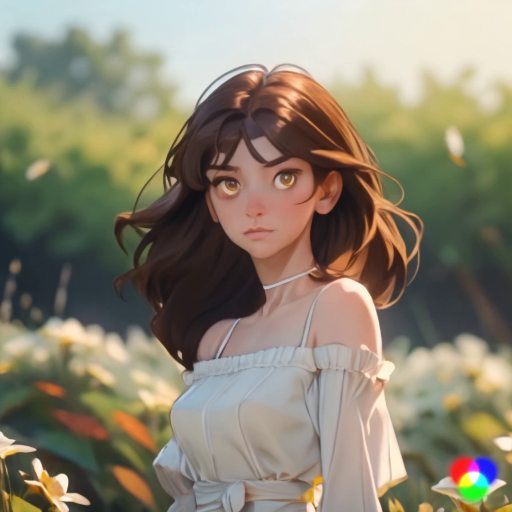}&
     \includegraphics[width=0.14\linewidth]{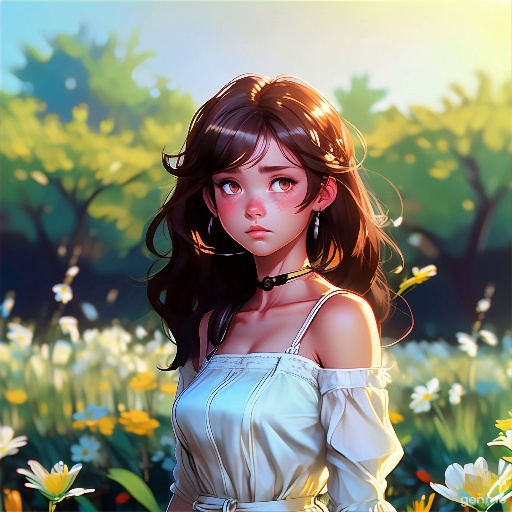}&
     \includegraphics[width=0.14\linewidth]{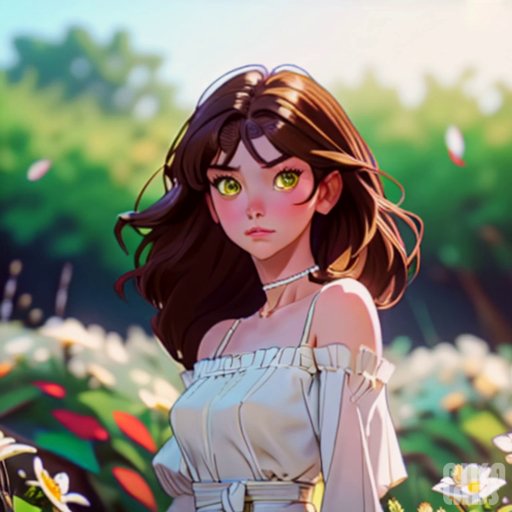}&
     \includegraphics[width=0.14\linewidth]{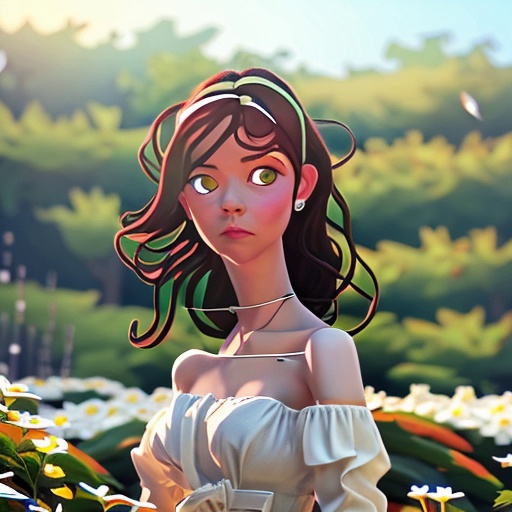}&
      \includegraphics[width=0.14\linewidth]{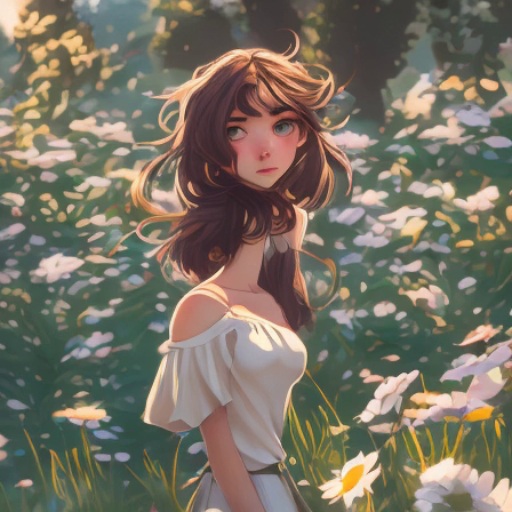}&
       \includegraphics[width=0.14\linewidth]{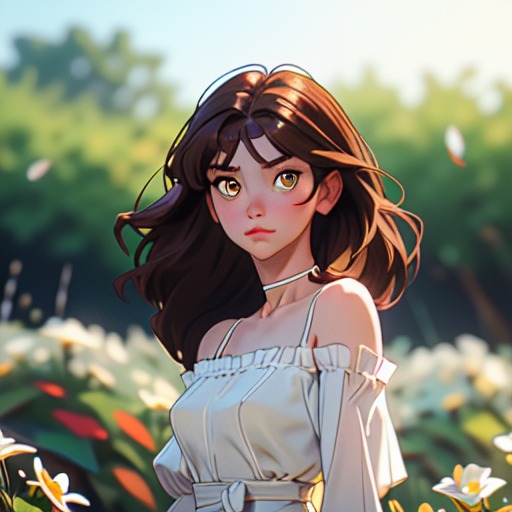}
       \\
       \multicolumn{7}{c}{\textit{``1girl, brown hair, a lot of white flowers, leaf, blurry foreground, ...''}} \vspace{0.5em}\\
       \includegraphics[width=0.14\linewidth]{gif/AZvsOthers/line2/007.jpg}&
       \includegraphics[width=0.14\linewidth]{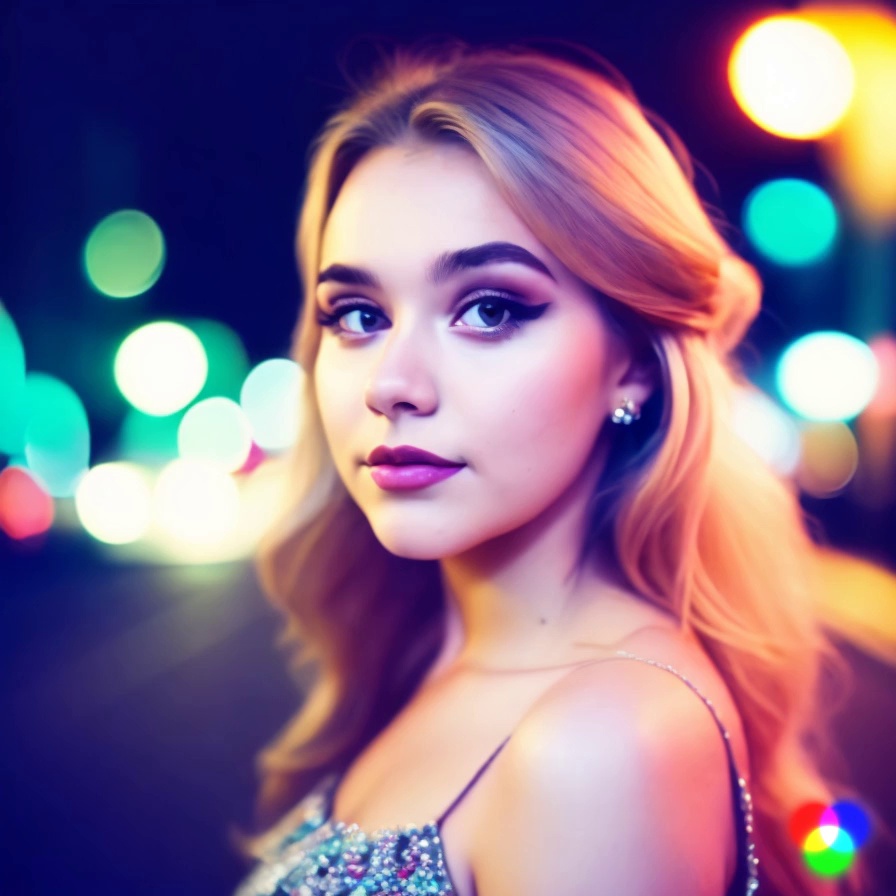}&
     \includegraphics[width=0.14\linewidth]{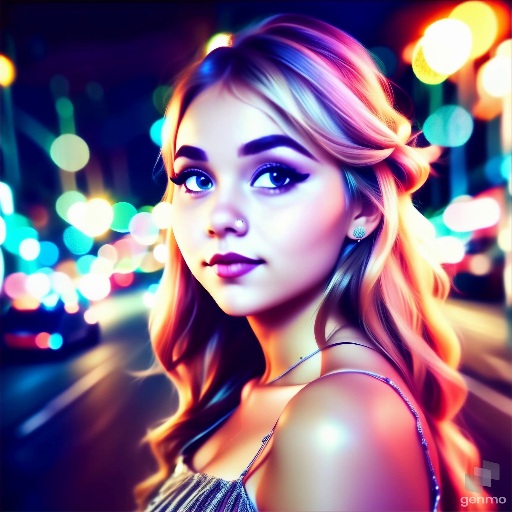}&
     \includegraphics[width=0.14\linewidth]{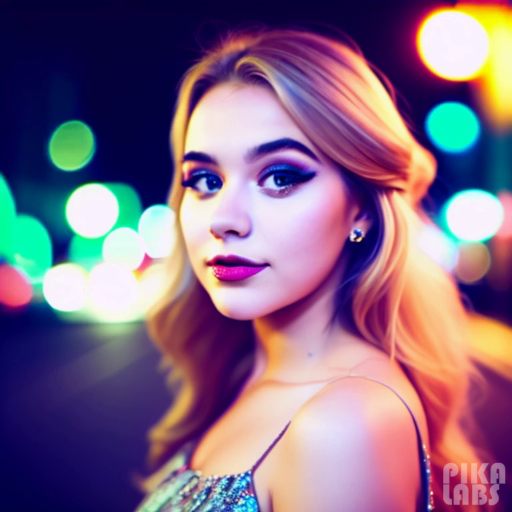}&
     \includegraphics[width=0.14\linewidth]{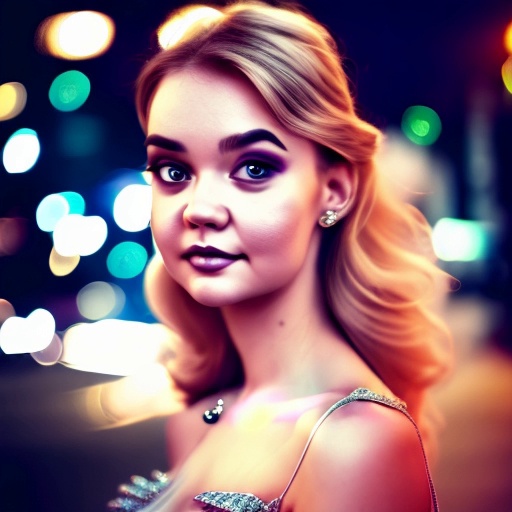}&
      \includegraphics[width=0.14\linewidth]{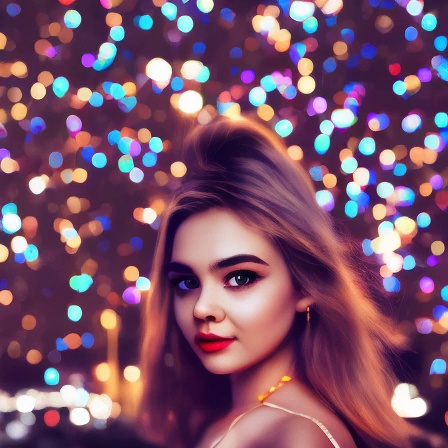}&
       \includegraphics[width=0.14\linewidth]{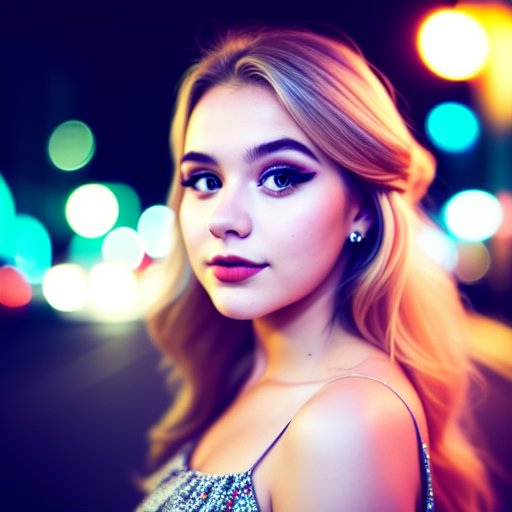}\\
       \multicolumn{7}{c}{\textit{``closeup face photo of 18 y.o swedish woman in dress, makeup, night city street, motion blur, ...''}}
    \end{tabular}
\vspace{-0.7em}
  \caption{Qualitative comparison results between publicly available image-to-video tools and our proposed AnimateZero. \textit{Best viewed with Acrobat Reader. Click the video to play the animation clips. \textbf{Static frames are provided in supplementary materials.}}}
\label{fig:ADvsOthers} 
\end{figure*}

\fi

%% file: all_subtex/comparison_all.tex
\begin{table*}[t]\footnotesize

\vspace{-0.2cm}
\center
\begin{tabular}{l | c c | c c c | c c c}
\toprule
\rowcolor{color3}&  \multicolumn{2}{c}{Basic Metrics } & \multicolumn{3}{c}{CLIP Metrics }& \multicolumn{3}{c}{User Study}\\
\rowcolor{color3} Method & $I_1$-MSE$\downarrow$ & Warping Error$\downarrow$ & Text-Sim$\uparrow$ & Domain-Sim$\uparrow$ & Style-Dist$\downarrow$ & Motion$\downarrow$ & Appearance$\downarrow$ & Subjective$\downarrow$  \\
\hline \hline
 Gen-2~\cite{gen2} & $59.93$ & $0.7353$ & $0.3282$ & $0.7796$ & $0.1707$ & $3.57$ & $2.52$ & $2.88$ \\
Genmo~\cite{genmo} & $90.76$ & $0.8284$ & $0.3184$ & $0.7801$ & $0.2752$ & $\mathbf{\color{blue}2.96}$ & $3.51$ & $3.21$ \\
Pika Labs~\cite{pikalabs} & $\mathbf{\color{blue}37.68}$ & $\mathbf{\color{red}0.6018}$ & $\mathbf{\color{red}0.3372}$ & $\mathbf{\color{blue}0.7876}$ & $\mathbf{\color{red}0.1275}$ & $3.71$ & $\mathbf{\color{blue}2.18}$ & $\mathbf{\color{blue}2.84}$ \\
\hline 
 VideoCrafter1~\cite{chen2023videocrafter1} & $96.23$ & $0.6596$ & $\mathbf{\color{blue}0.3325}$ & $0.7598$ &  $0.2762$ & $4.29$ & $5.09$ & $4.91$ \\
I2VGen-XL~\cite{i2vgenxl} & $104.8$ & $0.7724$ & $0.3009$ & $0.7272$ & $0.4308$ & $4.63$ & $5.79$ & $5.38$ \\
\hline
AnimateZero (Ours) & $\mathbf{\color{red}1.136}$ & $\mathbf{\color{blue}0.6562}$ & $0.3314$ & $\mathbf{\color{red}0.8671}$ & $\mathbf{\color{blue}0.1666}$ & $\mathbf{\color{red}1.83}$ & $\mathbf{\color{red}1.91}$ & $\mathbf{\color{red}1.78}$ \\
\bottomrule

\end{tabular}
\caption{Quantative comparison results between publicly available Image-to-Video tools and our proposed AnimateZero. Our proposed AnimateZero demonstrated best performance across multiple metrics or achieved comparable results to the best methods in other metrics. The metrics for the best-performing method are highlighted in \textcolor{red}{red}, while those for the second-best method are highlighted in \textcolor{blue}{blue}.}
\vspace{-1em}
\label{table:compare_all}
\end{table*}

%% file: all_subtex/ablation_study.tex
\begin{figure*}[ht]
  \centering
  \includegraphics[width=1\linewidth]{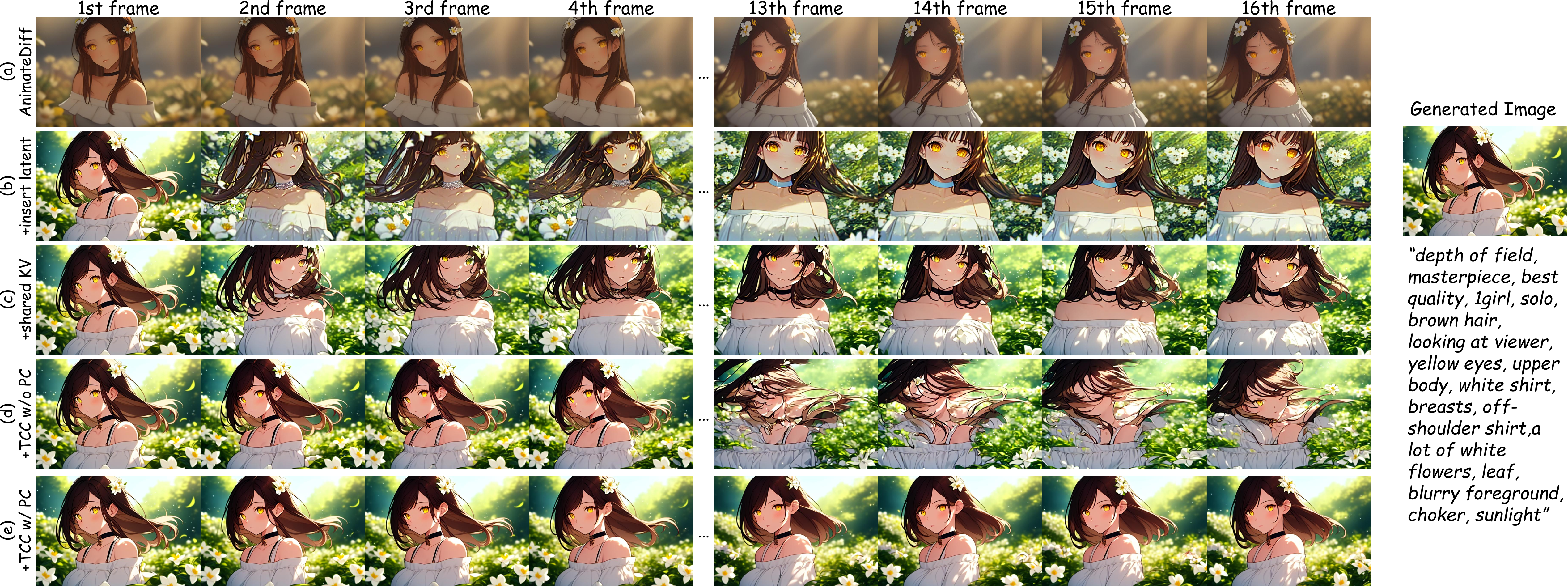}
  \vspace{-0.6cm}
  \caption{Demonstration for ablation study: (a) the video generated by AnimateDiff~\cite{guo2023animatediff}; (b) +inserting intermediate latents responsible for the generation of the given image; (c) +sharing keys and values from the generation of the given image; (d) +temporal consistency control without position correction (TCC w/o PC); (e) +temporal consistency control with position correction (TCC w/ PC). To clearly illustrate the role of each component, we present static frames, while \textbf{\textit{dynamic videos are provided in the supplementary materials.}}
  }
  \vspace{-1em}
\label{fig:ablation} 
\end{figure*}

%% file: all_subtex/limitation.tex
\if \animation 1

\begin{figure}[t]
  \centering
  \begin{tabular}{c@{\hspace{0.1em}}c@{\hspace{0.3em}}c@{\hspace{0.1em}}c}
    {\small AnimateDiff} & {\small AnimateZero} & {\small AnimateDiff} & {\small AnimateZero} \\
      \animategraphics[width=0.24\linewidth]{8}{gif/limitation/1/097_frame_}{0}{15}&
      \animategraphics[width=0.24\linewidth]{8}{gif/limitation/2/099_frame_}{0}{15}&
       \animategraphics[width=0.24\linewidth]{8}{gif/limitation/3/102_frame_}{0}{15}&
       \animategraphics[width=0.24\linewidth]{8}{gif/limitation/4/101_frame_}{0}{15}\\
       \multicolumn{2}{c}{\textit{``1boy, playing football, ...''}} & \multicolumn{2}{c}{\textit{``robot, running, ...''}}
  \end{tabular}
  
  \caption{AnimateZero is limited by the motion prior of AnimateDiff~\cite{guo2023animatediff}, and both perform poorly in complex movements. \textit{Best viewed with Acrobat Reader. Click the video to play the animation clips. \textbf{Static frames are provided in supplementary materials.}}}
  \vspace{-1em}
\label{fig:limitation} 
\end{figure}

\else

\begin{figure}[t]
  \centering
  \begin{tabular}{c@{\hspace{0.1em}}c@{\hspace{0.3em}}c@{\hspace{0.1em}}c}
    {\small AnimateDiff} & {\small AnimateZero} & {\small AnimateDiff} & {\small AnimateZero} \\
      \includegraphics[width=0.24\linewidth]{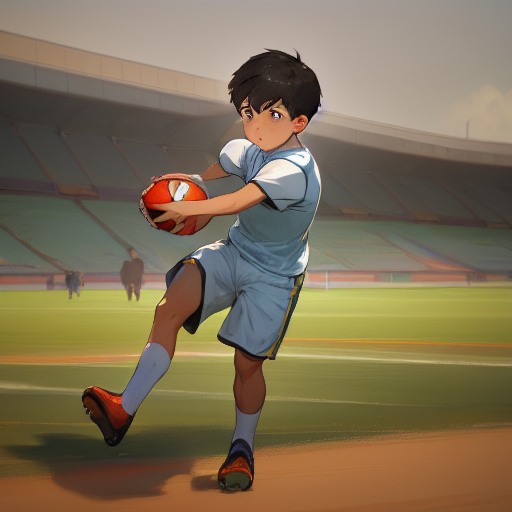} & \includegraphics[width=0.24\linewidth]{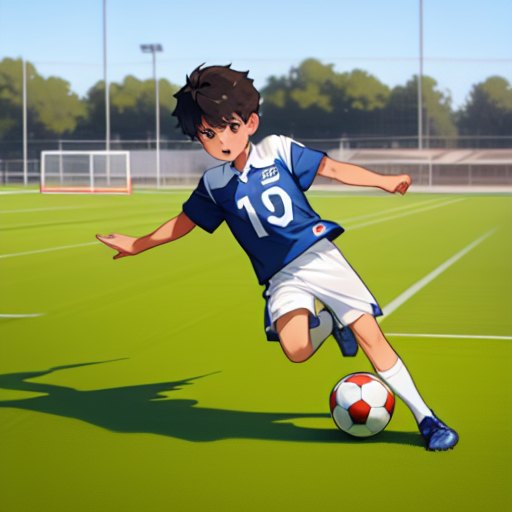}  &
       \includegraphics[width=0.24\linewidth]{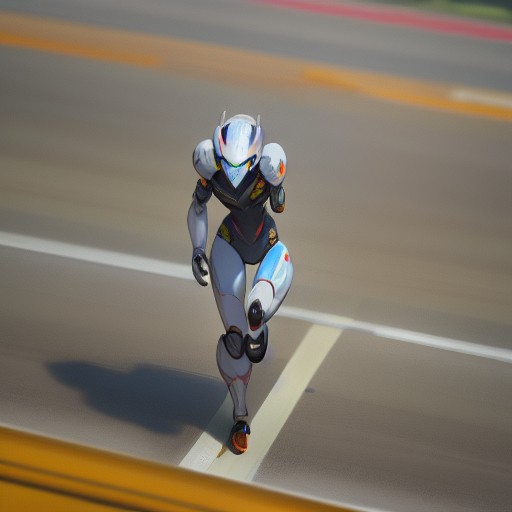} & \includegraphics[width=0.24\linewidth]{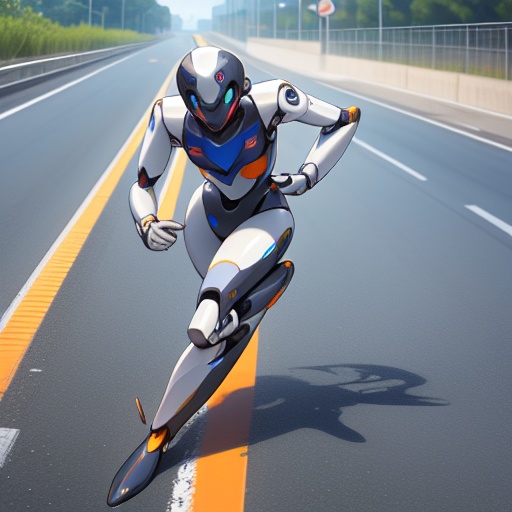} \\
       \multicolumn{2}{c}{\textit{``1boy, playing football, ...''}} & \multicolumn{2}{c}{\textit{``robot, running, ...''}}
  \end{tabular}
  
  \caption{AnimateZero is limited by the motion prior of AnimateDiff~\cite{guo2023animatediff}, and both perform poorly in complex movements. \textit{Best viewed with Acrobat Reader. Click the video to play the animation clips. \textbf{Static frames are provided in supplementary materials.}}}
  \vspace{-1em}
\label{fig:limitation} 
\end{figure}

\fi

%% file: tex/5_conclusion.tex
\section{Conclusions}
 In this paper, we present AnimateZero, which considers video generation as an image animation problem, allowing us to modify the pre-trained video diffusion model to enable more controllability in terms of appearance and motion. To achieve this, for appearance control, we inject the generative latents into the video generation so that we can generate the video concerning the first frame. For motion control, we propose a positional corrected window attention in the motion modules to generate temporally consistent results. Experiments show the advantage of AnimateZero compared with the AnimateDiff and the general image-to-video algorithms. 
AnimateZero is also the first to show that video diffusion models are zero-shot image animators, which not only allows controllable video generation but also opens up possibilities for various applications like animating real images, interactive video creation, and more.

%% file: appendix_tex/sec1.tex
\section{Implementation Details}
\label{sec:append-implemtation}
\subsection{Utilized Personalized T2I Checkpoints}
To thoroughly evaluate the performance of our approach across diverse image domains, we obtain personalized T2I checkpoints featuring various styles from Civitai~\cite{civitai} 
for assessment. Detailed information regarding these checkpoints is provided in Tab.~\ref{tb:t2i_ckpts}.

\begin{table}[ht]
    \centering
    \scriptsize
    \tabcolsep=0.18cm

    \begin{tabular}{l c c c}
        \toprule
        Model Name & Model Type & Image Domain & URL\\
        \hline
        ToonYou & checkpoints & Anime & \url{https://civitai.com/models/30240} \\
        CarDos Anime & checkpoints & Anime & \url{https://civitai.com/models/25399} \\
        Anything V5 & checkpoints & Anime & \url{https://civitai.com/models/9409} \\
        Counterfeit V3.0 & checkpoints & Anime & \url{https://civitai.com/models/4468} \\
        Realistic Vision V5.1 & checkpoints & Realistic & \url{https://civitai.com/models/4201} \\
        Photon & checkpoints & Realistic & \url{https://civitai.com/models/84728} \\
        helloObject & checkpoints & Realistic & \url{https://civitai.com/models/121716} \\
        
        \bottomrule
    \end{tabular}

    \caption{The sources of all personalized T2I checkpoints used in our experiments.}
    \vspace{-0.6cm}
    \label{tb:t2i_ckpts}
\end{table}

\subsection{Position Embedding Correction in Self-Attention}
This subsection presents the detailed calculation process of self-attention with position embedding correction. First, we omit the batch size, height, and width dimensions in our notation, assuming that the input tokens of self-attention is $Z_{in}=\{z_1, z_2, ..., z_f; z_i\in\mathbb{R}^{c\times 1}\}$ where $c$ and $f$ represent the numbers of channels and frames. 
In the \textbf{first} step, we add position embeddings, denoted as $P=\{p_1, p_2, ..., p_f; p_i\in \mathbb{R}^{c\times 1}\}$, to input tokens. We perform pairwise addition of each element in $P$ and $Z_{in}$, constructing a set $\{a_i^{j}; a_i^{j} = z_i + p_j, 1\leq i \leq f, 1\leq j \leq f\}$. 
In the \textbf{second} step, we construct a position-corrected pool for queries, keys and values:
\begin{equation}
\{q_i^j, k_i^j, v_i^j; q_i^j=W_qa_i^j, k_i^j=W_ka_i^j, v_i^j=W_va_i^j, 1\leq i \leq f, 1\leq j \leq f\},    
\end{equation}
where $W_q$, $W_k$ and $W_v$ represent the linear projection weights of queries, keys and values. 
In the \textbf{third} step, we obtain the output $Z_{out}=\{\hat{z}_1, \hat{z}_2, ..., \hat{z}_f; \hat{z}_i\in\mathbb{R}^{c\times 1} \}$ by calculating the proposed window attention:
\begin{equation}
    \hat{z}_i = \Tilde{V}_i\cdot \mathbf{Softmax}((q_i^f)^\top\Tilde{K})_i/\sqrt{c})^{\top}.
    \label{eq:get_out_token}
\end{equation}
The used keys $\Tilde{K}_i$ and values $\Tilde{V}_i$ are all limited in first $i$ frames, that is,
\begin{equation}
    \Tilde{K}_i=\{
    k_{1}^{1},k_{1}^{2},...,k_{1}^{f-i+1},k_{2}^{f-i+2}...,k_{i}^{f}\}, \Tilde{V}_i=\{v_{1}^{1},v_{1}^{2},...,v_{1}^{f-i+1},v_{2}^{f-i+2}...,v_{i}^{f}\},
    \label{eq:kv_list}
\end{equation}
where the token from first frame have been copied for $f-i+1$ times to keep the total numbers of utilized keys and values equal to $f$. One detail worth noting is that the superscript of the query token from the $i$-th frame in Eq.~\ref{eq:get_out_token} should be consistent with the superscript of the corresponding key and value tokens from $i$-th frame in Eq.~\ref{eq:kv_list}.

%% file: appendix_tex/sec2.tex
\section{Effect of Time-Travel Sampling Strategy}
\label{sec:append_tt}

\input{append-fig-tex/time-travel-fig}

As proposed in \cite{wang2023zeroshot, yu2023freedom}, the time-travel sampling strategy has the ability to improve the visual quality of sampled results. In our experiments, this strategy generates smoother video according to the chosen hyperparameters. Assuming that the intermediate video latent code at $t$-th timestep is $\mathbf{Z}_t=\{\mathbf{z}_{t}^{1}, \mathbf{z}_{t}^{2}, ..., \mathbf{z}_{t}^{f}\}$, where $\mathbf{z}_{t}^{i}$ is the intermediate latent code for $i$-th frame. For one denoising step, the calculation formula is 
\begin{equation}
    \mathbf{Z}_{t-1}=Denoiser(\mathbf{Z}_{t}, \mathbf{t}_{prompt}, t), 
    \label{eq:denoising}
\end{equation}
in which $Denoiser(\cdot)$ represents the denoising process and $\mathbf{t}_{prompt}$ is the given prompt text. The time-travel sampling strategy iteratively performs denoising operation in Eq.~\ref{eq:denoising} and diffusion operation~\cite{ho2020denoising} in: 
\begin{equation}
    \mathbf{Z}_{t}=\sqrt{\alpha_t}\mathbf{Z}_{t-1}+\sqrt{1-\alpha_t}\mathbf{N}, \mathbf{N}=\{\epsilon_1, \epsilon_2, ..., \epsilon_f; \epsilon_i\sim \mathcal{N}(\mathbf{0}, \mathbf{I})\}. 
\end{equation}
This approach aims to achieve more harmonious and natural generation results. In the context of video generation tasks, we observed that it has the ability to make the final video smoother.
The number of iterations in each timestep determines the degree of smoothness in the final results, and we set the default number as $5$. According to the findings in \cite{yu2023freedom}, it is unnecessary to apply this strategy at every timestep. Instead, we use the time-travel sampling strategy only between the $10$-th timestep and the $20$-th timestep.
We show the comparison results before and after using time-travel sampling strategy in Fig.~\ref{fig:time-travel}.

%% file: append-fig-tex/time-travel-fig.tex

\if \animation 1
\begin{figure*}[t]
  \centering
  \begin{tabular}{c@{\hspace{0.3em}}c@{\hspace{0.3em}}c@{\hspace{0.8em}}c@{\hspace{0.3em}}c@{\hspace{0.3em}}c}
  Generated Images & w/o time-travel & w/ time-travel & Generated Images & w/o time-travel & w/ time-travel \\
    \includegraphics[width=0.16\linewidth]{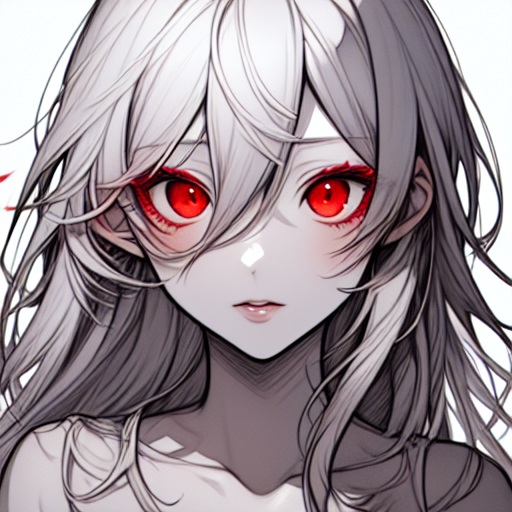}&
    \animategraphics[width=0.16\linewidth]{8}{gif/time-travel/080/080_frame_}{0}{15}&
     \animategraphics[width=0.16\linewidth]{8}{gif/time-travel/081/081_frame_}{0}{15}&
     \includegraphics[width=0.16\linewidth]{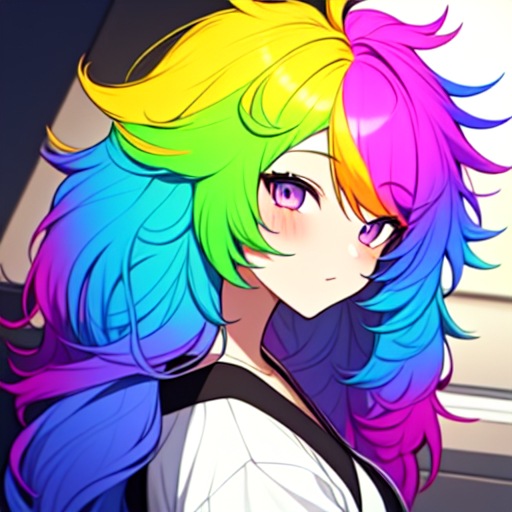}&
      \animategraphics[width=0.16\linewidth]{8}{gif/time-travel/084/084_frame_}{0}{15}&
       \animategraphics[width=0.16\linewidth]{8}{gif/time-travel/085/085_frame_}{0}{15}
       \\
       \multicolumn{3}{c}{\textit{``1girl, red eyes, silver hair, shiny skin, ...''}} & \multicolumn{3}{c}{\textit{``1girl with rainbow hair, really wild hair, ...''}}
    \end{tabular}
\vspace{-0.5em}
  \caption{Demonstrate the role of time-travel sampling strategy~\cite{wang2023zeroshot, yu2023freedom}. The time-travel sampling strategy can produce smoother and more natural results. However, it should be emphasized that in most cases, AnimateZero can already obtain satisfactory results. The time-travel sampling strategy is only used in certain T2I models (such as Anything V5) or certain complex textures (such as hair). \textit{Best viewed with Acrobat Reader. Click the video to play the animation clips. \textbf{Static frames are provided in Sec.~\ref{subsec:static}.}}}
  \vspace{-1em}
\label{fig:time-travel} 
\end{figure*}

\else
\begin{figure*}[t]
  \centering
  \begin{tabular}{c@{\hspace{0.3em}}c@{\hspace{0.3em}}c@{\hspace{0.8em}}c@{\hspace{0.3em}}c@{\hspace{0.3em}}c}
  Generated Images & w/o time-travel & w/ time-travel & Generated Images & w/o time-travel & w/ time-travel \\
    \includegraphics[width=0.16\linewidth]{gif/time-travel/080.jpg}&
    \includegraphics[width=0.16\linewidth]{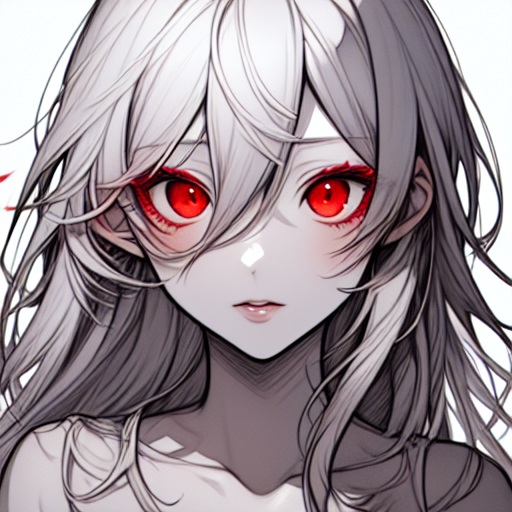}&
     \includegraphics[width=0.16\linewidth]{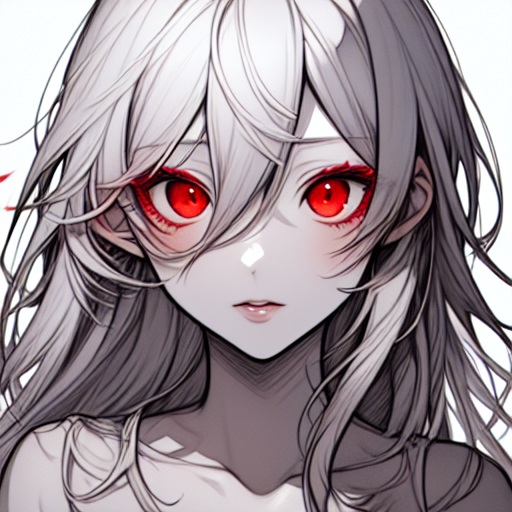}&
     \includegraphics[width=0.16\linewidth]{gif/time-travel/084.jpg}&
      \includegraphics[width=0.16\linewidth]{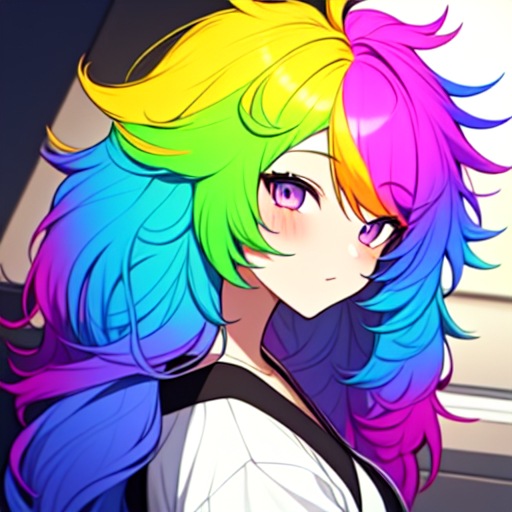}&
       \includegraphics[width=0.16\linewidth]{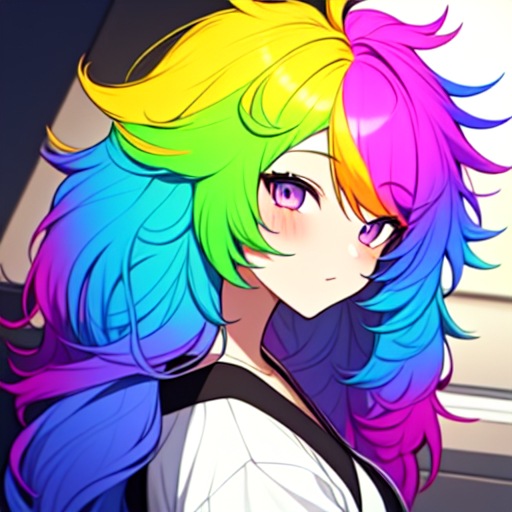}
       \\
       \multicolumn{3}{c}{\textit{``1girl, red eyes, silver hair, shiny skin, ...''}} & \multicolumn{3}{c}{\textit{``1girl with rainbow hair, really wild hair, ...''}}
    \end{tabular}
\vspace{-0.5em}
  \caption{Demonstrate the role of time-travel sampling strategy~\cite{wang2023zeroshot, yu2023freedom}. The time-travel sampling strategy can produce smoother and more natural results. However, it should be emphasized that in most cases, AnimateZero can already obtain satisfactory results. The time-travel sampling strategy is only used in certain T2I models (such as Anything V5) or certain complex textures (such as hair). \textit{Best viewed with Acrobat Reader. Click the video to play the animation clips. \textbf{Static frames are provided in Sec.~\ref{subsec:static}.}}}
  \vspace{-1em}
\label{fig:time-travel} 
\end{figure*}

\fi

%% file: appendix_tex/sec3.tex
\section{Applications}
\label{sec:append_applications}
\subsection{Improved Video Editing Compared to AnimateDiff~\cite{guo2023animatediff}}
The temporal consistency of videos generated by AnimateDiff is notable, and a common use of AnimateDiff is to assist ControlNet~\cite{zhang2023adding} in video editing, aiming to achieve smooth editing results. The specific approach involves inputting feature information for each frame of the original video (such as extracted depth maps, edge maps, \etc.) to ControlNet, thereby controlling each frame of the video generated by AnimateDiff.
The primary challenge encountered in this video editing process is the inherent domain gap issue of AnimateDiff. This issue significantly degrades the subjective quality of the edited video, and the alignment degree between the text and the generated video is also substantially reduced.
As demonstrated in the experimental section of the main paper, AnimateZero exhibits a significant advantage in maintaining the T2I domain compared to AnimateDiff. Therefore, we attempted to use AnimateZero to assist ControlNet in video editing. The results of the editing process showed a noticeable improvement in subjective quality and text-video matching degree compared to AnimateDiff. Additionally, AnimateZero still ensures that the generated video remains smooth and good temporal consistency. \textbf{\textit{We showcase some results of AnimateZero performing video editing in our project page.}}

\subsection{Frame Interpolation and Looped Video Generation}

\begin{figure*}[t]
  \centering
  \includegraphics[width=1.0\linewidth]{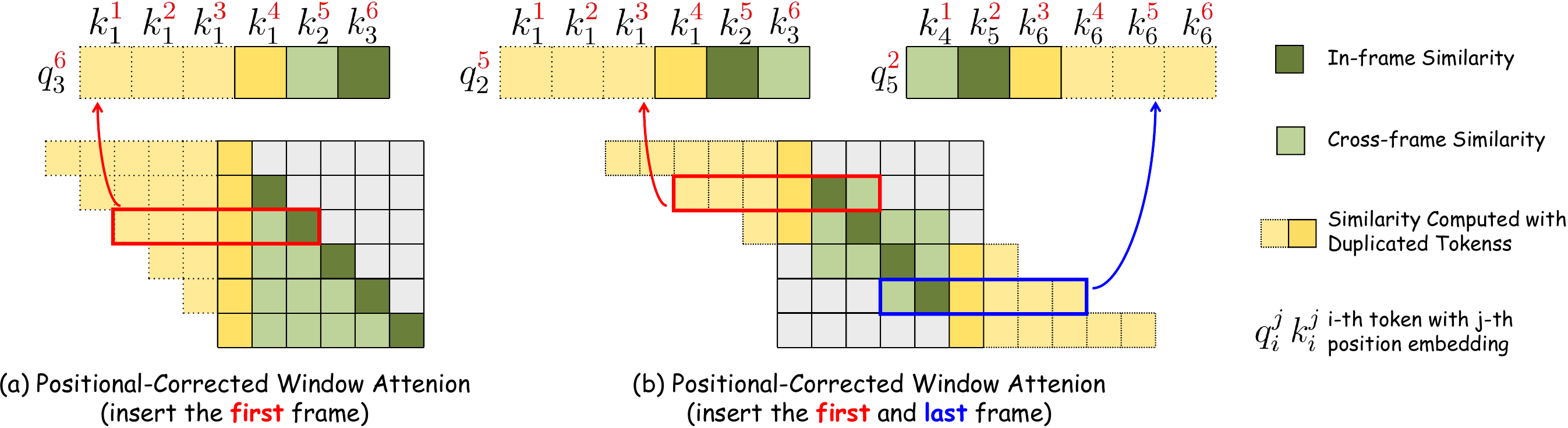}
  \vspace{-0.6cm}
  \caption{ Demonstrate the difference between (a) insertion of the \textcolor{red}{first} frame and (b) insertion of both \textcolor{red}{first} and \textcolor{blue}{last} frames. The technique illustrated in (b) is the basis for achieving applications like frame interpolation and looped video generation. }
  \vspace{-0.6cm}
\label{fig:insert_two} 
\end{figure*}

AnimateZero attempts to insert the first frame into the generated video to achieve image animation from the generated image. An extended idea is whether similar techniques can be used to insert \textbf{multiple} frames. In Fig.~\ref{fig:insert_two}, we propose an extension to the original position-corrected window attention used by AnimateZero. This extension allows for the simultaneous insertion of both the first and last frames. The key modification involves simultaneously emphasizing the tokens corresponding to both the first and last frames, ensuring that the final generated video's first and last frames match the given images.
This technique has the potential application for frame interpolation, allowing interpolation between any two generated images. Additionally, when the first and last frames are the same, it can be considered as achieving looped video generation. \textbf{\textit{Relevant results are showcased in our project page.}}

\subsection{Real Image Animation}
AnimateZero has been demonstrated to perform image animation on generated images, but it also has the potential to handle image animation on real images. The main difference between real and generated images is the absence of readily available intermediate latents. However, we can obtain pseudo intermediate latents through methods like DDIM Inversion~\cite{ddim} or by directly diffusing the clean latents~\cite{ho2020denoising}, enabling image generation on real images.
Nevertheless, for real images, the issue of domain gap is challenging to avoid. This is influenced not only by the style of the real image but also factors such as its resolution and whether it has been degraded by some degradation operators. \textbf{\textit{We showcase some results of AnimateZero performing real image animation in our project page.}}

%% file: appendix_tex/sec4.tex
\section{More Visual Results}
\label{sec:append_moreresults}
\label{subsec:static}
To facilitate the accessibility of visual results for non-Adobe users, we have included \textit{\textbf{static frames}} of videos in main paper (Fig.~\ref{fig:teaser}, ~\ref{fig:ADvsAZ}, ~\ref{fig:ADvsOthers}, ~\ref{fig:limitation}) and supplementary materials (Fig.~\ref{fig:time-travel}) in Fig.~\ref{fig:static-fig1-1}, ~\ref{fig:static-fig1-2}, ~\ref{fig:static-fig4}, ~\ref{fig:static-fig5-1}, ~\ref{fig:static-fig5-2}, ~\ref{fig:static-fig7}, ~\ref{fig:static-figa}, . 
We also provide an \textit{\textbf{HTML file}} with many generated video examples, which we highly recommend readers to check out. In addition, we have also attached the \textit{\textbf{source files}} of all videos in the paper and HTML file for the convenience of readers.
\input{append-fig-tex/static-fig1}
\input{append-fig-tex/static-fig4}
\input{append-fig-tex/static-fig5}
\input{append-fig-tex/static-fig7}
\input{append-fig-tex/static-figa}

%% file: append-fig-tex/static-fig1.tex
\begin{figure*}[ht]
  \centering
  \begin{tabular}{c@{\hspace{0.em}}c@{\hspace{0.em}}c@{\hspace{0.em}}c@{\hspace{0.em}}c@{\hspace{0.em}}c@{\hspace{0.em}}c@{\hspace{0.em}}c} 
    \includegraphics[width=0.125\linewidth]{gif/teaser/1/082_frame_0.jpg}&
    \includegraphics[width=0.125\linewidth]{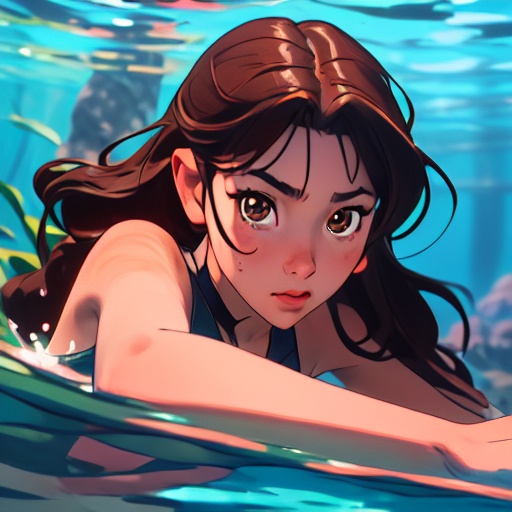}&
    \includegraphics[width=0.125\linewidth]{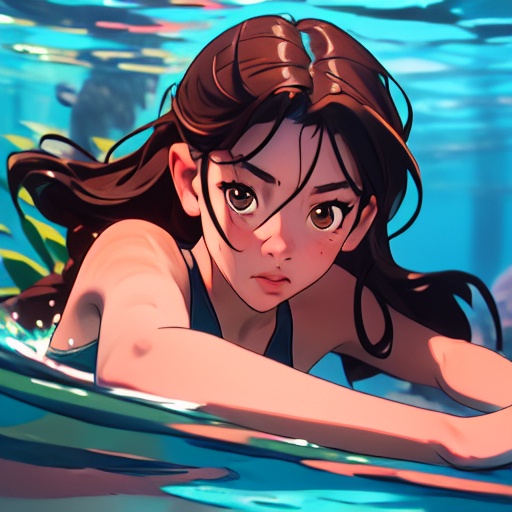}&
    \includegraphics[width=0.125\linewidth]{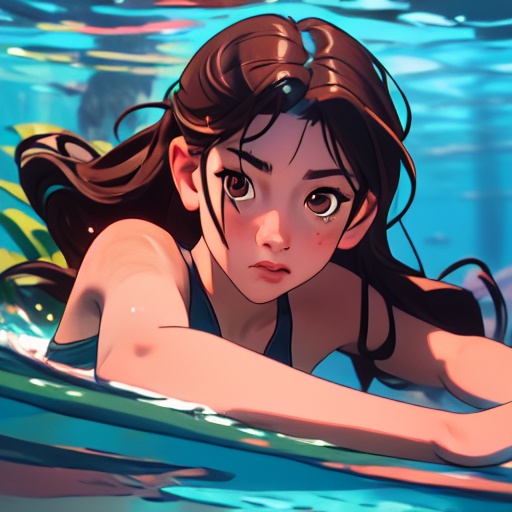}&
    \includegraphics[width=0.125\linewidth]{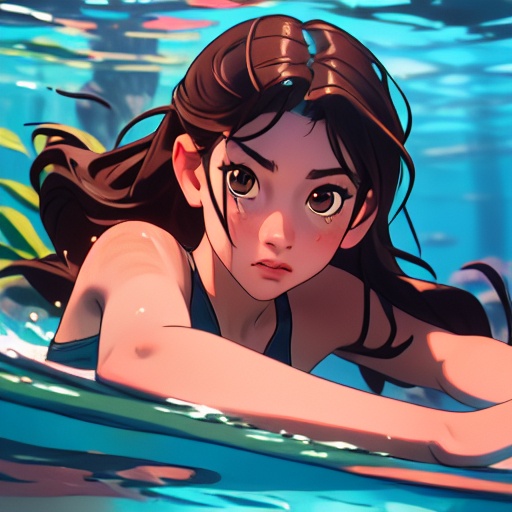}&
    \includegraphics[width=0.125\linewidth]{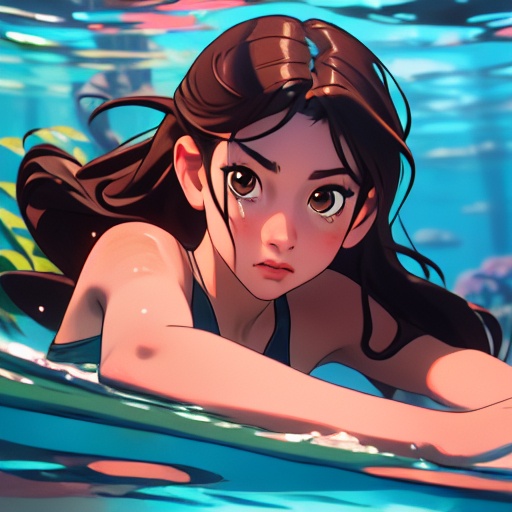}&
    \includegraphics[width=0.125\linewidth]{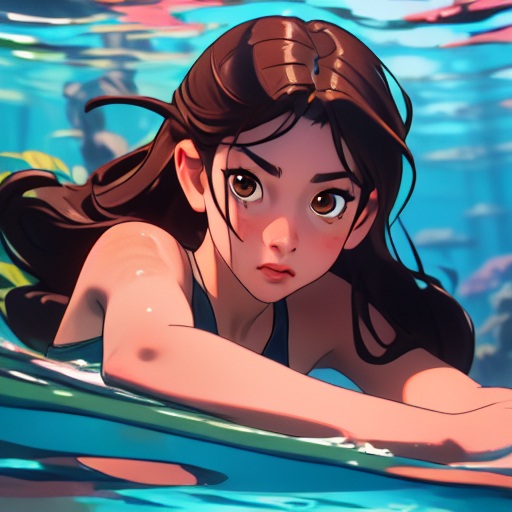}&
    \includegraphics[width=0.125\linewidth]{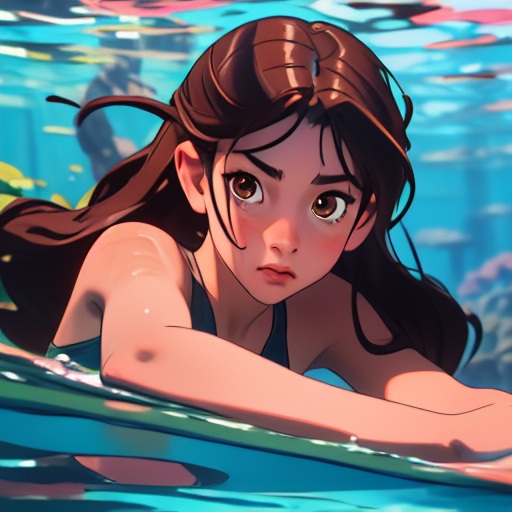}
    \\
    \multicolumn{8}{c}{\textit{``1girl, underwater, swimsuit, air bubble, looking at viewer, swimming, dappled sunlight, ...''}}\vspace{0.5em}
    \\
    \includegraphics[width=0.125\linewidth]{gif/teaser/3/012_frame_0.jpg}&
    \includegraphics[width=0.125\linewidth]{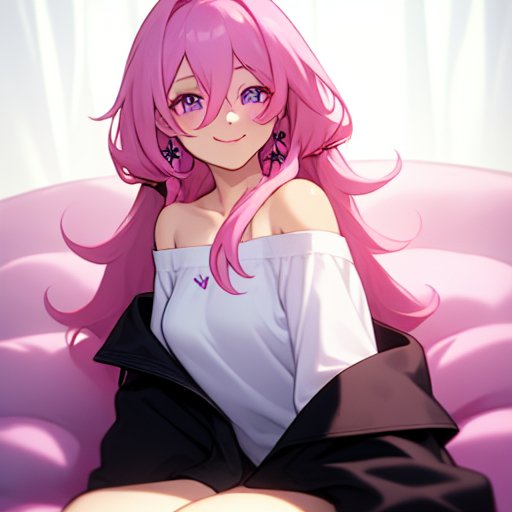}&
    \includegraphics[width=0.125\linewidth]{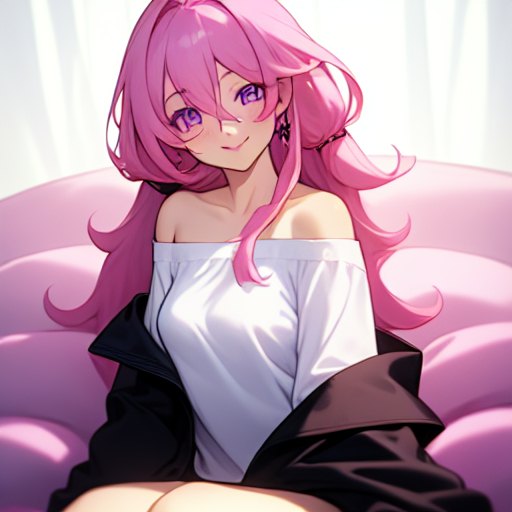}&
    \includegraphics[width=0.125\linewidth]{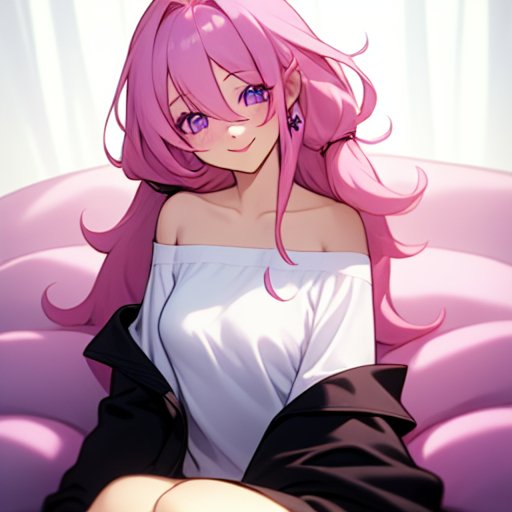}&
    \includegraphics[width=0.125\linewidth]{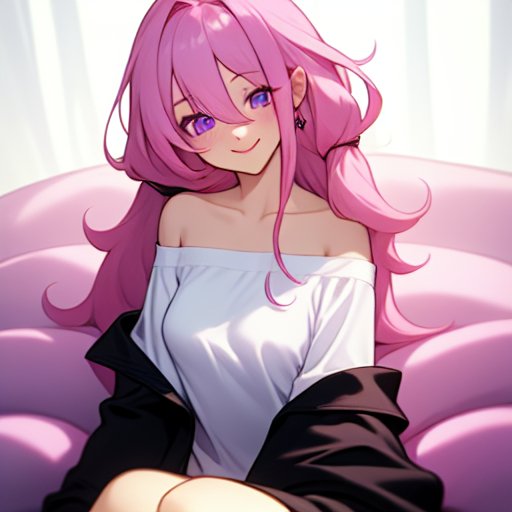}&
    \includegraphics[width=0.125\linewidth]{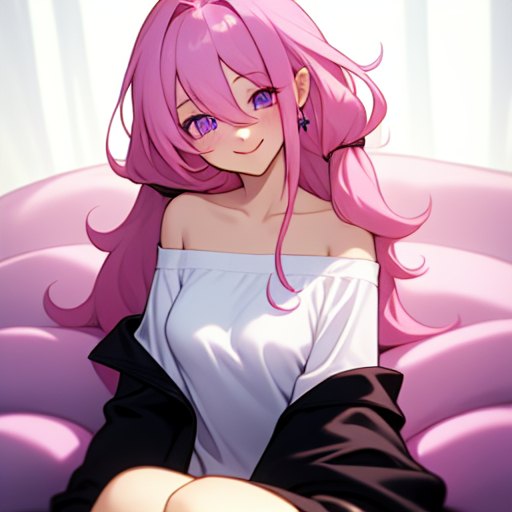}&
    \includegraphics[width=0.125\linewidth]{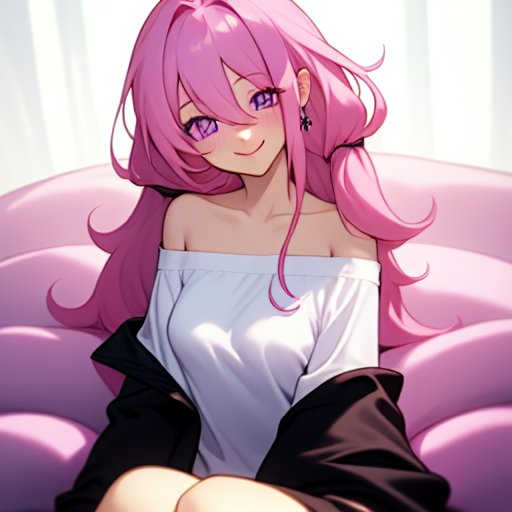}&
    \includegraphics[width=0.125\linewidth]{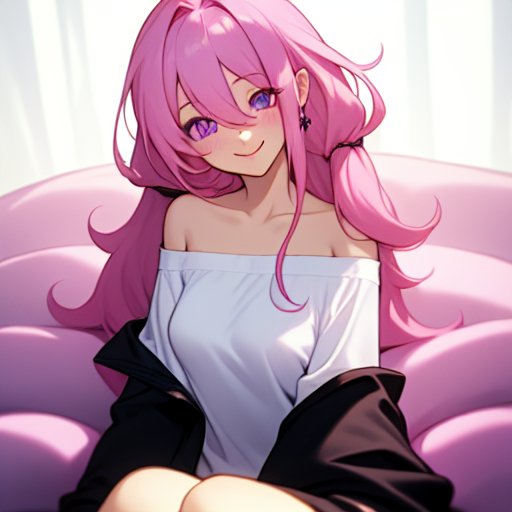}
    \\
    \multicolumn{8}{c}{\textit{``1girl, black jacket, long sleeves, pink hair, hair between eyes, sitting, ...''}}\vspace{0.5em}
    \\
    \includegraphics[width=0.125\linewidth]{gif/teaser/4/020_frame_0.jpg}&
    \includegraphics[width=0.125\linewidth]{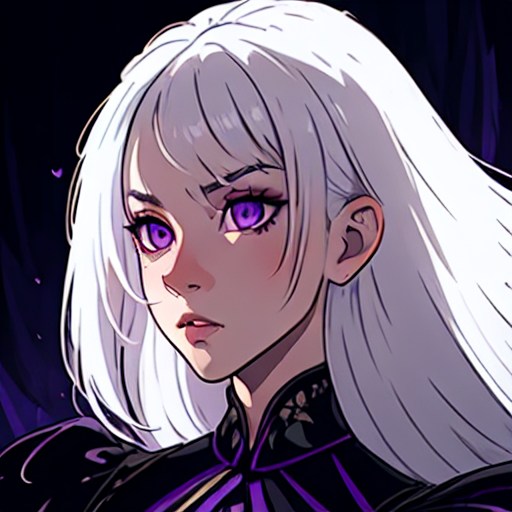}&
    \includegraphics[width=0.125\linewidth]{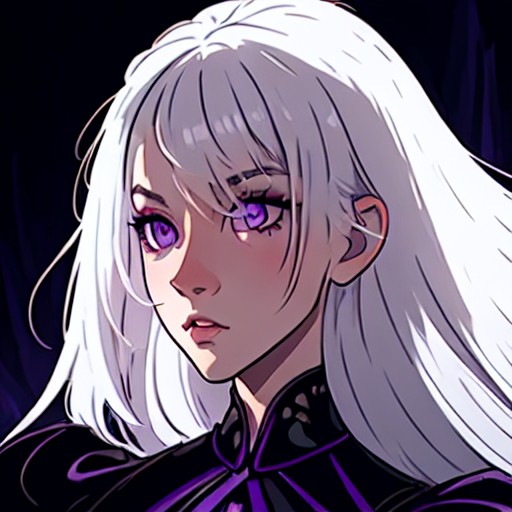}&
    \includegraphics[width=0.125\linewidth]{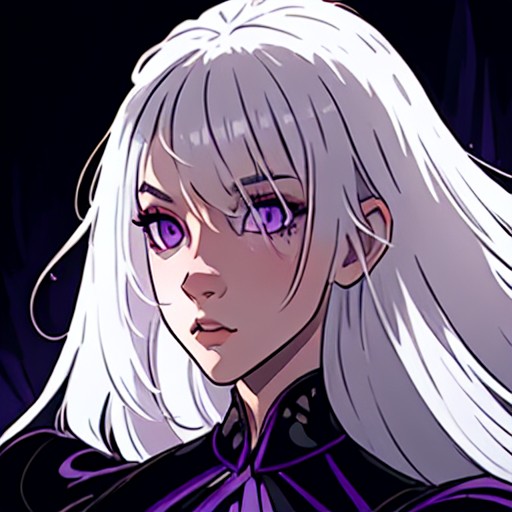}&
    \includegraphics[width=0.125\linewidth]{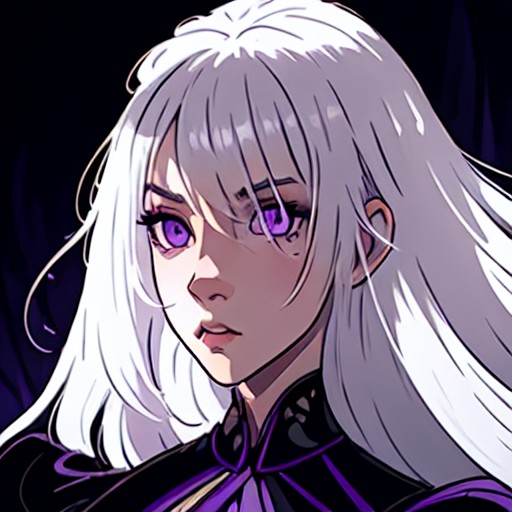}&
    \includegraphics[width=0.125\linewidth]{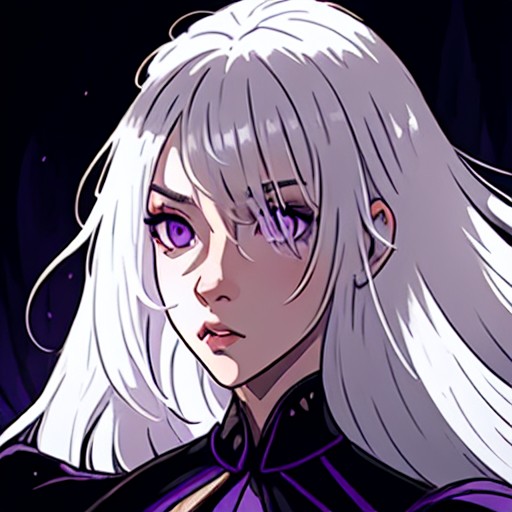}&
    \includegraphics[width=0.125\linewidth]{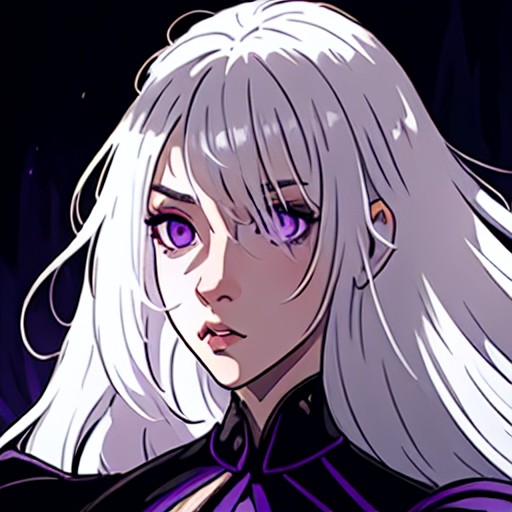}&
    \includegraphics[width=0.125\linewidth]{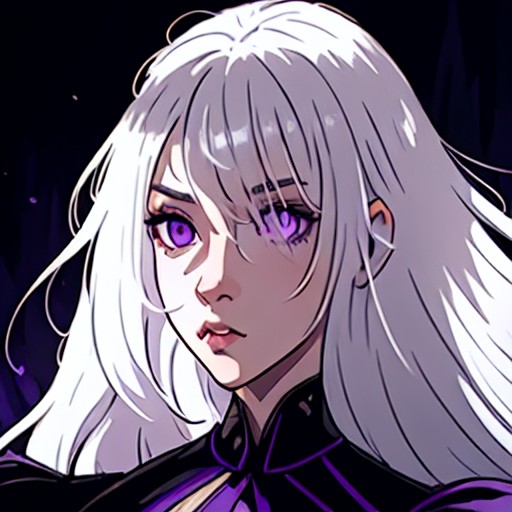}
    \\
    \multicolumn{8}{c}{\textit{``dark fantasy, purple eyes, cinematic light, white hair, sharp face, hair between eyes, ...''}}\vspace{0.5em}
    \\
    \includegraphics[width=0.125\linewidth]{gif/teaser/5/034_frame_0.jpg}&
    \includegraphics[width=0.125\linewidth]{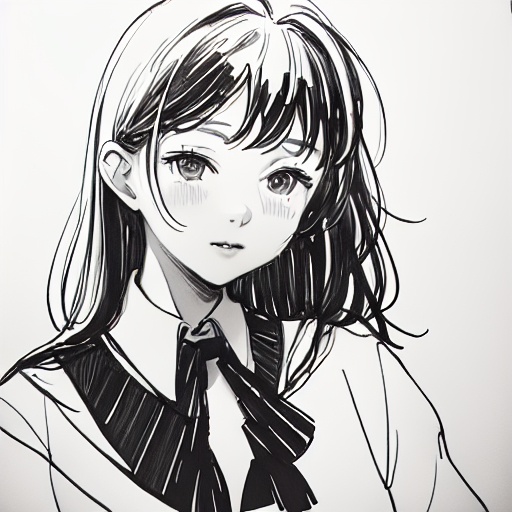}&
    \includegraphics[width=0.125\linewidth]{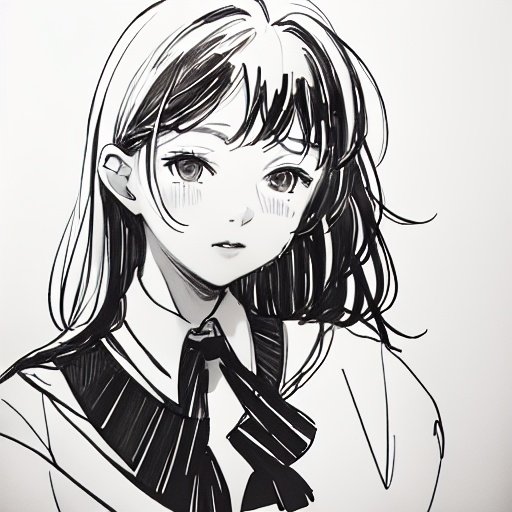}&
    \includegraphics[width=0.125\linewidth]{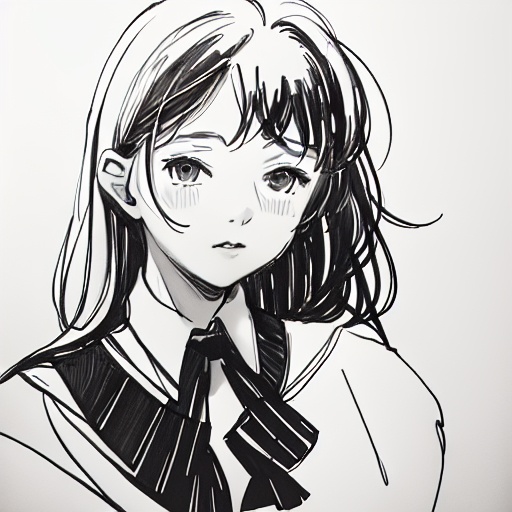}&
    \includegraphics[width=0.125\linewidth]{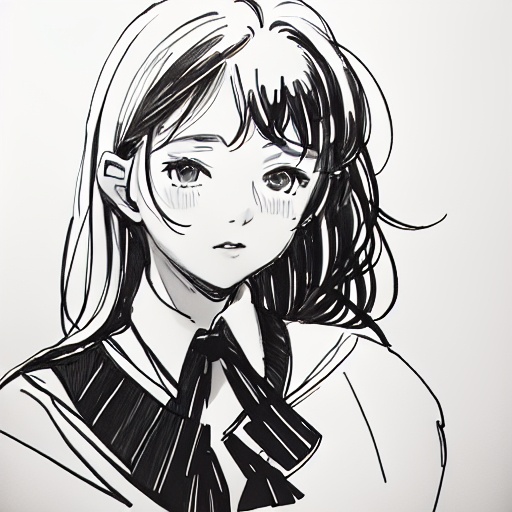}&
    \includegraphics[width=0.125\linewidth]{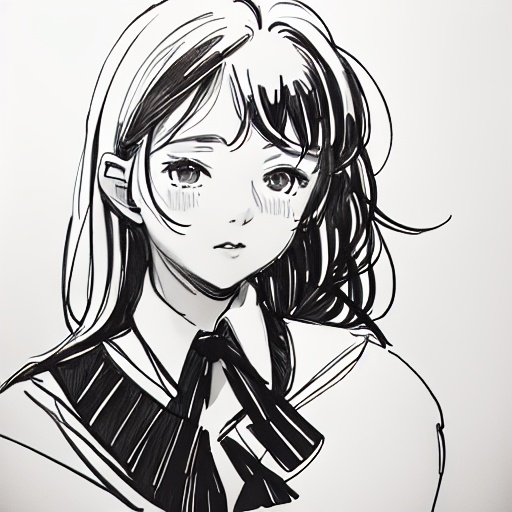}&
    \includegraphics[width=0.125\linewidth]{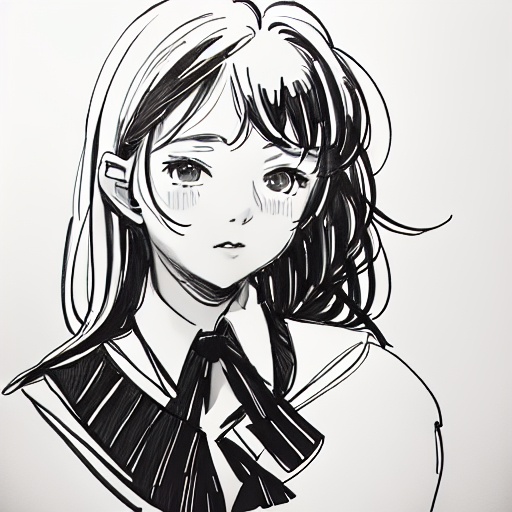}&
    \includegraphics[width=0.125\linewidth]{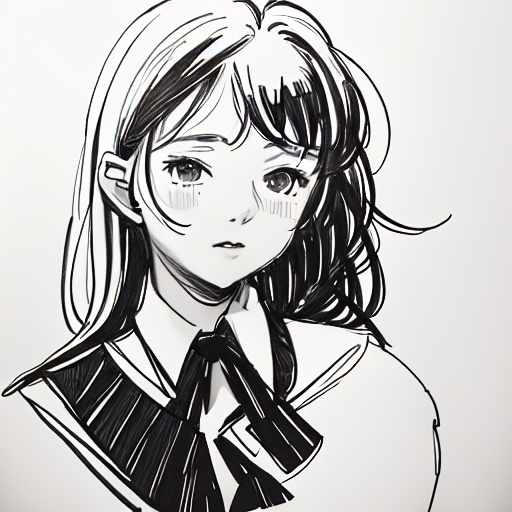}
    \\
    \multicolumn{8}{c}{\textit{``school uniform, JK, sketch, long hair, clam down, looking at the camera, ...''}}\vspace{0.5em}
    \\
    \includegraphics[width=0.125\linewidth]{gif/teaser/6/051_frame_0.jpg}&
    \includegraphics[width=0.125\linewidth]{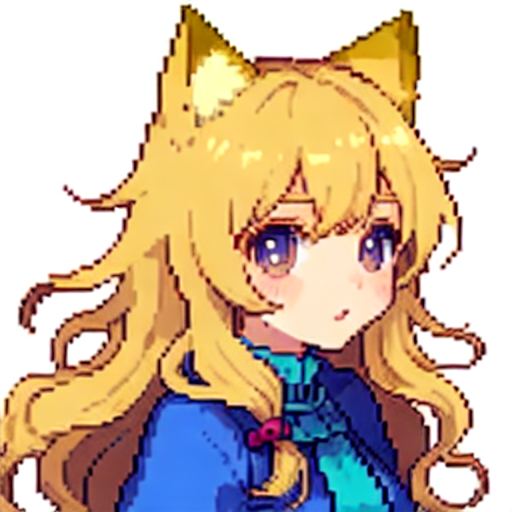}&
    \includegraphics[width=0.125\linewidth]{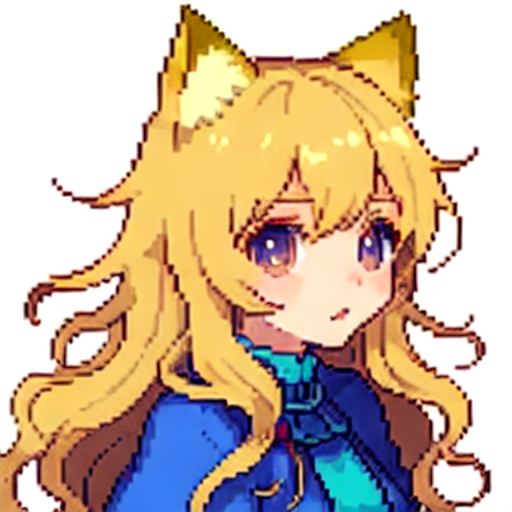}&
    \includegraphics[width=0.125\linewidth]{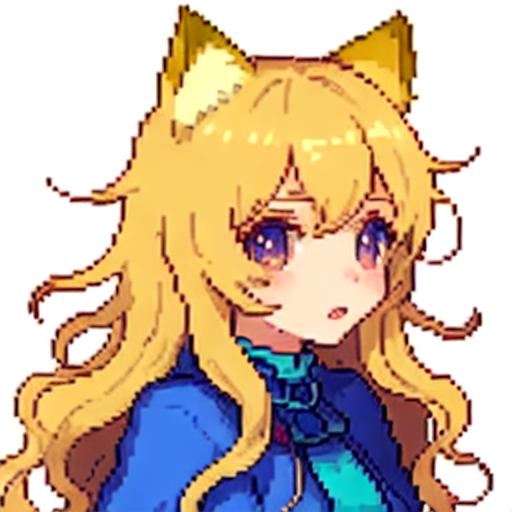}&
    \includegraphics[width=0.125\linewidth]{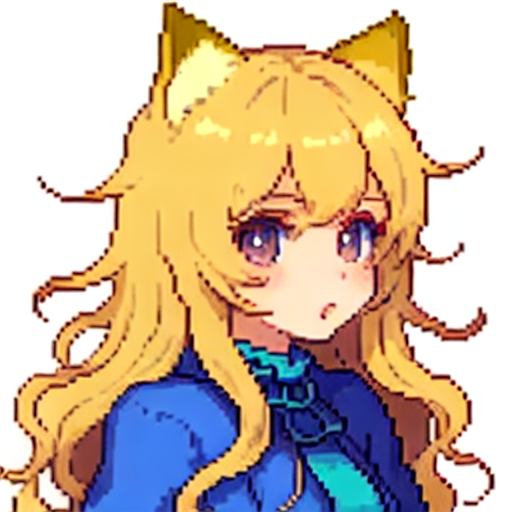}&
    \includegraphics[width=0.125\linewidth]{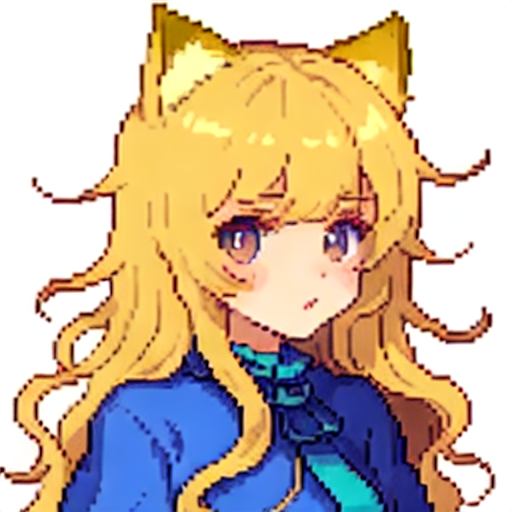}&
    \includegraphics[width=0.125\linewidth]{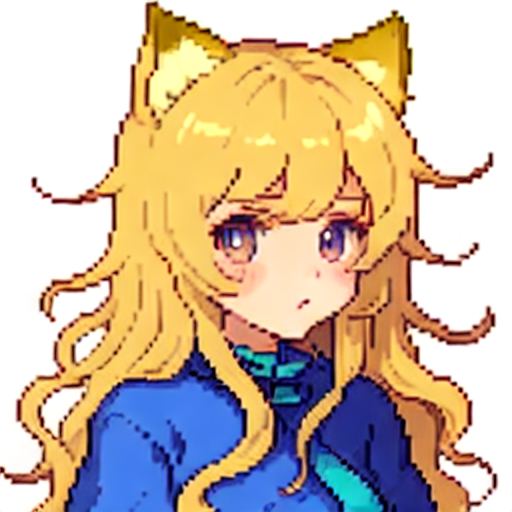}&
    \includegraphics[width=0.125\linewidth]{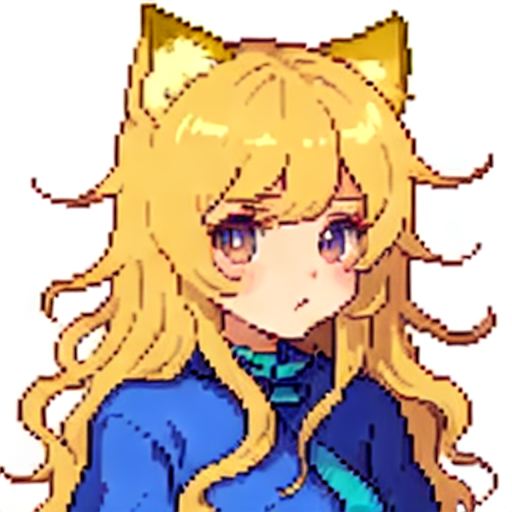}
    \\
    \multicolumn{8}{c}{\textit{``pixel art, cat ears, blonde hair,  wavy hair, portrait of cute girl, ...''}}
    \end{tabular}
\vspace{-0.5em}
  \caption{Static frames sequences in Fig.~\ref{fig:teaser} (part 1).}
  \vspace{-1em}
\label{fig:static-fig1-1} 
\end{figure*}

\begin{figure*}[ht]
  \centering
  \begin{tabular}{c@{\hspace{0.em}}c@{\hspace{0.em}}c@{\hspace{0.em}}c@{\hspace{0.em}}c@{\hspace{0.em}}c@{\hspace{0.em}}c@{\hspace{0.em}}c} 
    \includegraphics[width=0.125\linewidth]{gif/teaser/10/141_frame_0.jpg}&
    \includegraphics[width=0.125\linewidth]{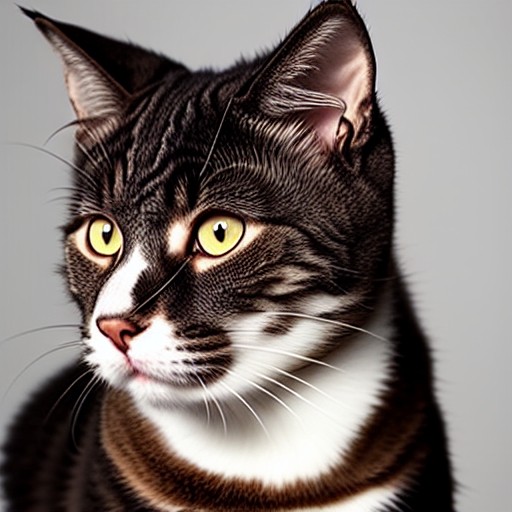}&
    \includegraphics[width=0.125\linewidth]{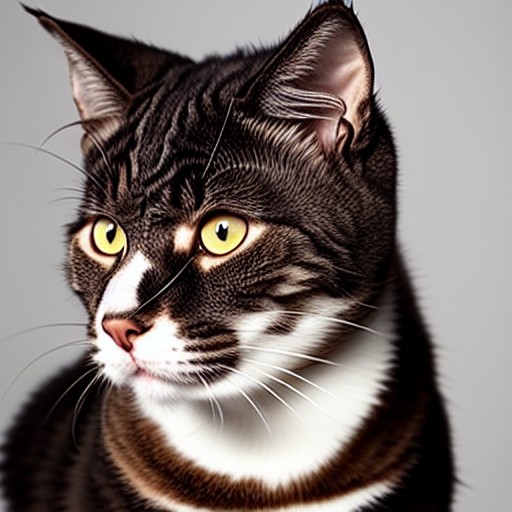}&
    \includegraphics[width=0.125\linewidth]{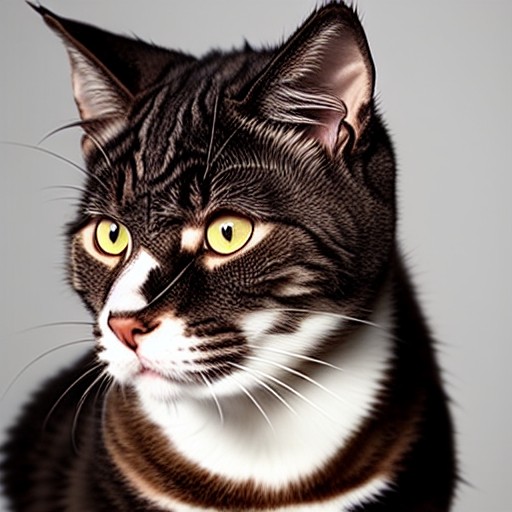}&
    \includegraphics[width=0.125\linewidth]{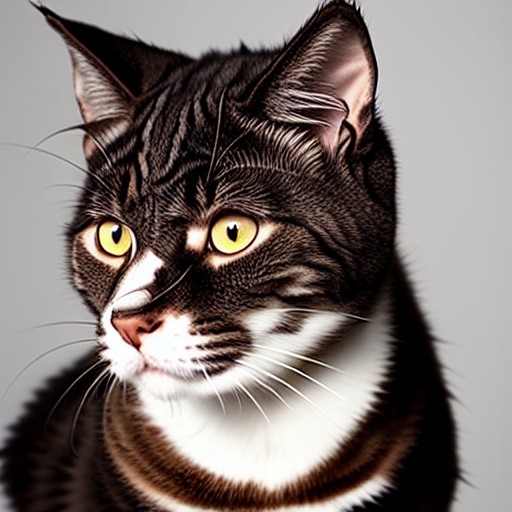}&
    \includegraphics[width=0.125\linewidth]{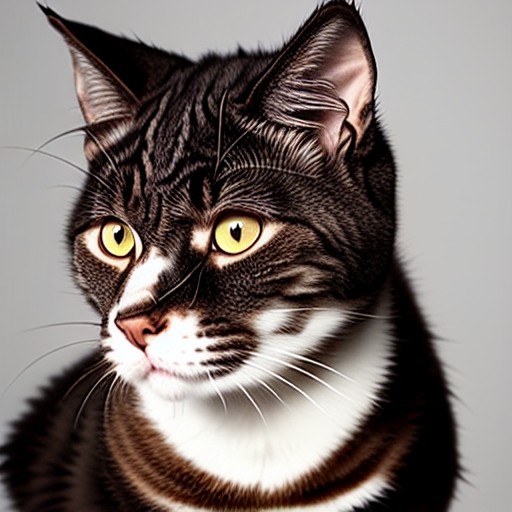}&
    \includegraphics[width=0.125\linewidth]{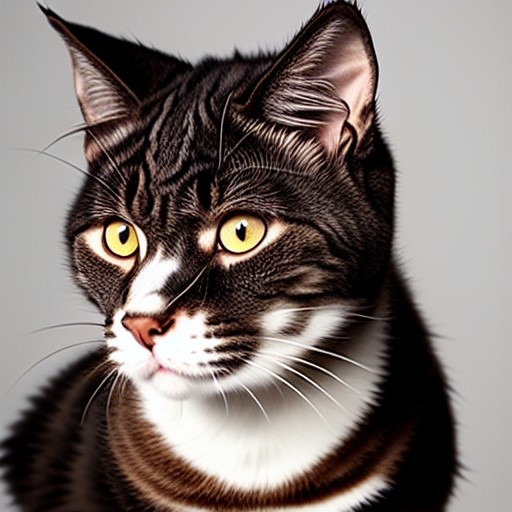}&
    \includegraphics[width=0.125\linewidth]{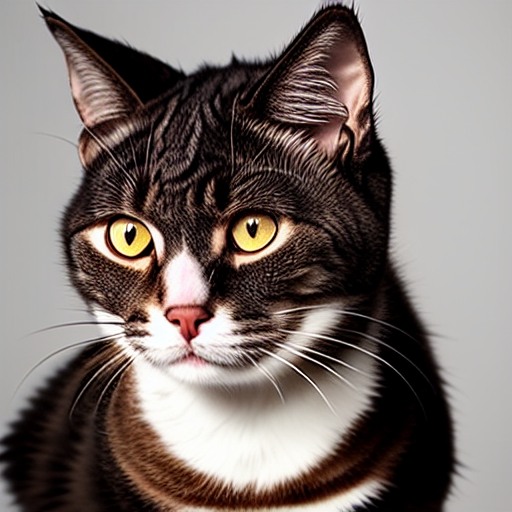}
    \\
    \multicolumn{8}{c}{\textit{``a cat head, look to one side''}}\vspace{0.5em}
    \\
    \includegraphics[width=0.125\linewidth]{gif/teaser/7/100-0.jpg}&
    \includegraphics[width=0.125\linewidth]{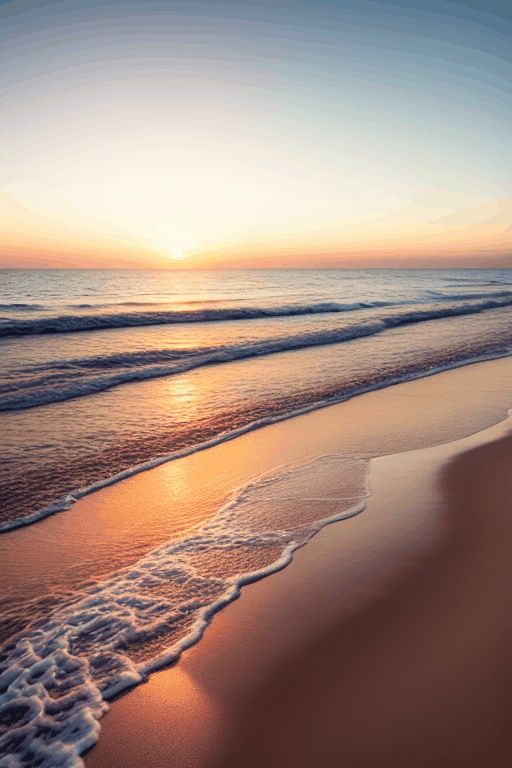}&
    \includegraphics[width=0.125\linewidth]{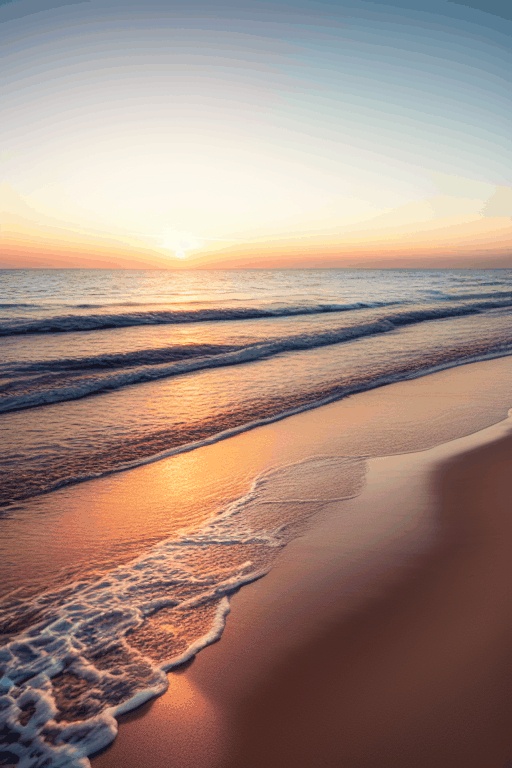}&
    \includegraphics[width=0.125\linewidth]{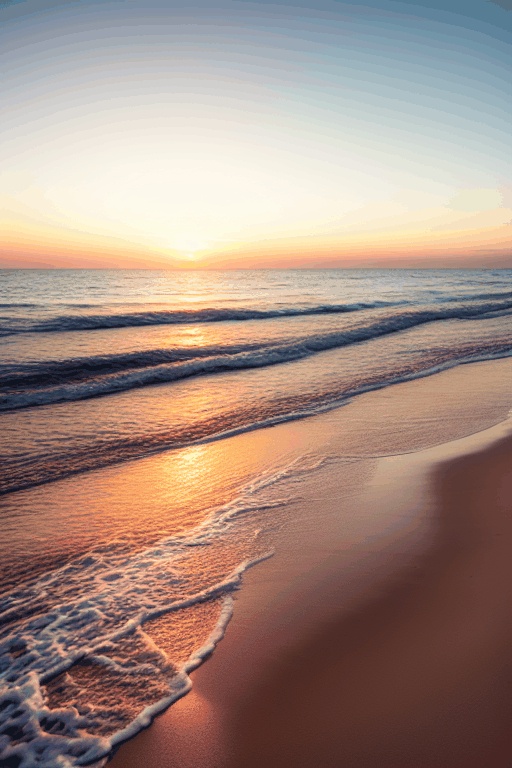}&
    \includegraphics[width=0.125\linewidth]{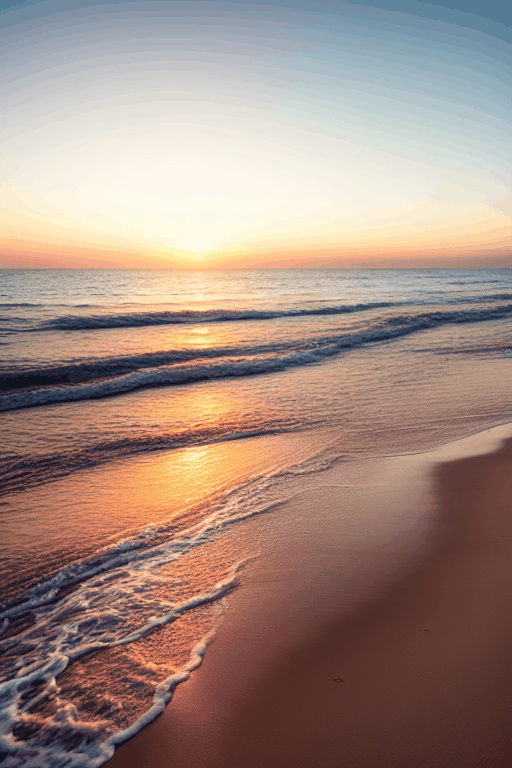}&
    \includegraphics[width=0.125\linewidth]{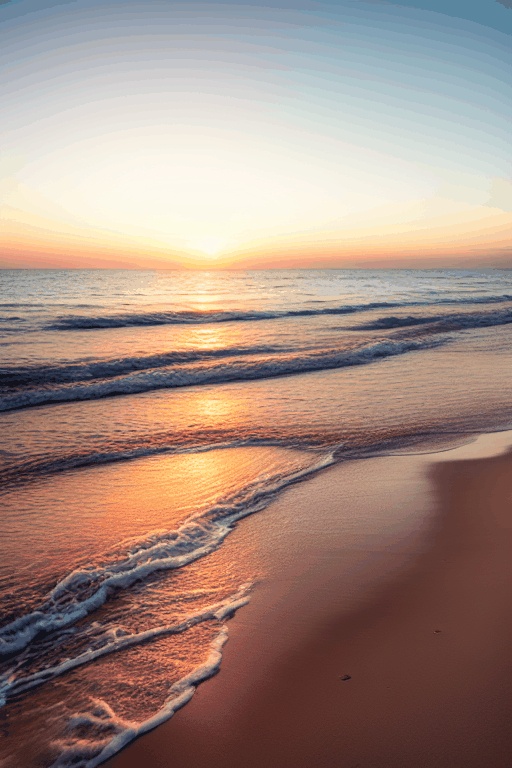}&
    \includegraphics[width=0.125\linewidth]{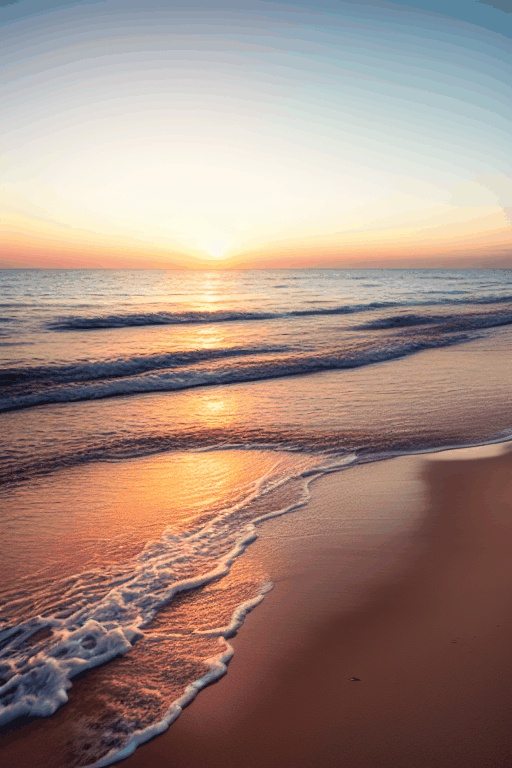}&
    \includegraphics[width=0.125\linewidth]{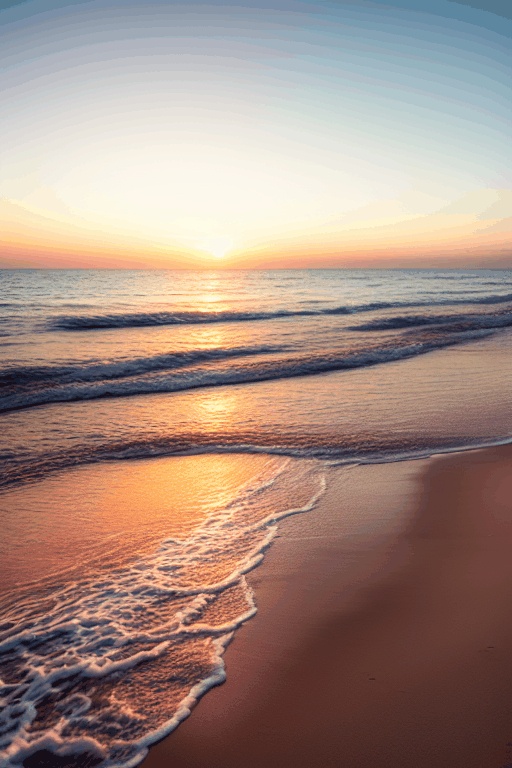}
    \\
    \multicolumn{8}{c}{\textit{``waves hit the beach''}}\vspace{0.5em}
    \\
    \includegraphics[width=0.125\linewidth]{gif/teaser/8/580-0.jpg}&
    \includegraphics[width=0.125\linewidth]{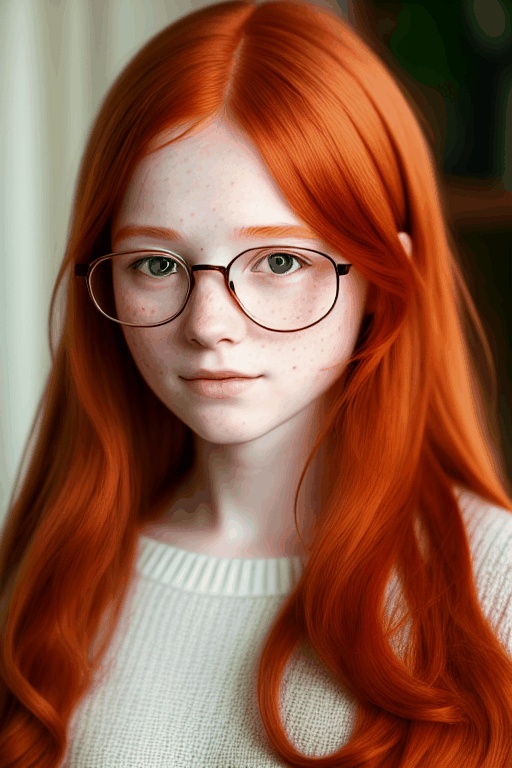}&
    \includegraphics[width=0.125\linewidth]{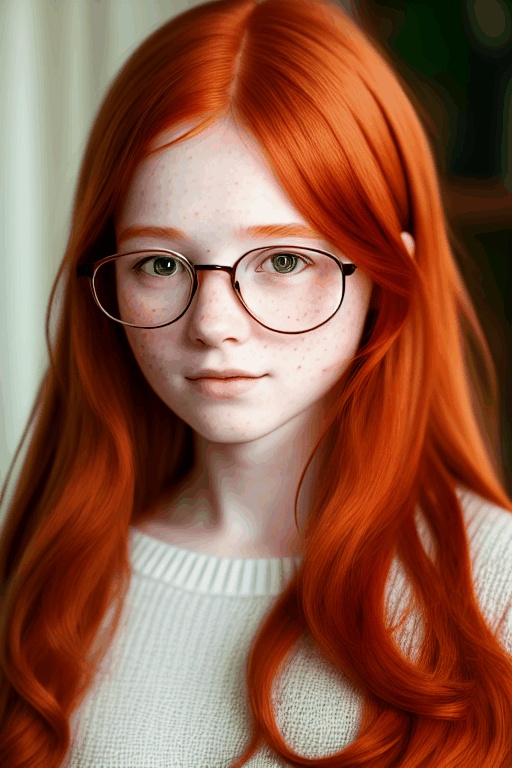}&
    \includegraphics[width=0.125\linewidth]{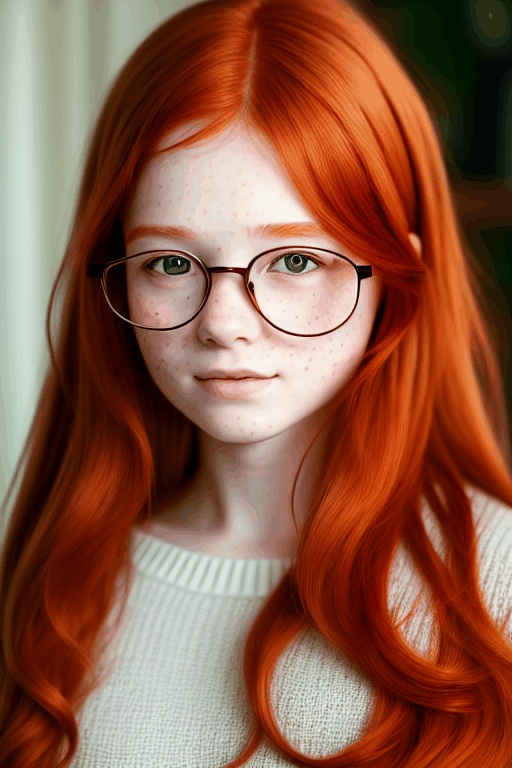}&
    \includegraphics[width=0.125\linewidth]{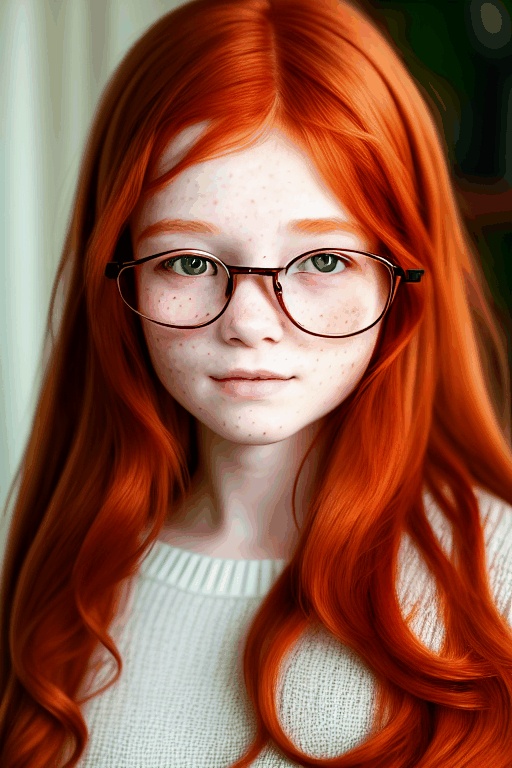}&
    \includegraphics[width=0.125\linewidth]{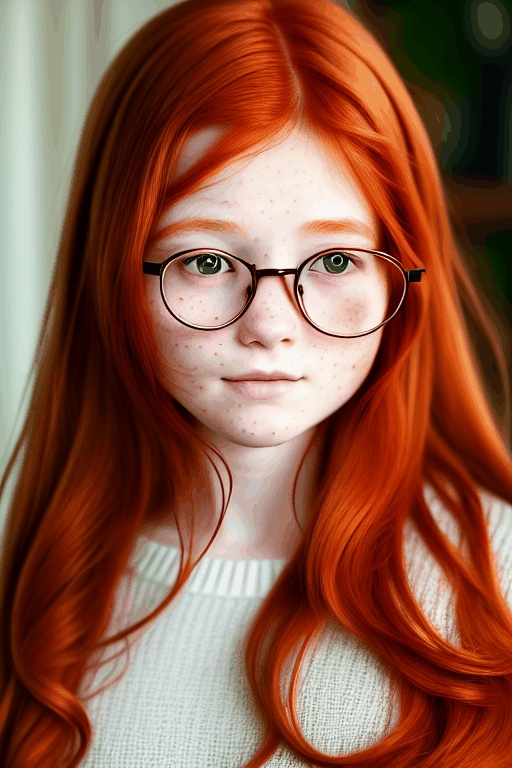}&
    \includegraphics[width=0.125\linewidth]{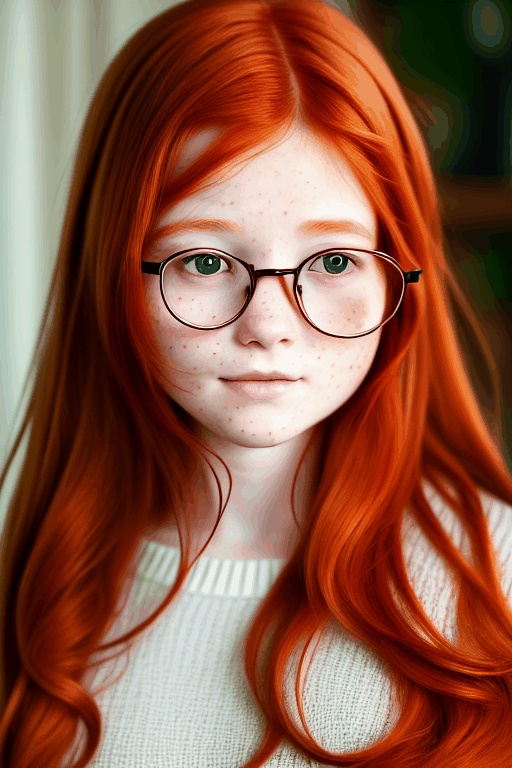}&
    \includegraphics[width=0.125\linewidth]{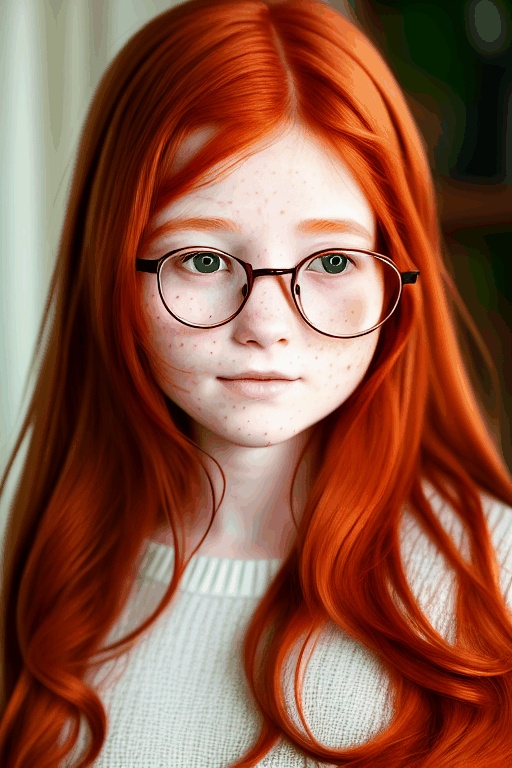}
    \\
    \multicolumn{8}{c}{\textit{``freckles, orange hair, glasses, ...''}}\vspace{0.5em}
    \\
    \includegraphics[width=0.125\linewidth]{gif/teaser/12/153_frame_0.jpg}&
    \includegraphics[width=0.125\linewidth]{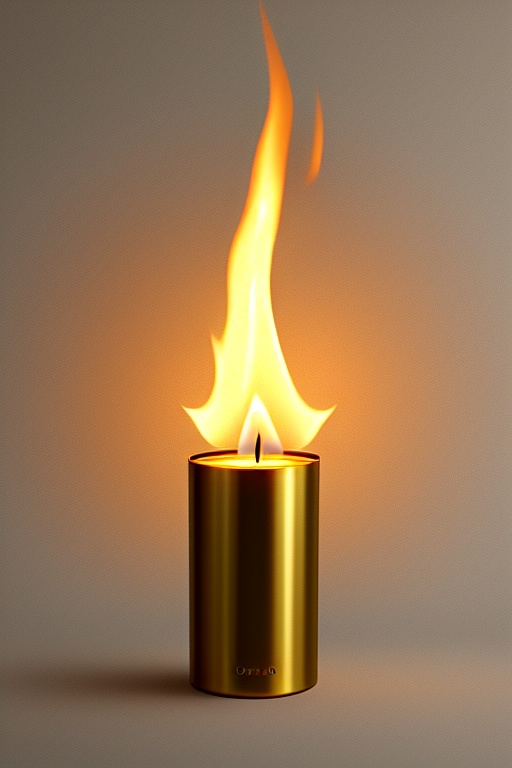}&
    \includegraphics[width=0.125\linewidth]{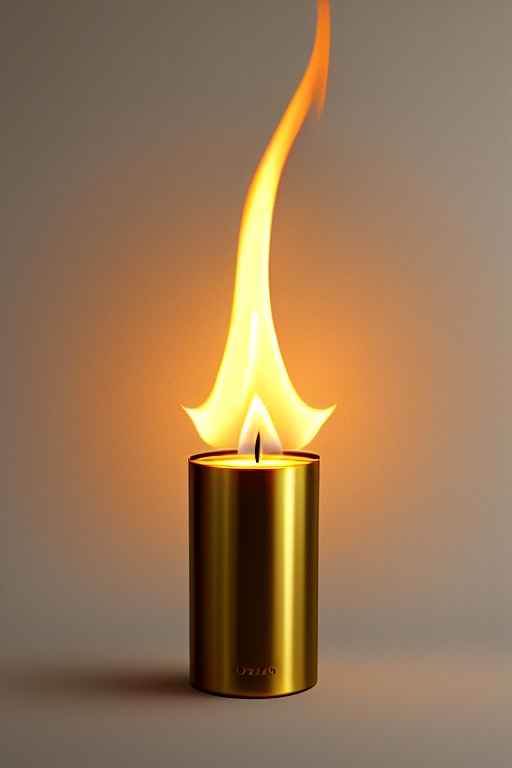}&
    \includegraphics[width=0.125\linewidth]{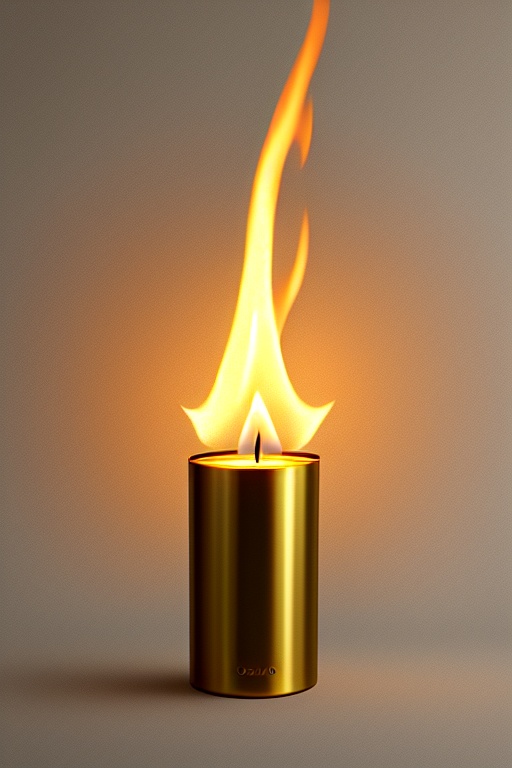}&
    \includegraphics[width=0.125\linewidth]{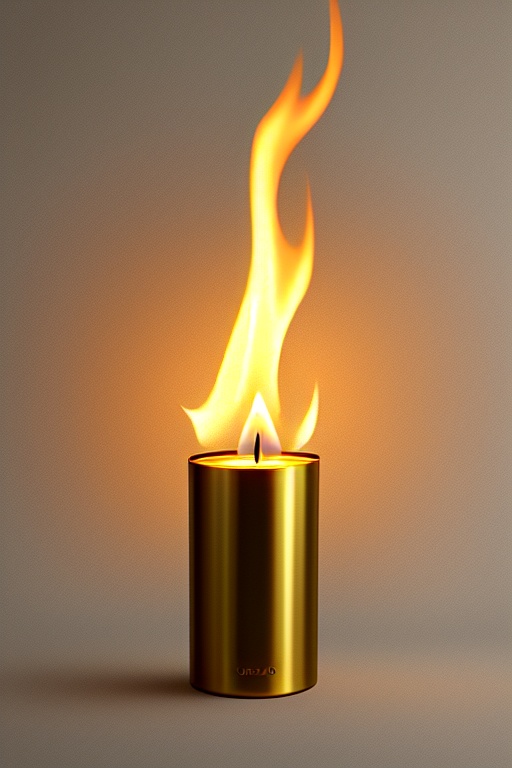}&
    \includegraphics[width=0.125\linewidth]{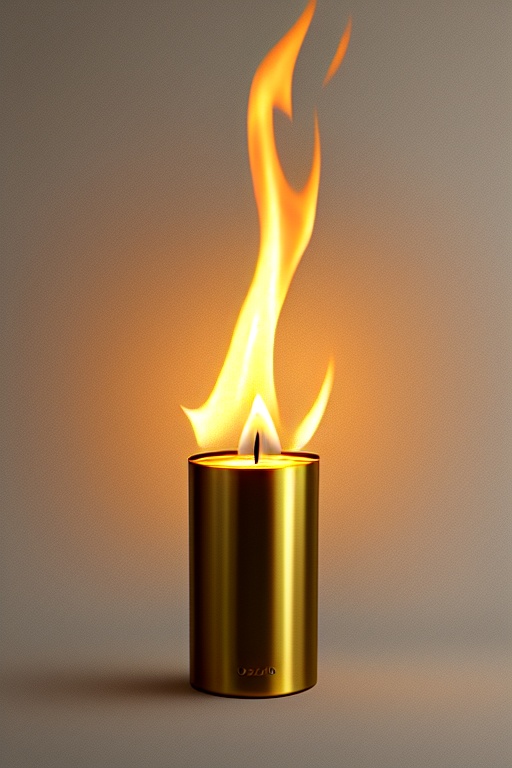}&
    \includegraphics[width=0.125\linewidth]{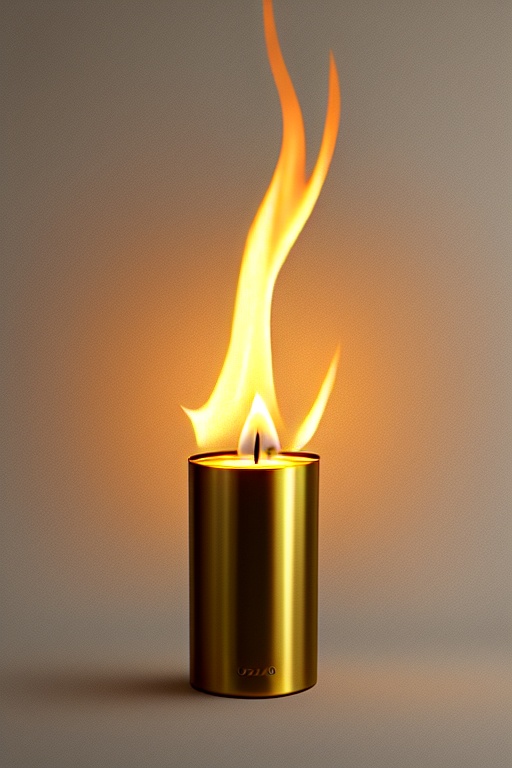}&
    \includegraphics[width=0.125\linewidth]{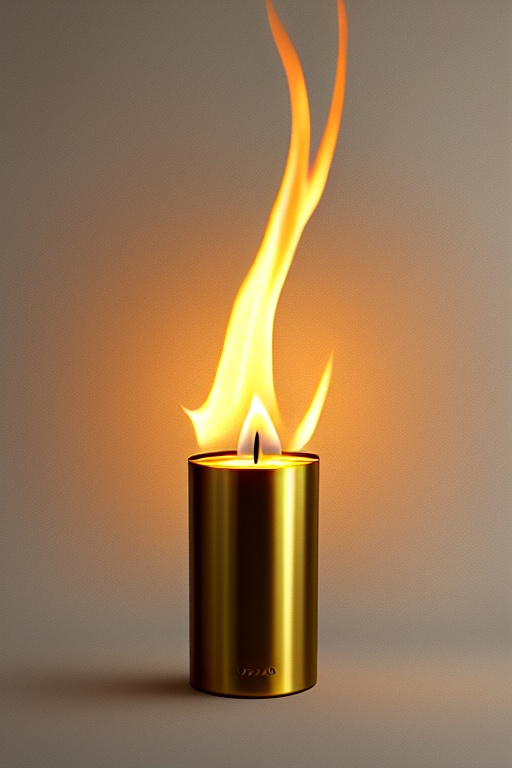}
    \\
    \multicolumn{8}{c}{\textit{``lighter, flame, candle''}}\vspace{0.5em}
    \end{tabular}
\vspace{-0.5em}
  \caption{Static frames sequences in Fig.~\ref{fig:teaser}(part 2).}
  \vspace{-1em}
\label{fig:static-fig1-2} 
\end{figure*}

%% file: append-fig-tex/static-fig4.tex
\begin{figure*}[ht]
  \centering
  \begin{tabular}{c@{\hspace{0.em}}c@{\hspace{0.em}}c@{\hspace{0.em}}c@{\hspace{0.5em}}c@{\hspace{0.em}}c@{\hspace{0.em}}c@{\hspace{0.em}}c} 
    \multicolumn{4}{c}{AnimateDiff~\cite{guo2023animatediff}} & \multicolumn{4}{c}{AnimateZero}\\
    \includegraphics[width=0.12\linewidth]{gif/ADvsAZ/526/526-0.jpg}&
    \includegraphics[width=0.12\linewidth]{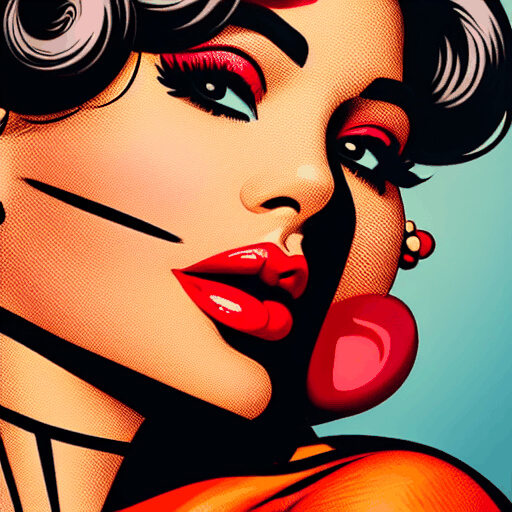}&
    \includegraphics[width=0.12\linewidth]{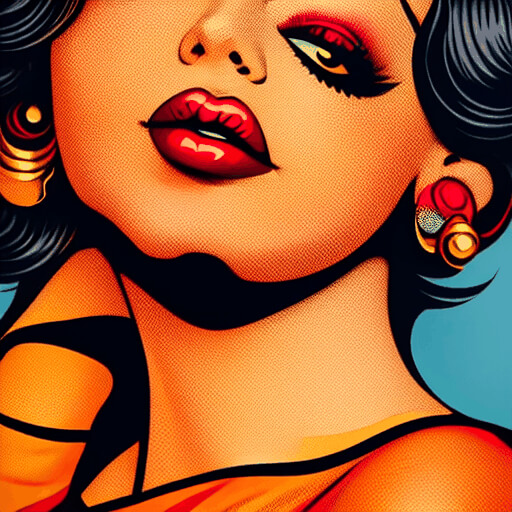}&
    \includegraphics[width=0.12\linewidth]{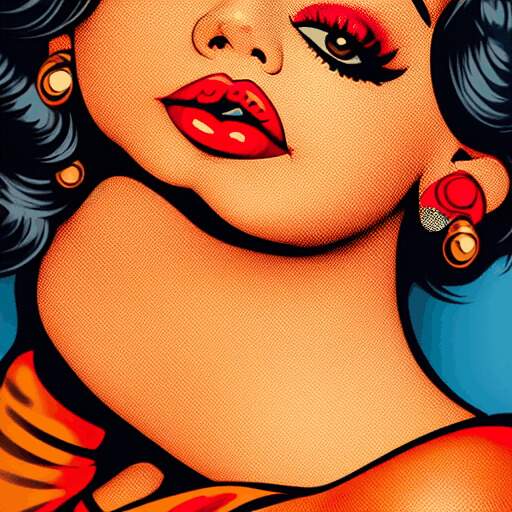}&
    \includegraphics[width=0.12\linewidth]{gif/ADvsAZ/525/525-0.jpg}&
    \includegraphics[width=0.12\linewidth]{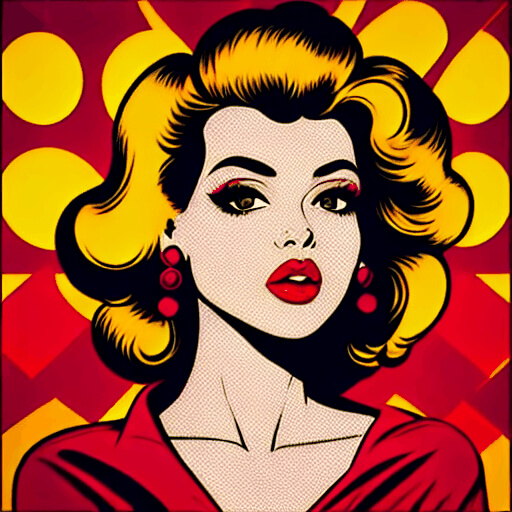}&
    \includegraphics[width=0.12\linewidth]{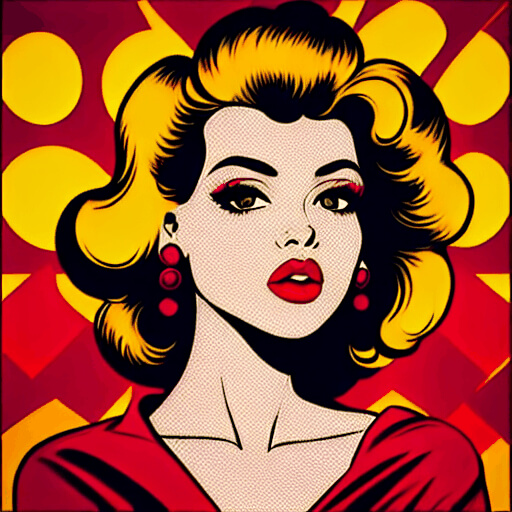}&
    \includegraphics[width=0.12\linewidth]{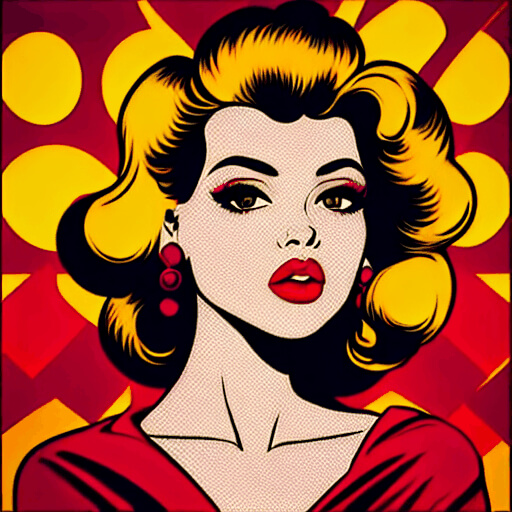}
    \\
    \multicolumn{8}{c}{\textit{``1girl, jewelry, \textcolor{red}{upper body}, earrings, pop art, ...''}}\vspace{0.5em}
    \\
    \includegraphics[width=0.12\linewidth]{gif/ADvsAZ/528/528-0.jpg}&
    \includegraphics[width=0.12\linewidth]{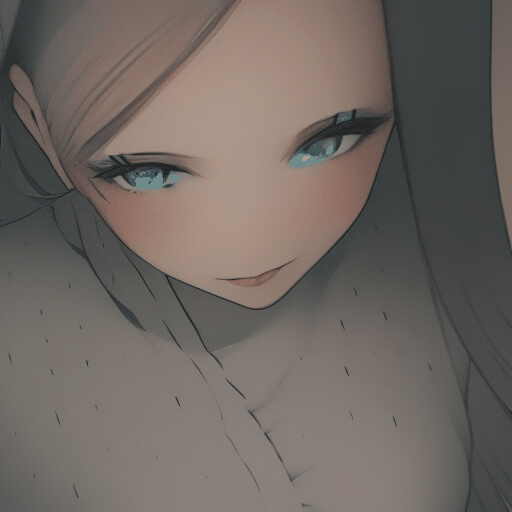}&
    \includegraphics[width=0.12\linewidth]{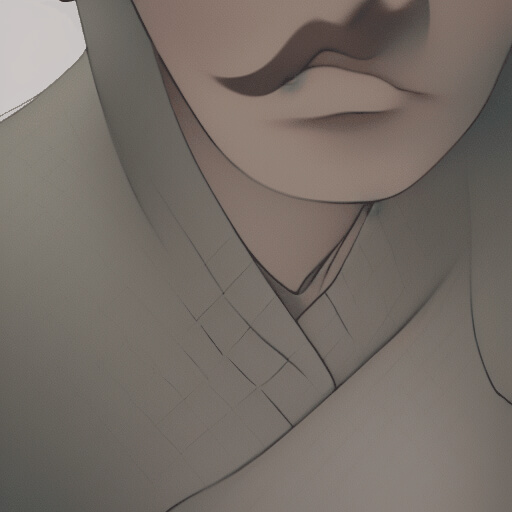}&
    \includegraphics[width=0.12\linewidth]{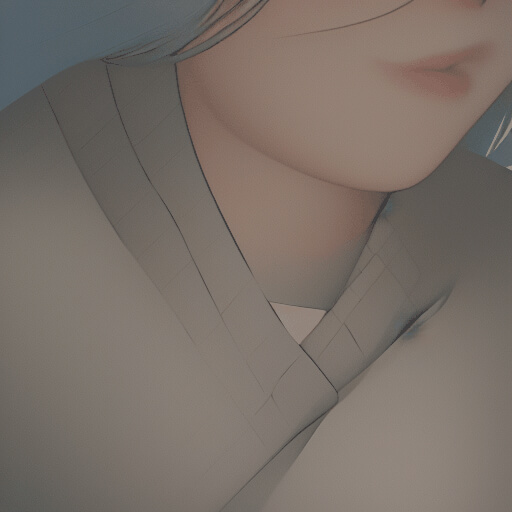}&
    \includegraphics[width=0.12\linewidth]{gif/ADvsAZ/531/531-0.jpg}&
    \includegraphics[width=0.12\linewidth]{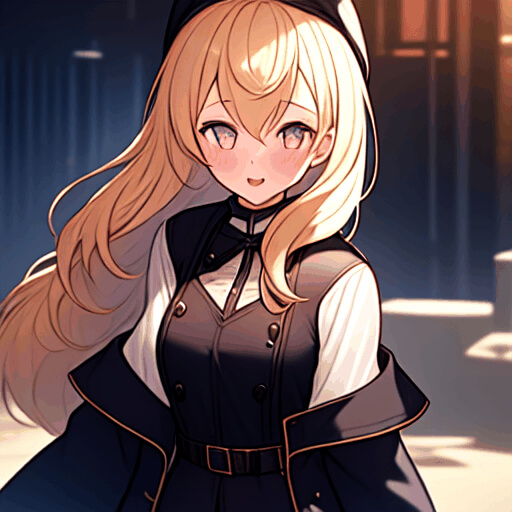}&
    \includegraphics[width=0.12\linewidth]{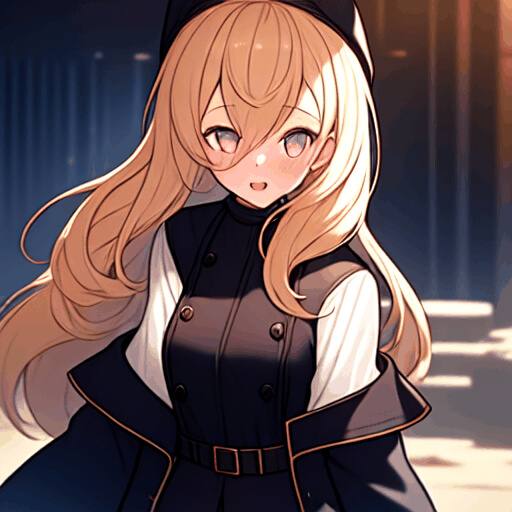}&
    \includegraphics[width=0.12\linewidth]{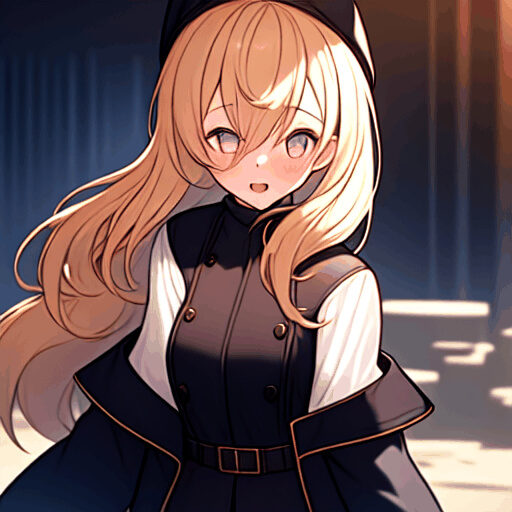}
    \\
    \multicolumn{8}{c}{\textit{``1girl, long hair, looking at the camera, ...''}}\vspace{0.5em}
    \\
    \includegraphics[width=0.12\linewidth]{gif/ADvsAZ/532/532-0.jpg}&
    \includegraphics[width=0.12\linewidth]{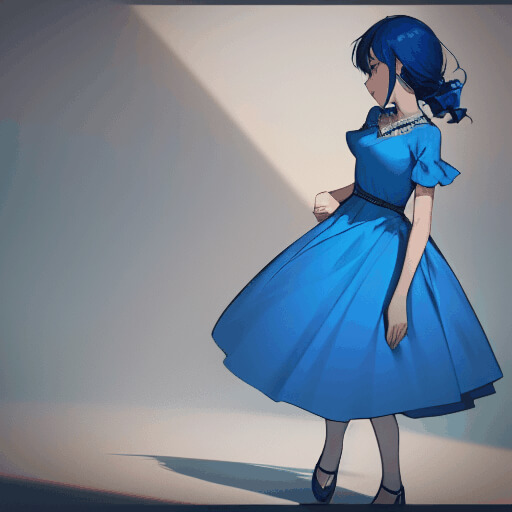}&
    \includegraphics[width=0.12\linewidth]{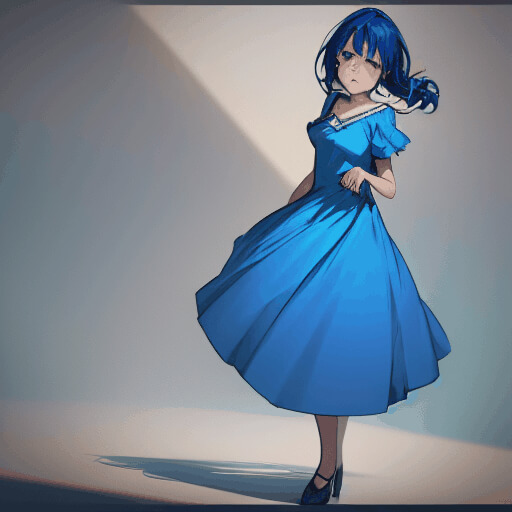}&
    \includegraphics[width=0.12\linewidth]{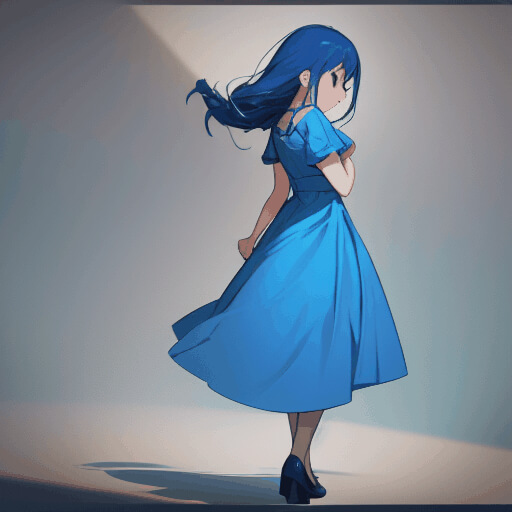}&
    \includegraphics[width=0.12\linewidth]{gif/ADvsAZ/533/533-0.jpg}&
    \includegraphics[width=0.12\linewidth]{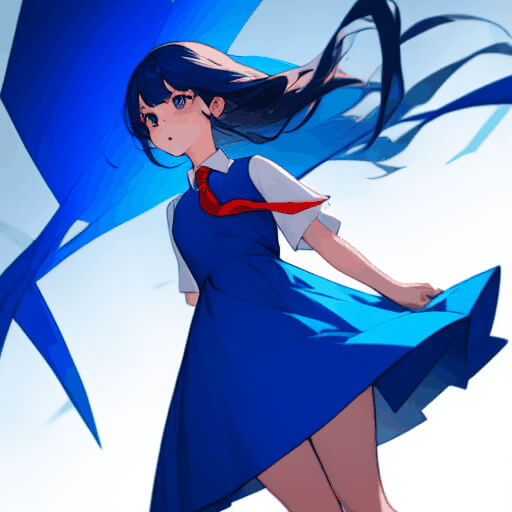}&
    \includegraphics[width=0.12\linewidth]{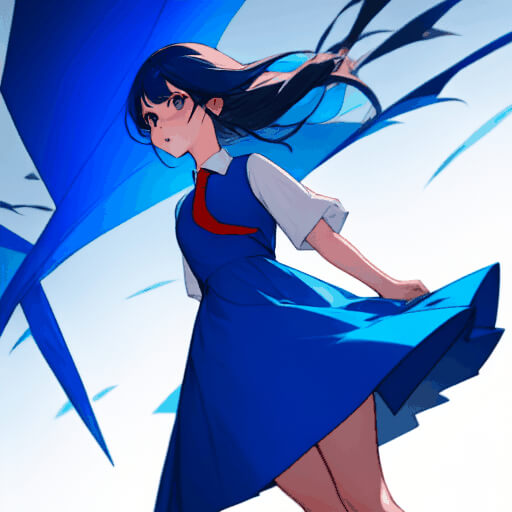}&
    \includegraphics[width=0.12\linewidth]{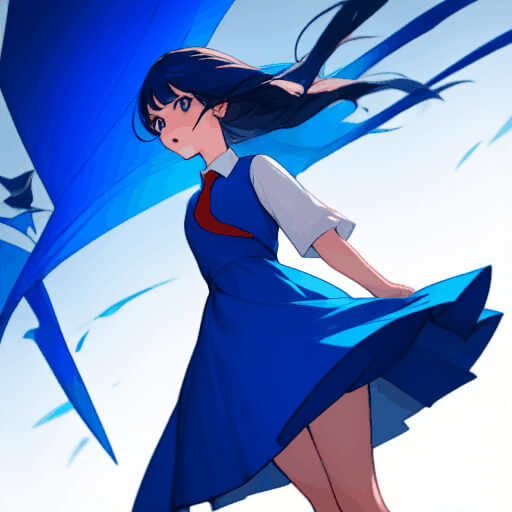}
    \\
    \multicolumn{8}{c}{\textit{``1girl, blue dress, \textcolor{red}{red tie}, floating blue, ...''}}\vspace{0.5em}
    \\
    \includegraphics[width=0.12\linewidth]{gif/ADvsAZ/545/545-0.jpg}&
    \includegraphics[width=0.12\linewidth]{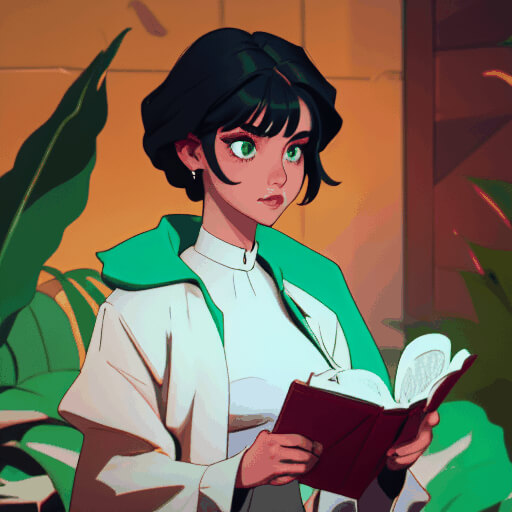}&
    \includegraphics[width=0.12\linewidth]{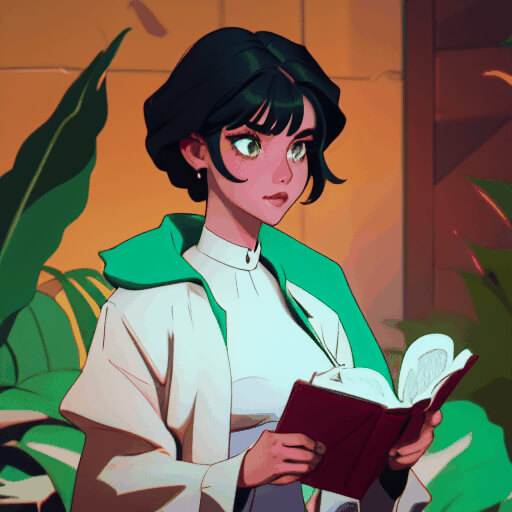}&
    \includegraphics[width=0.12\linewidth]{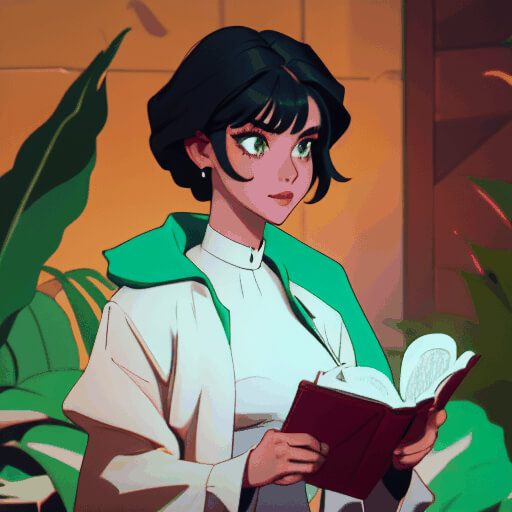}&
    \includegraphics[width=0.12\linewidth]{gif/ADvsAZ/549/549-0.jpg}&
    \includegraphics[width=0.12\linewidth]{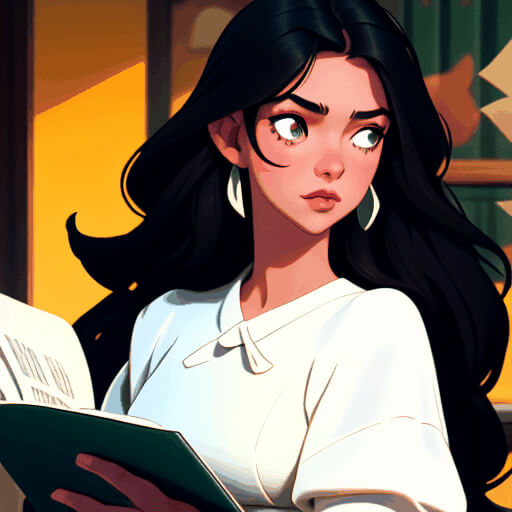}&
    \includegraphics[width=0.12\linewidth]{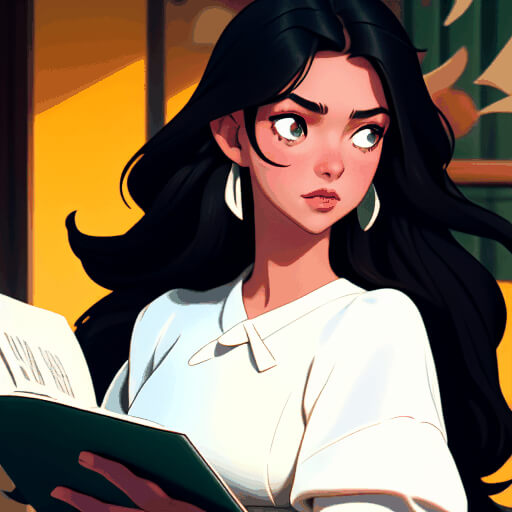}&
    \includegraphics[width=0.12\linewidth]{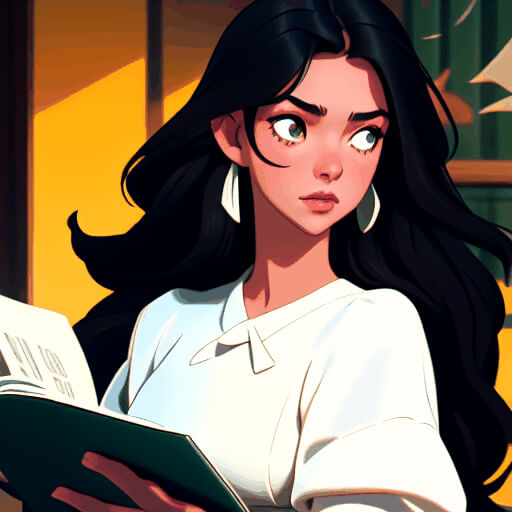}
    \\
    \multicolumn{8}{c}{\textit{``1girl wearing white \textcolor{red}{dress} is reading \textcolor{red}{green} book, ...''}}\vspace{0.5em}
    \end{tabular}
\vspace{-0.5em}
  \caption{Static frames sequences in Fig.~\ref{fig:ADvsAZ}.}
  \vspace{-1em}
\label{fig:static-fig4} 
\end{figure*}

%% file: append-fig-tex/static-fig5.tex
\begin{figure*}[ht]
  \centering
  \begin{tabular}{c@{\hspace{0.5em}}c@{\hspace{0.em}}c@{\hspace{0.em}}c@{\hspace{0.em}}c@{\hspace{0.0em}}c@{\hspace{0.em}}c@{\hspace{0.em}}c@{\hspace{0.em}}c} 
  \rotatebox[origin=l]{90}{Gen-2~\cite{gen2}}&
    \includegraphics[width=0.12\linewidth]{gif/AZvsOthers/line3/gen2/011_0.jpg}&
    \includegraphics[width=0.12\linewidth]{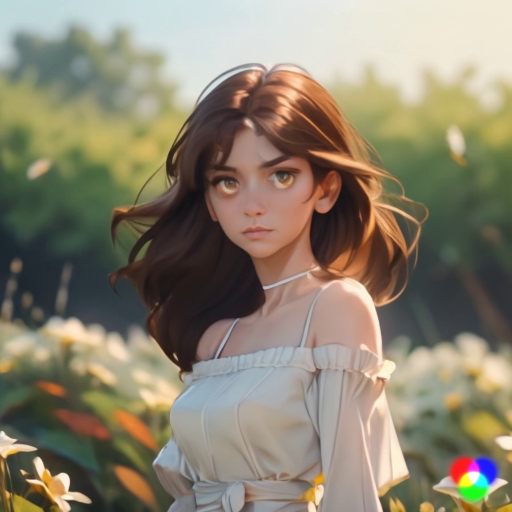}&
    \includegraphics[width=0.12\linewidth]{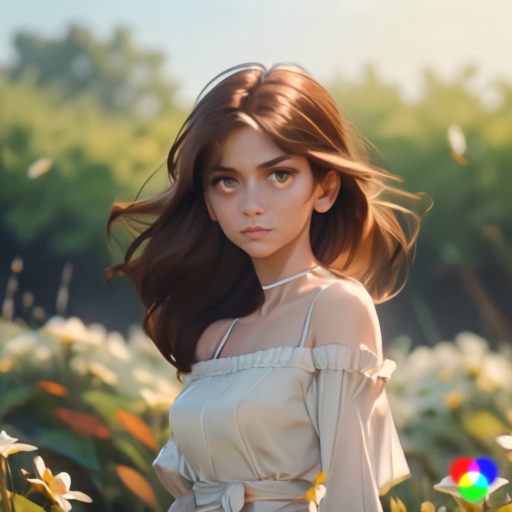}&
    \includegraphics[width=0.12\linewidth]{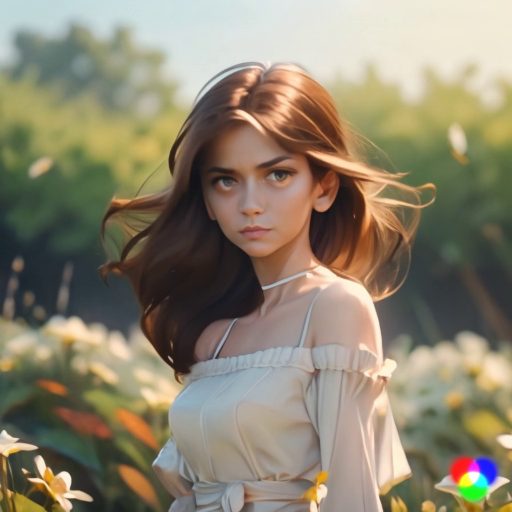}&
    \includegraphics[width=0.12\linewidth]{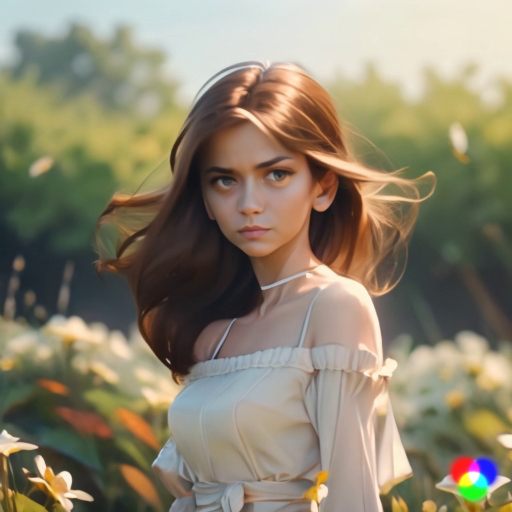}&
    \includegraphics[width=0.12\linewidth]{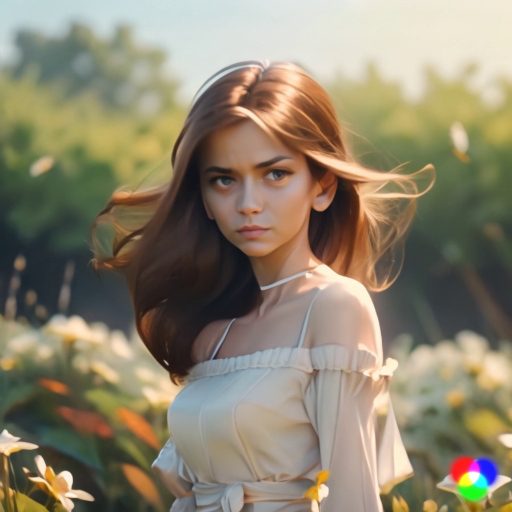}&
    \includegraphics[width=0.12\linewidth]{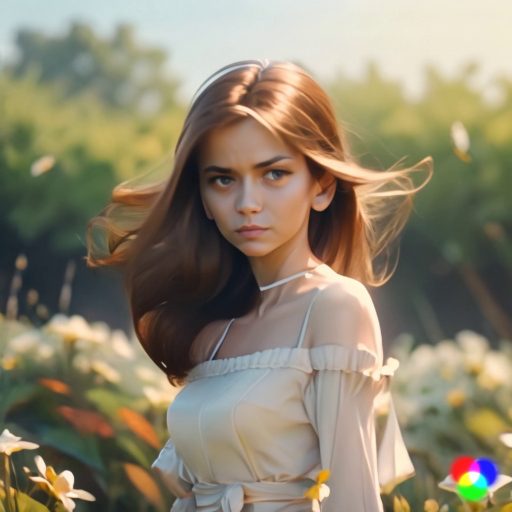}&
    \includegraphics[width=0.12\linewidth]{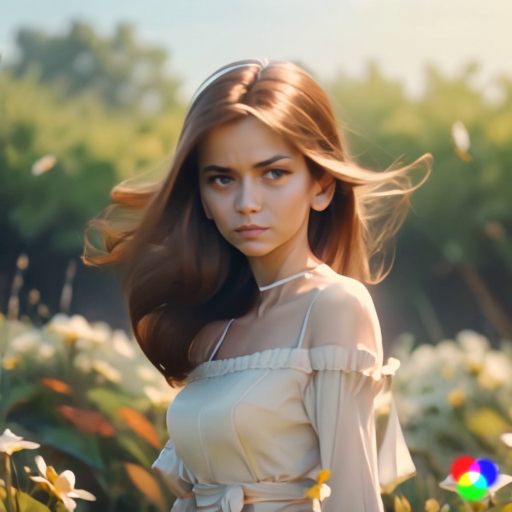}
    \\
    \rotatebox[origin=l]{90}{Genmo~\cite{genmo}}&
    \includegraphics[width=0.12\linewidth]{gif/AZvsOthers/line3/genmo/011_0.jpg}&
    \includegraphics[width=0.12\linewidth]{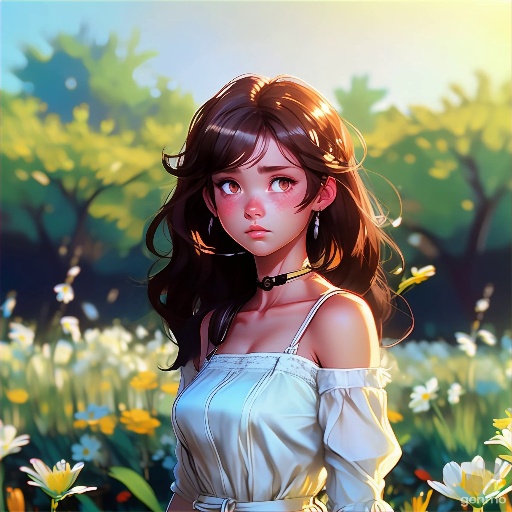}&
    \includegraphics[width=0.12\linewidth]{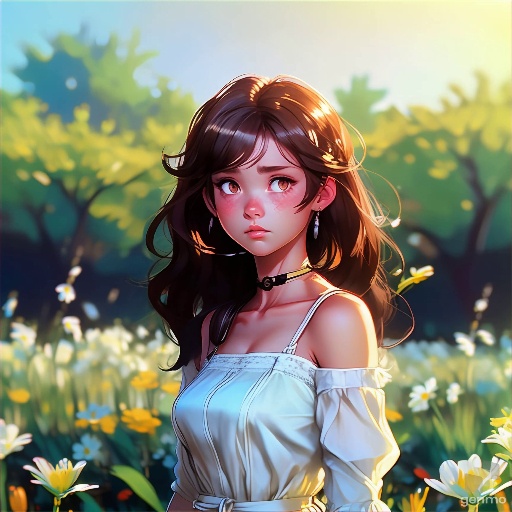}&
    \includegraphics[width=0.12\linewidth]{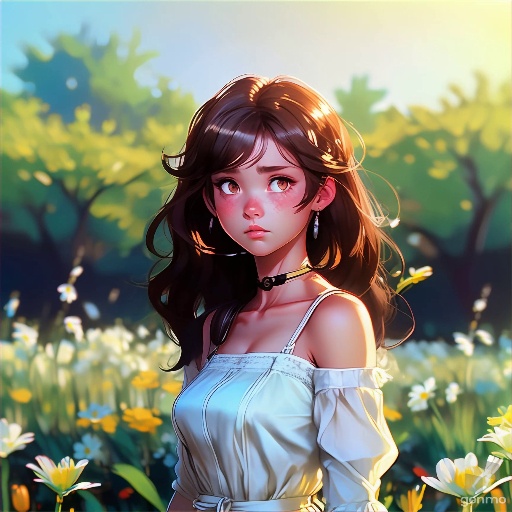}&
    \includegraphics[width=0.12\linewidth]{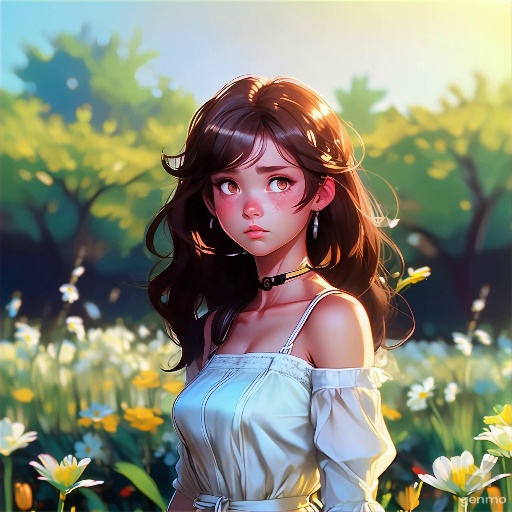}&
    \includegraphics[width=0.12\linewidth]{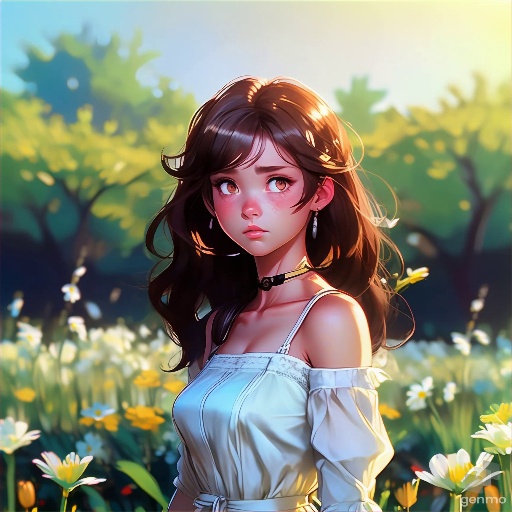}&
    \includegraphics[width=0.12\linewidth]{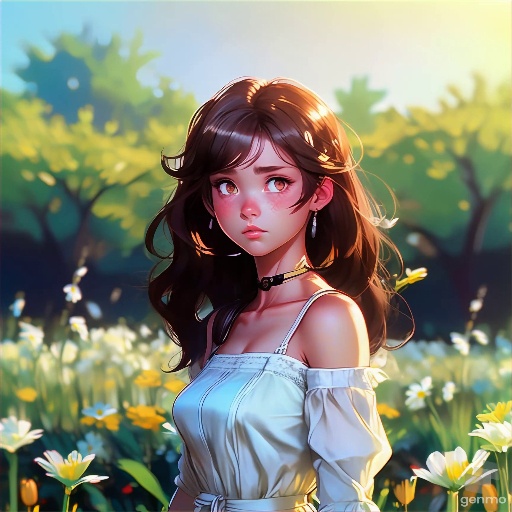}&
    \includegraphics[width=0.12\linewidth]{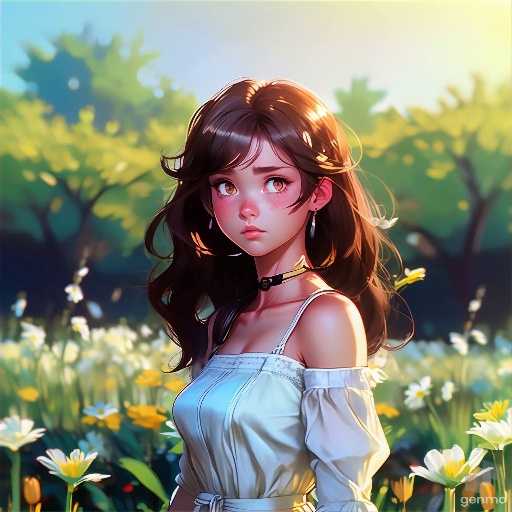}
    \\
    \rotatebox[origin=l]{90}{Pika Labs~\cite{pikalabs}}&
    \includegraphics[width=0.12\linewidth]{gif/AZvsOthers/line3/pika/011_0.jpg}&
    \includegraphics[width=0.12\linewidth]{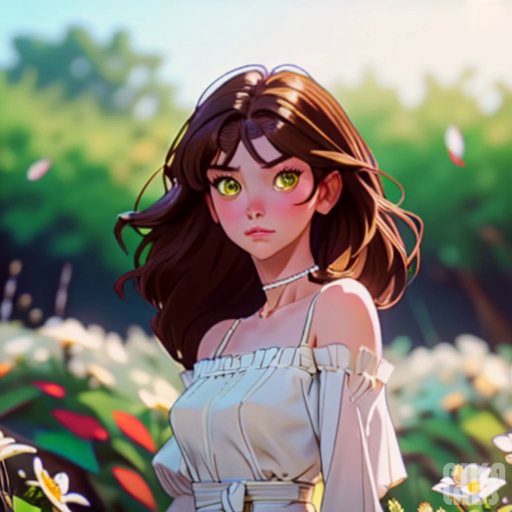}&
    \includegraphics[width=0.12\linewidth]{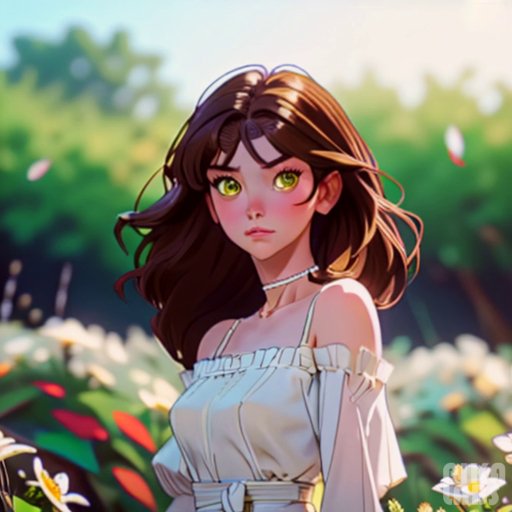}&
    \includegraphics[width=0.12\linewidth]{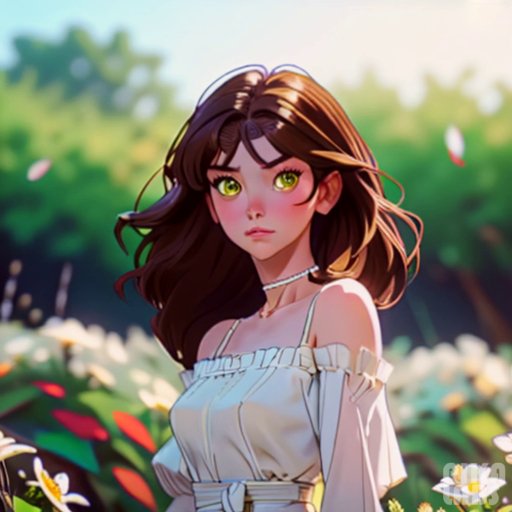}&
    \includegraphics[width=0.12\linewidth]{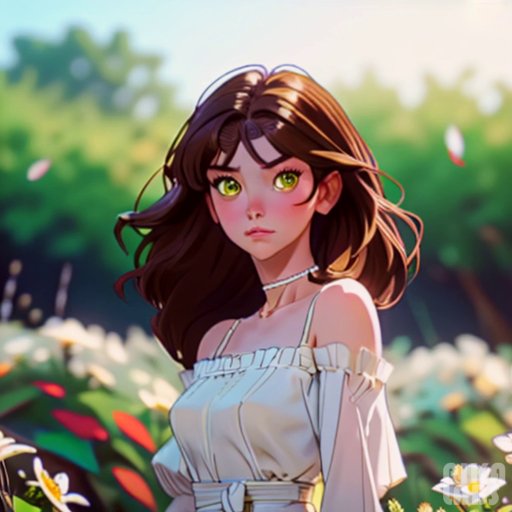}&
    \includegraphics[width=0.12\linewidth]{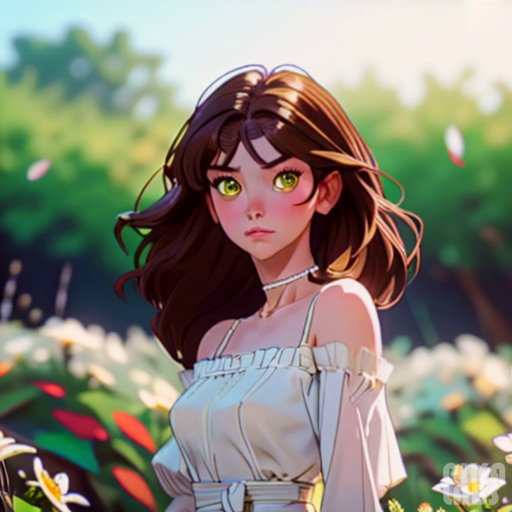}&
    \includegraphics[width=0.12\linewidth]{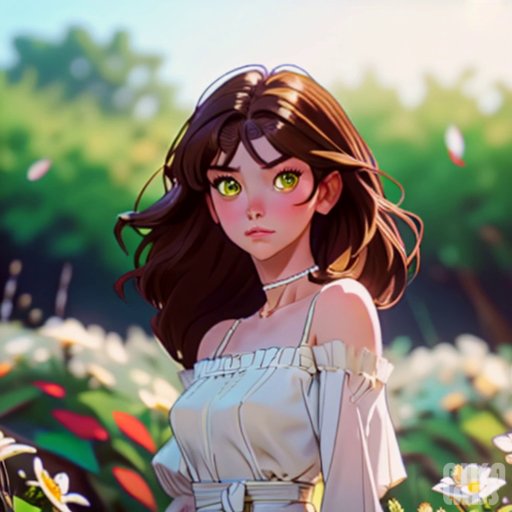}&
    \includegraphics[width=0.12\linewidth]{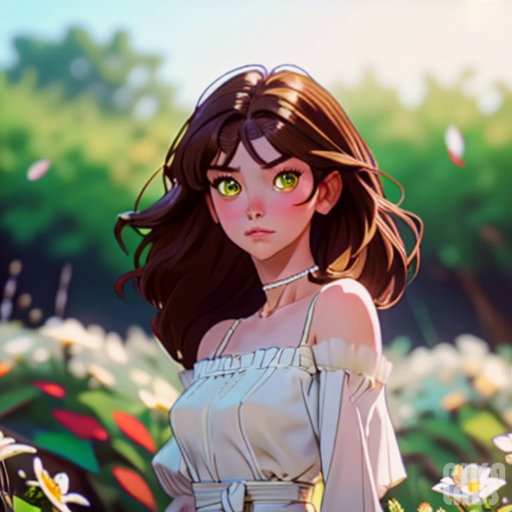}
    \\
    \rotatebox[origin=l]{90}{\scriptsize VideoCrafter1~\cite{chen2023videocrafter1}}&
    \includegraphics[width=0.12\linewidth]{gif/AZvsOthers/line3/videocrafter/011_0.jpg}&
    \includegraphics[width=0.12\linewidth]{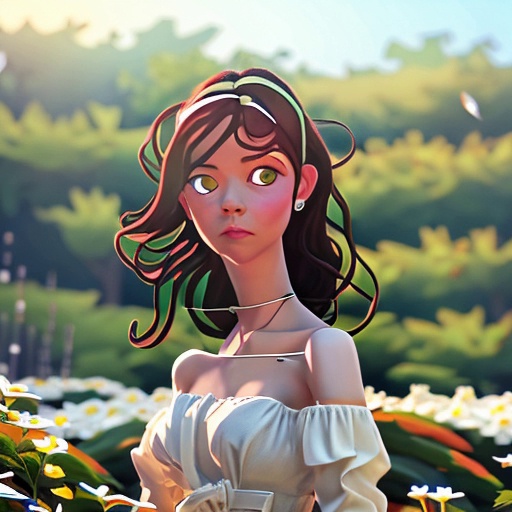}&
    \includegraphics[width=0.12\linewidth]{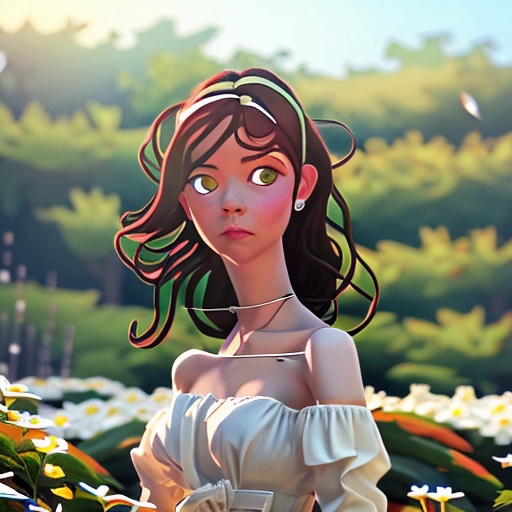}&
    \includegraphics[width=0.12\linewidth]{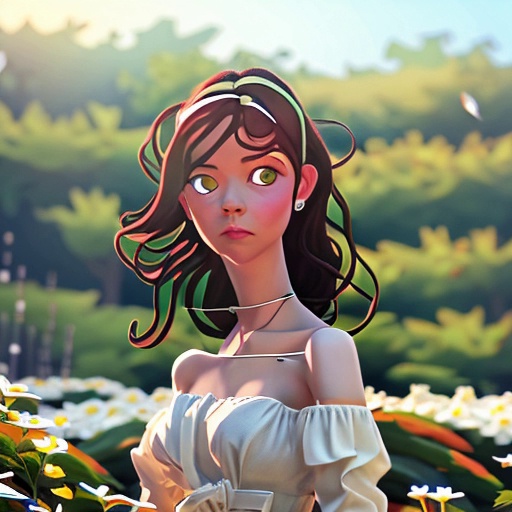}&
    \includegraphics[width=0.12\linewidth]{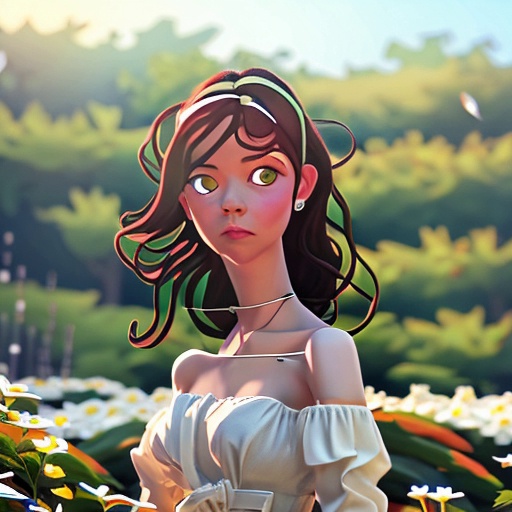}&
    \includegraphics[width=0.12\linewidth]{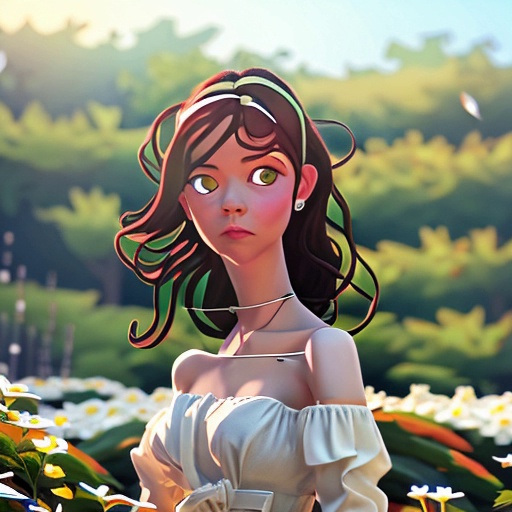}&
    \includegraphics[width=0.12\linewidth]{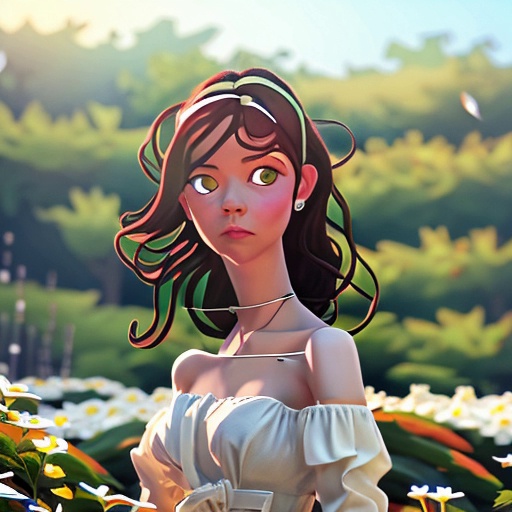}&
    \includegraphics[width=0.12\linewidth]{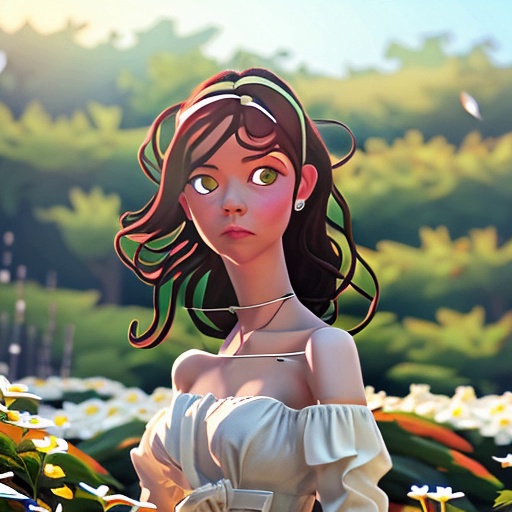}
    \\
    \rotatebox[origin=l]{90}{\small I2VGen-XL~\cite{i2vgenxl}}&
    \includegraphics[width=0.12\linewidth]{gif/AZvsOthers/line3/i2vgenxl/011_0.jpg}&
    \includegraphics[width=0.12\linewidth]{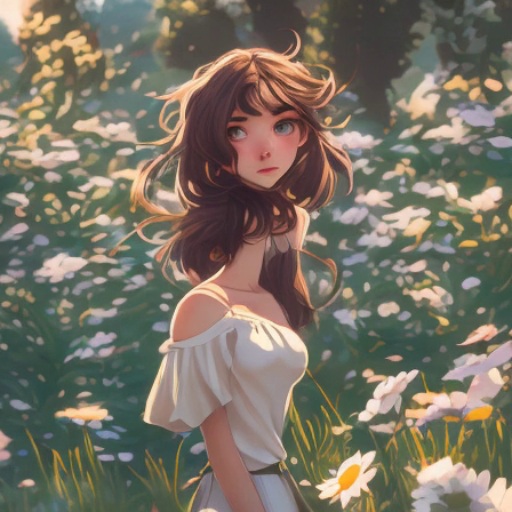}&
    \includegraphics[width=0.12\linewidth]{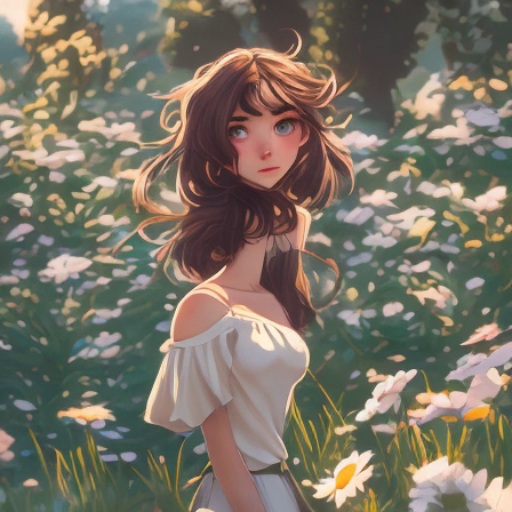}&
    \includegraphics[width=0.12\linewidth]{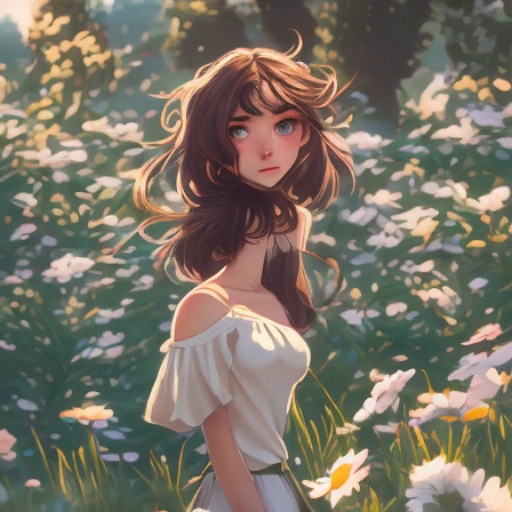}&
    \includegraphics[width=0.12\linewidth]{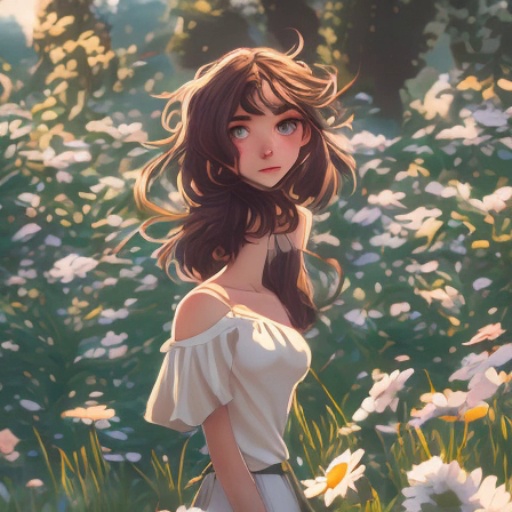}&
    \includegraphics[width=0.12\linewidth]{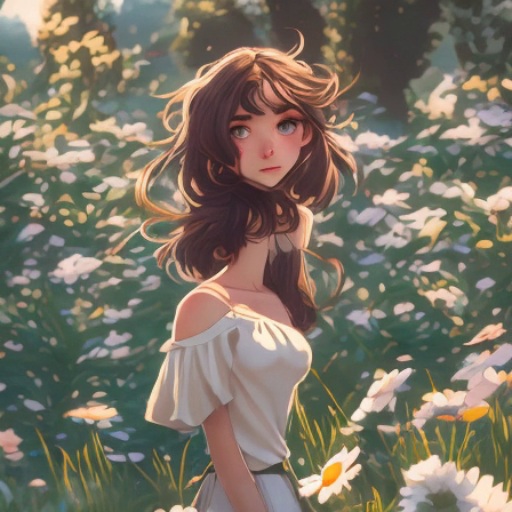}&
    \includegraphics[width=0.12\linewidth]{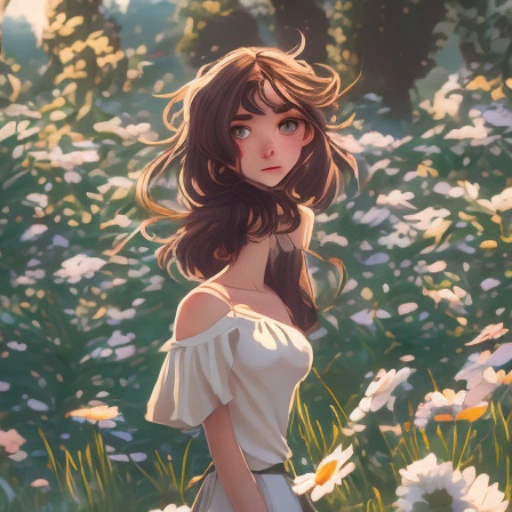}&
    \includegraphics[width=0.12\linewidth]{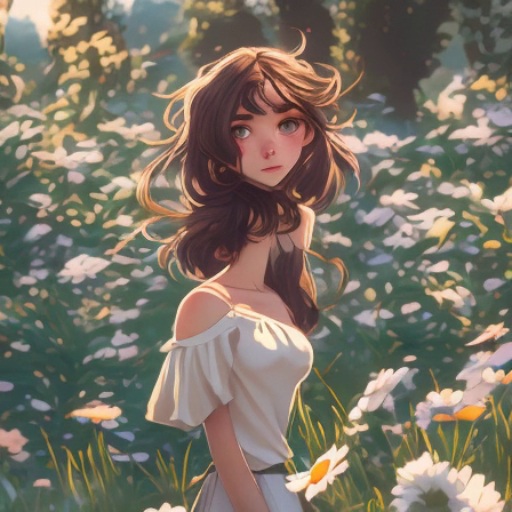}
    \\
    \rotatebox[origin=l]{90}{AnimateZero}&
    \includegraphics[width=0.12\linewidth]{gif/AZvsOthers/line3/az/011_0.jpg}&
    \includegraphics[width=0.12\linewidth]{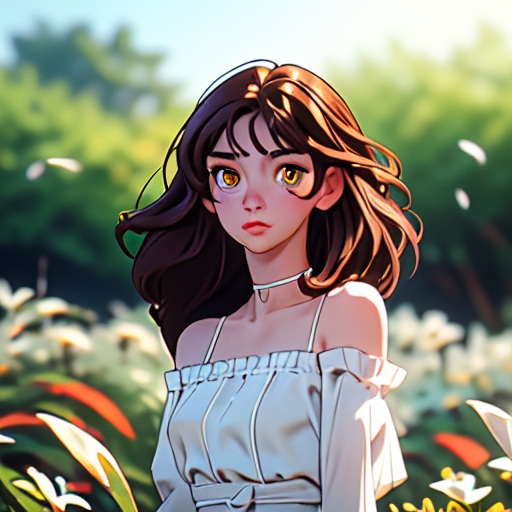}&
    \includegraphics[width=0.12\linewidth]{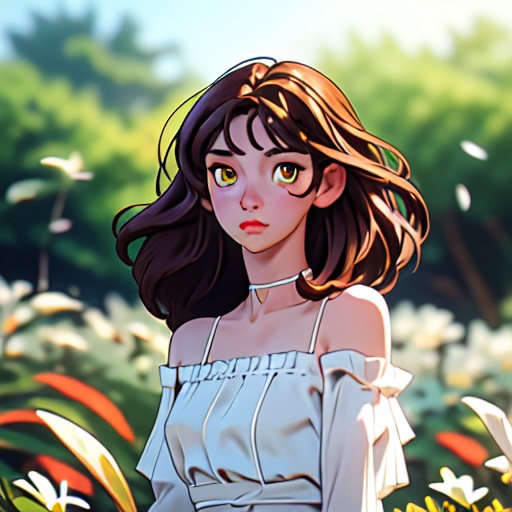}&
    \includegraphics[width=0.12\linewidth]{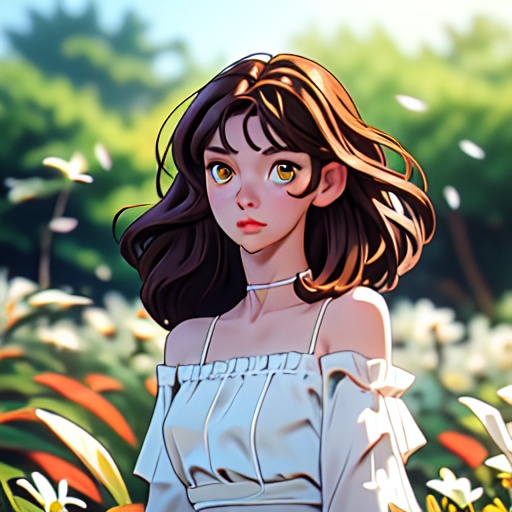}&
    \includegraphics[width=0.12\linewidth]{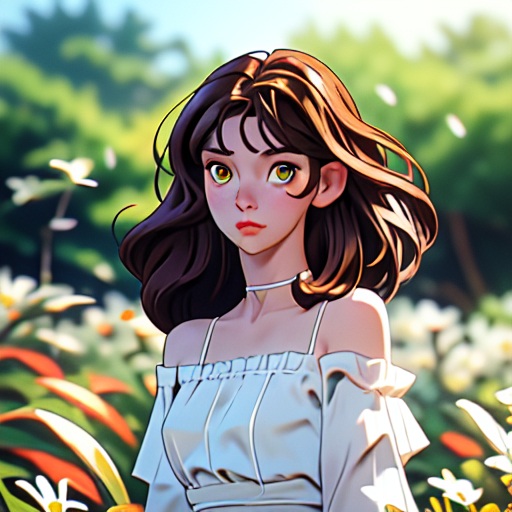}&
    \includegraphics[width=0.12\linewidth]{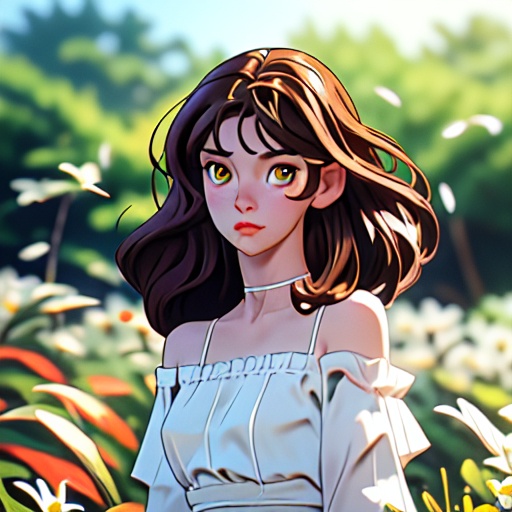}&
    \includegraphics[width=0.12\linewidth]{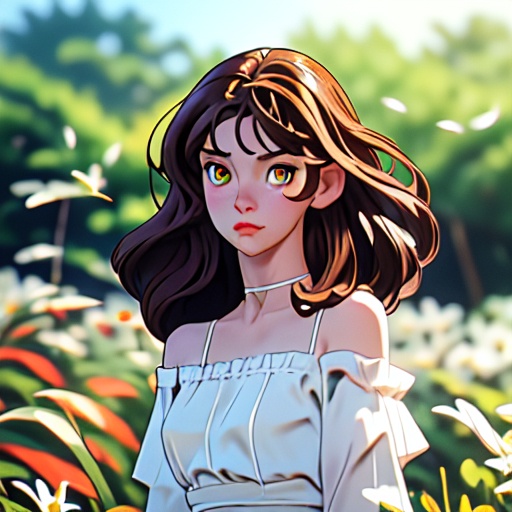}&
    \includegraphics[width=0.12\linewidth]{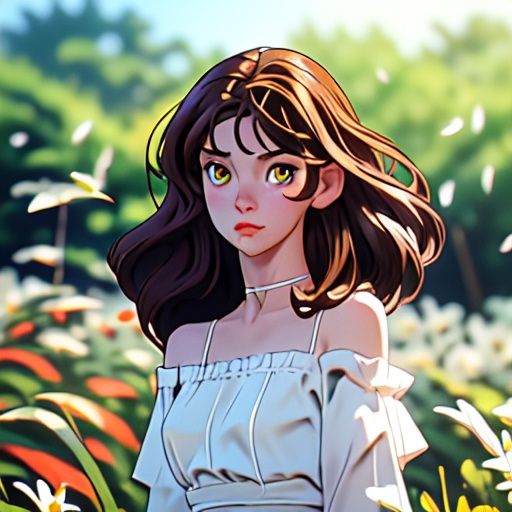}
    \\
    \multicolumn{9}{c}{\textit{``1girl, brown hair, a lot of white flowers, leaf, blurry foreground, ...''}}
    \end{tabular}
\vspace{-0.5em}
  \caption{Static frames sequences in Fig.~\ref{fig:ADvsOthers} (part 1).}
  \vspace{-1em}
\label{fig:static-fig5-1} 
\end{figure*}

\begin{figure*}[ht]
  \centering
  \begin{tabular}{c@{\hspace{0.5em}}c@{\hspace{0.em}}c@{\hspace{0.em}}c@{\hspace{0.em}}c@{\hspace{0.0em}}c@{\hspace{0.em}}c@{\hspace{0.em}}c@{\hspace{0.em}}c} 
  \rotatebox[origin=l]{90}{Gen-2~\cite{gen2}}&
    \includegraphics[width=0.12\linewidth]{gif/AZvsOthers/line2/gen2/007_0.jpg}&
    \includegraphics[width=0.12\linewidth]{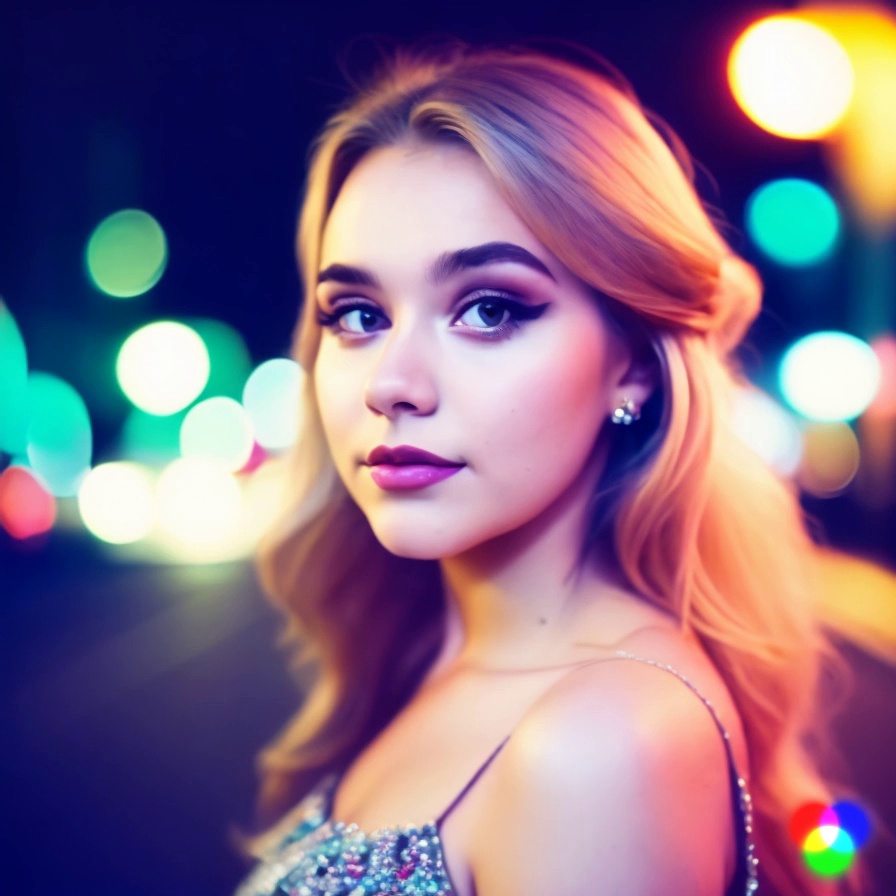}&
    \includegraphics[width=0.12\linewidth]{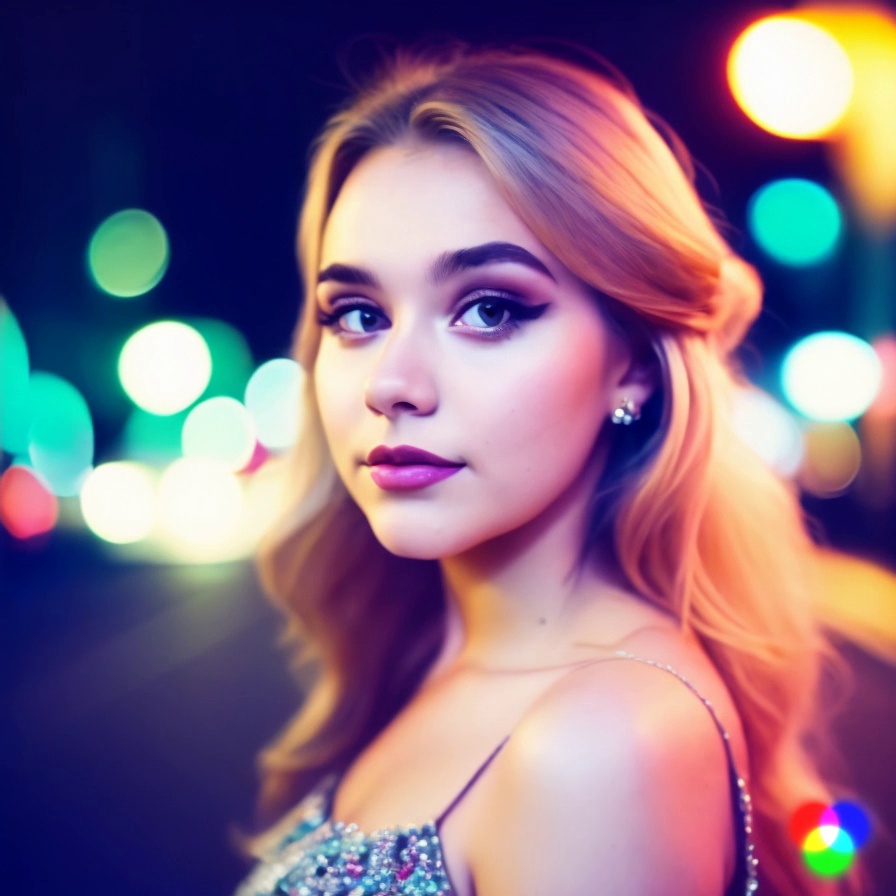}&
    \includegraphics[width=0.12\linewidth]{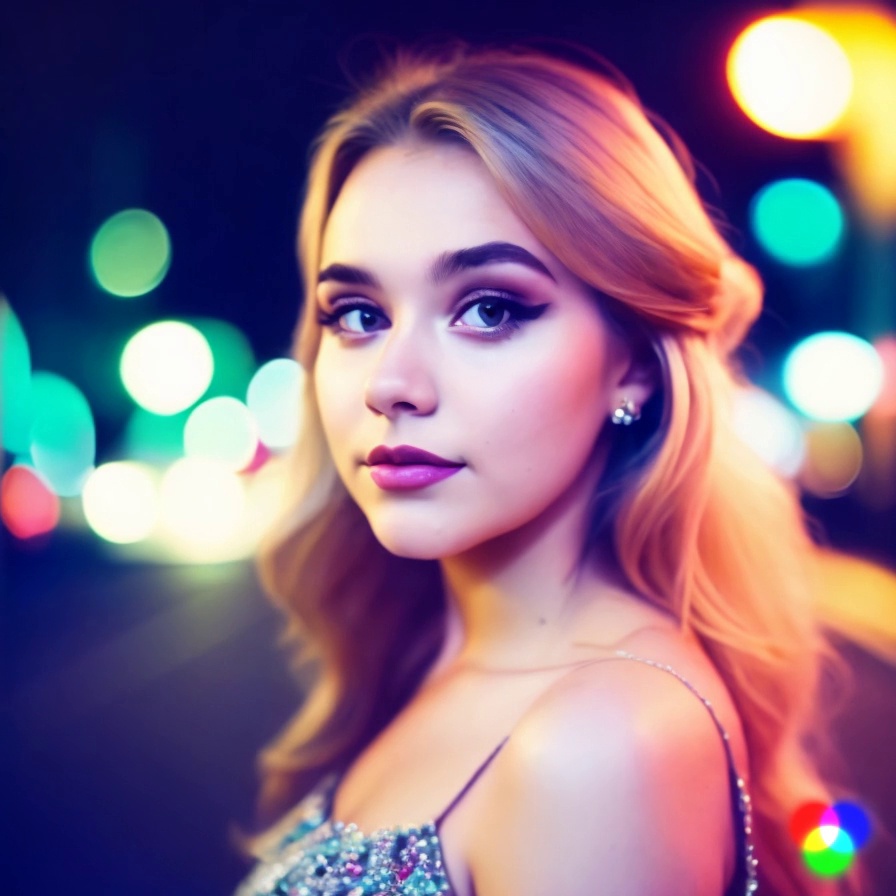}&
    \includegraphics[width=0.12\linewidth]{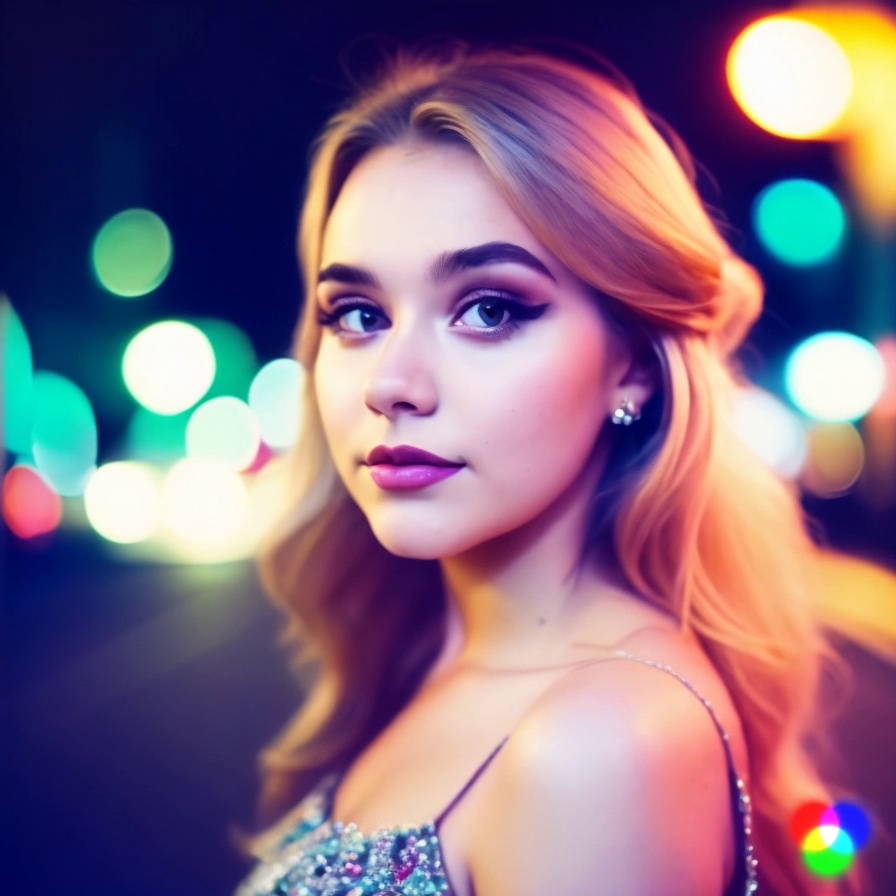}&
    \includegraphics[width=0.12\linewidth]{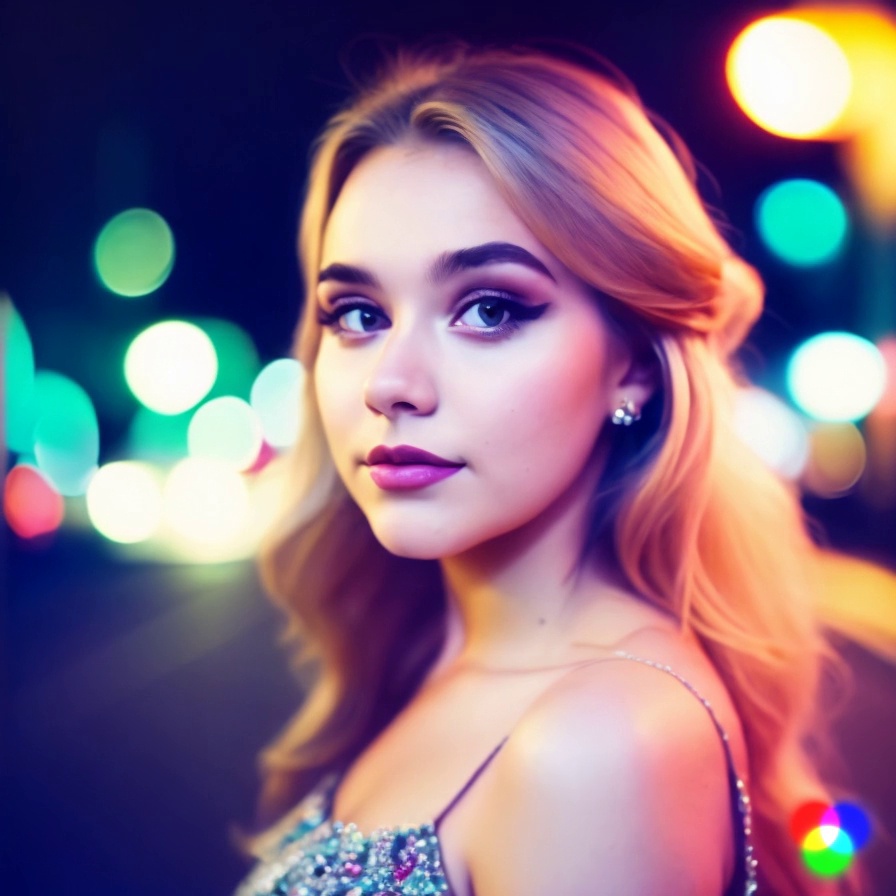}&
    \includegraphics[width=0.12\linewidth]{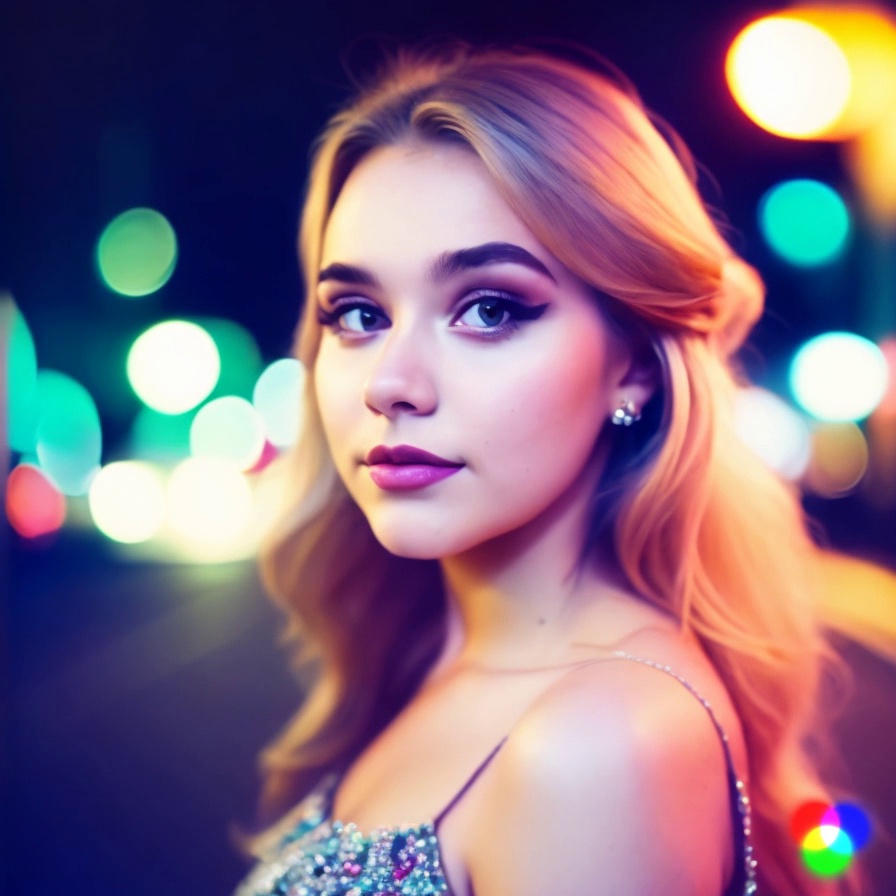}&
    \includegraphics[width=0.12\linewidth]{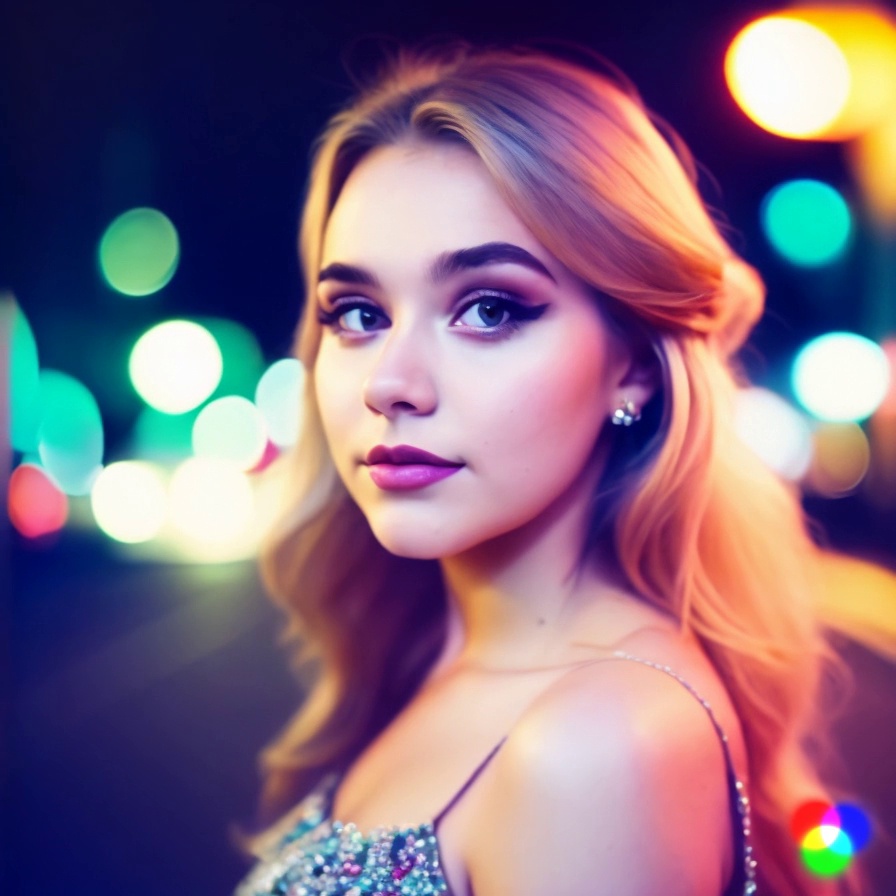}
    \\
    \rotatebox[origin=l]{90}{Genmo~\cite{genmo}}&
    \includegraphics[width=0.12\linewidth]{gif/AZvsOthers/line2/genmo/007_0.jpg}&
    \includegraphics[width=0.12\linewidth]{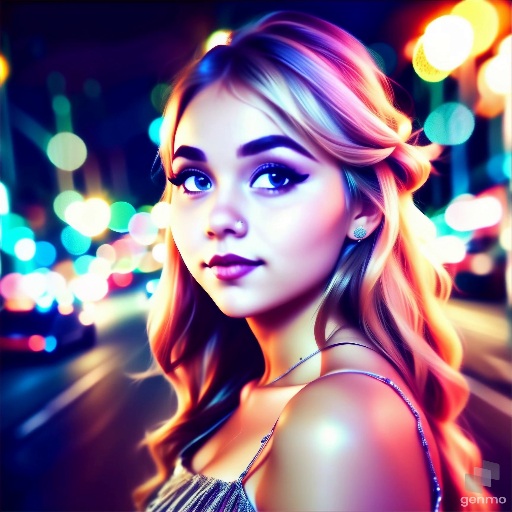}&
    \includegraphics[width=0.12\linewidth]{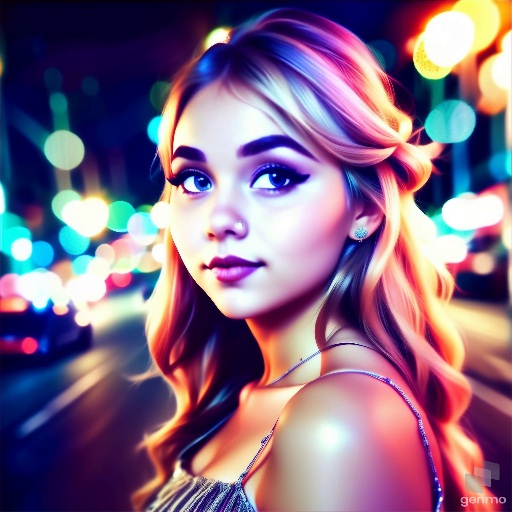}&
    \includegraphics[width=0.12\linewidth]{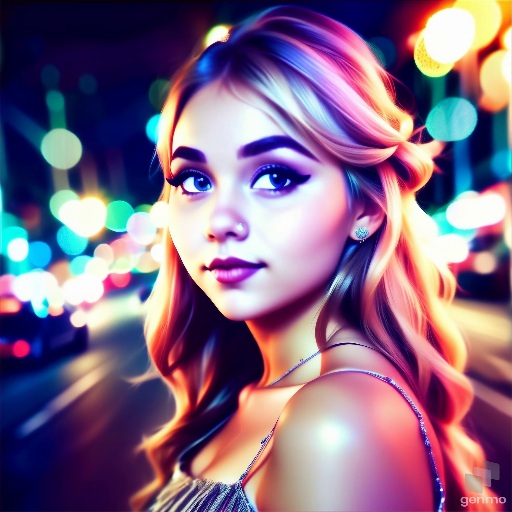}&
    \includegraphics[width=0.12\linewidth]{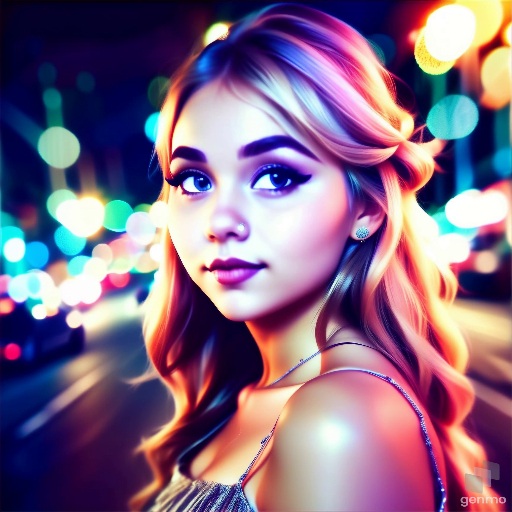}&
    \includegraphics[width=0.12\linewidth]{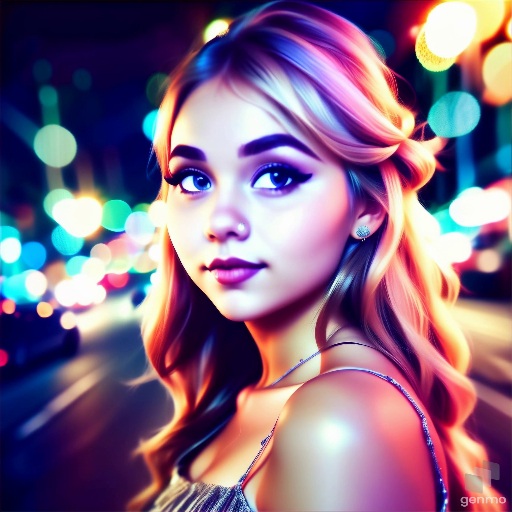}&
    \includegraphics[width=0.12\linewidth]{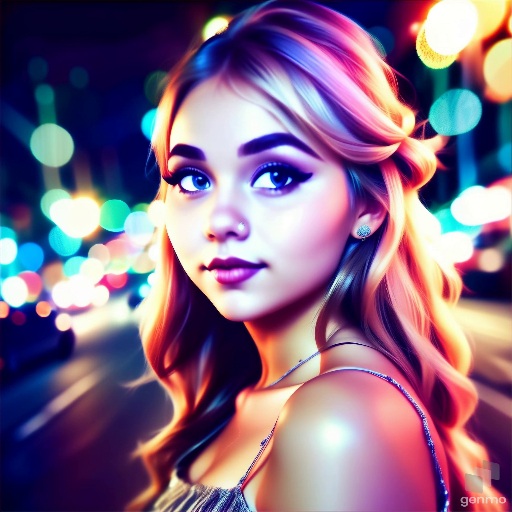}&
    \includegraphics[width=0.12\linewidth]{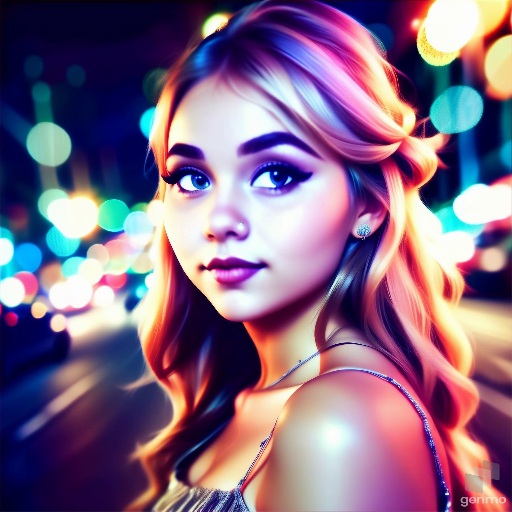}
    \\
    \rotatebox[origin=l]{90}{Pika Labs~\cite{pikalabs}}&
    \includegraphics[width=0.12\linewidth]{gif/AZvsOthers/line2/pika/007_0.jpg}&
    \includegraphics[width=0.12\linewidth]{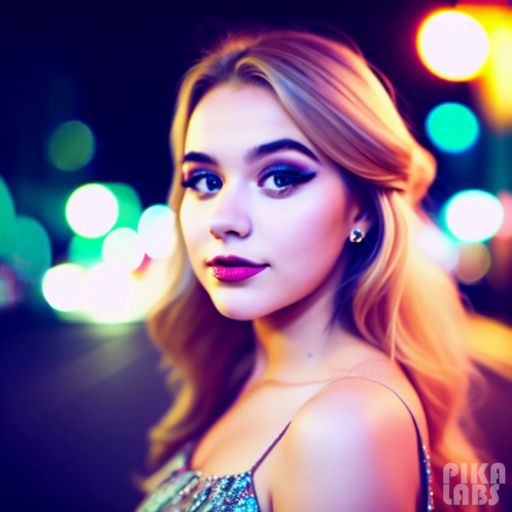}&
    \includegraphics[width=0.12\linewidth]{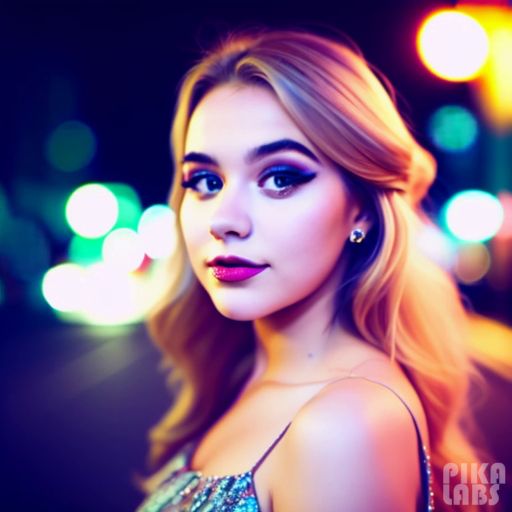}&
    \includegraphics[width=0.12\linewidth]{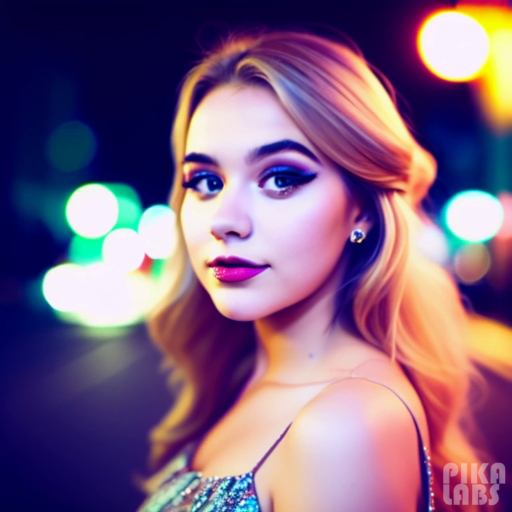}&
    \includegraphics[width=0.12\linewidth]{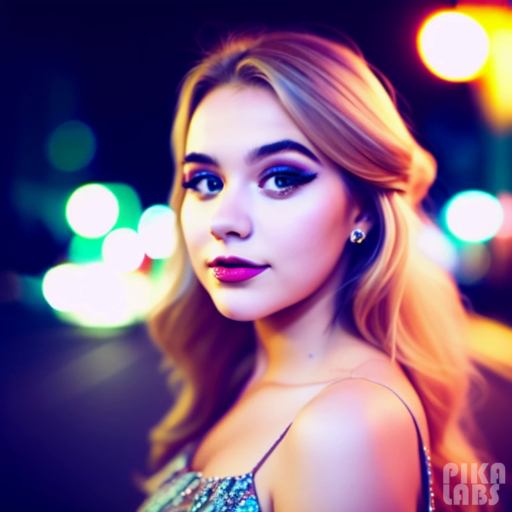}&
    \includegraphics[width=0.12\linewidth]{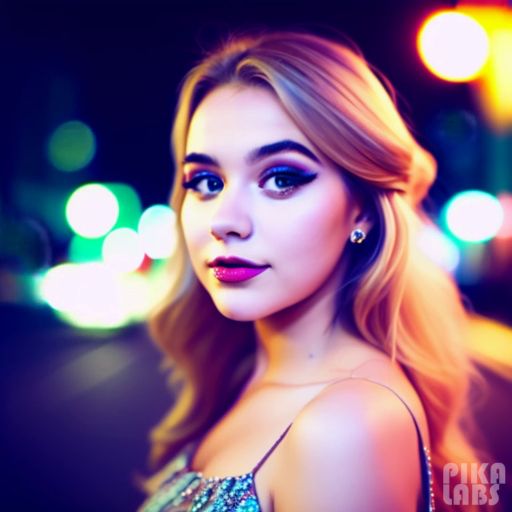}&
    \includegraphics[width=0.12\linewidth]{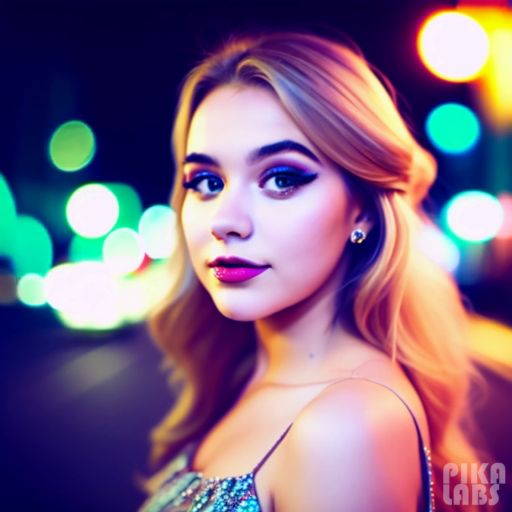}&
    \includegraphics[width=0.12\linewidth]{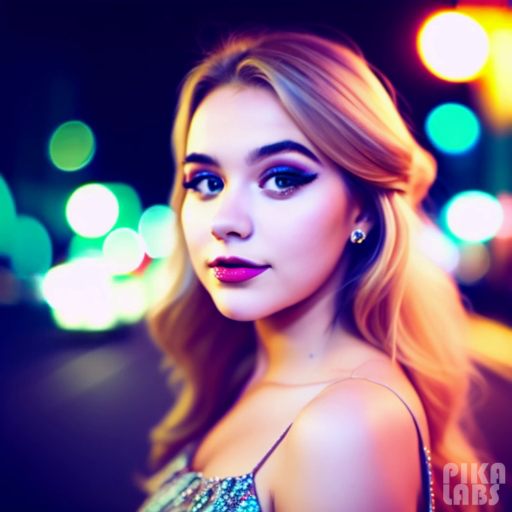}
    \\
    \rotatebox[origin=l]{90}{\scriptsize VideoCrafter1~\cite{chen2023videocrafter1}}&
    \includegraphics[width=0.12\linewidth]{gif/AZvsOthers/line2/videocrafter/007_0.jpg}&
    \includegraphics[width=0.12\linewidth]{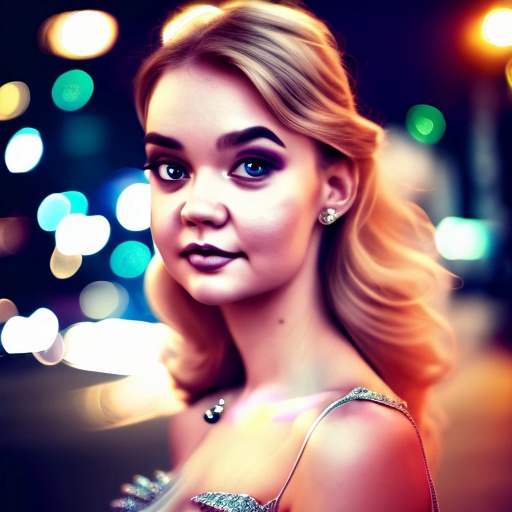}&
    \includegraphics[width=0.12\linewidth]{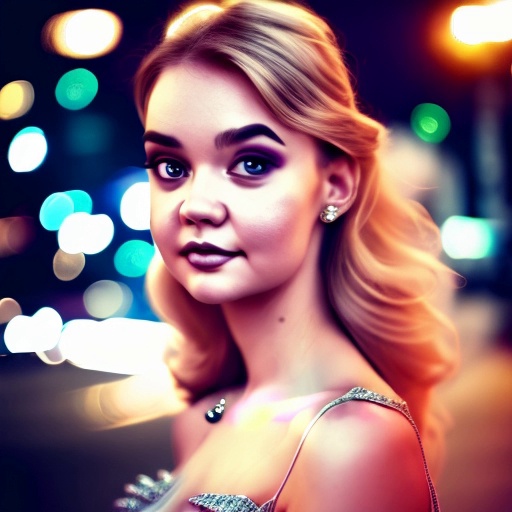}&
    \includegraphics[width=0.12\linewidth]{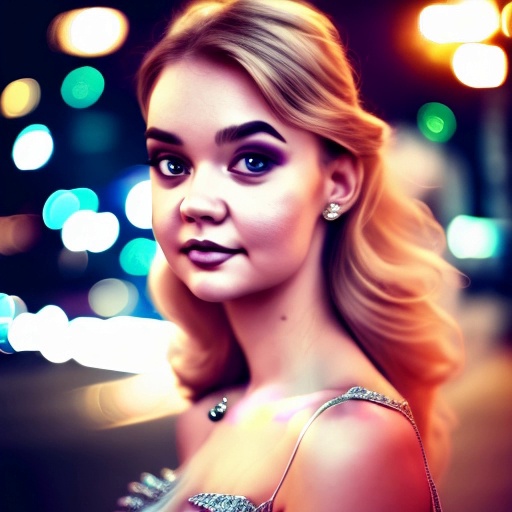}&
    \includegraphics[width=0.12\linewidth]{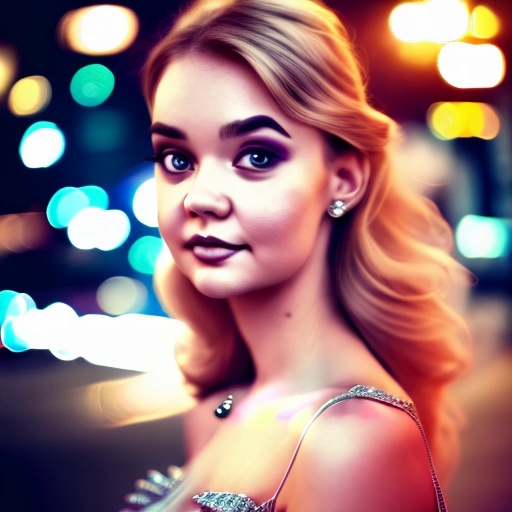}&
    \includegraphics[width=0.12\linewidth]{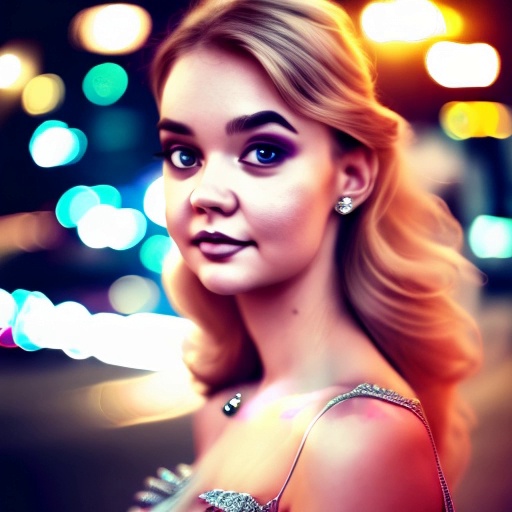}&
    \includegraphics[width=0.12\linewidth]{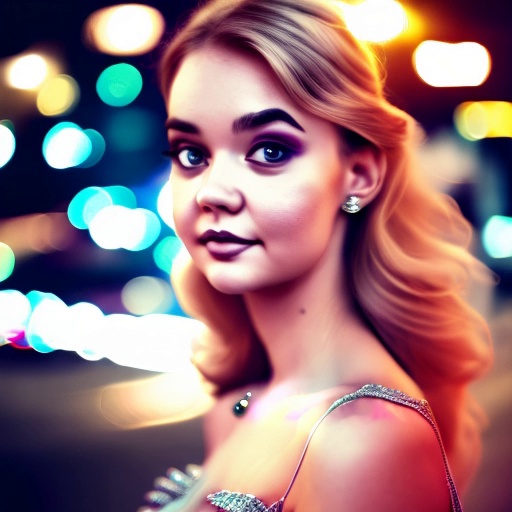}&
    \includegraphics[width=0.12\linewidth]{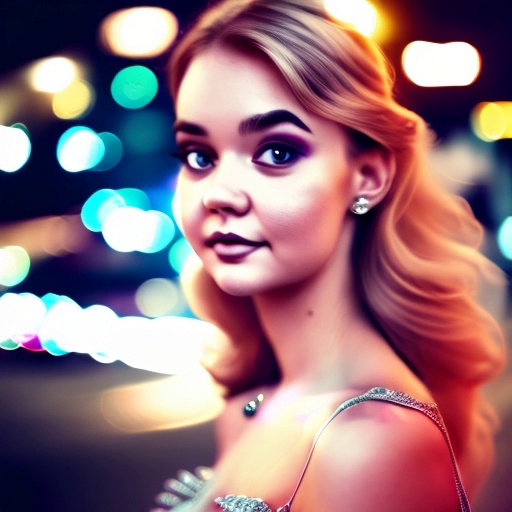}
    \\
    \rotatebox[origin=l]{90}{\small I2VGen-XL~\cite{i2vgenxl}}&
    \includegraphics[width=0.12\linewidth]{gif/AZvsOthers/line2/i2vgenxl/007_0.jpg}&
    \includegraphics[width=0.12\linewidth]{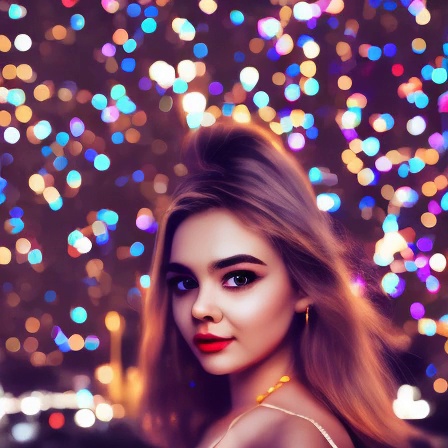}&
    \includegraphics[width=0.12\linewidth]{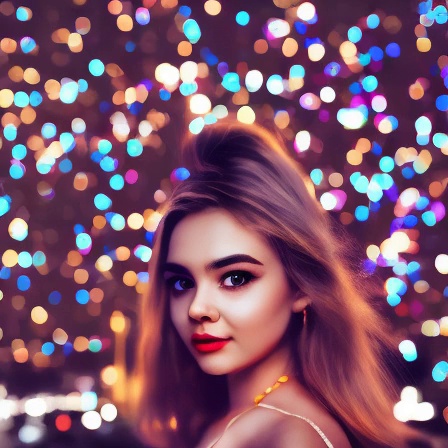}&
    \includegraphics[width=0.12\linewidth]{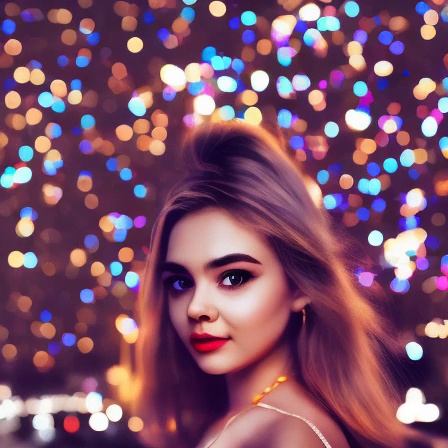}&
    \includegraphics[width=0.12\linewidth]{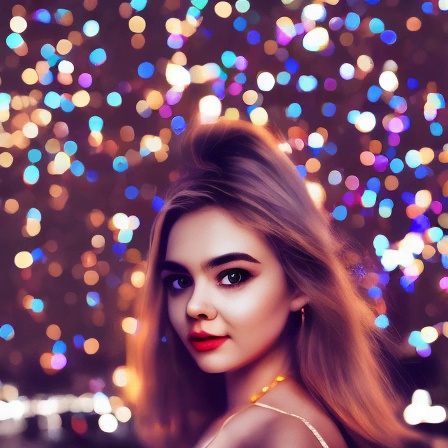}&
    \includegraphics[width=0.12\linewidth]{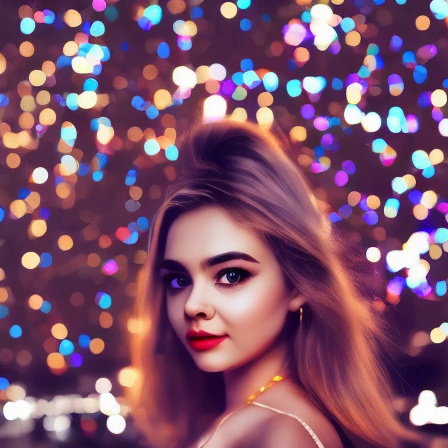}&
    \includegraphics[width=0.12\linewidth]{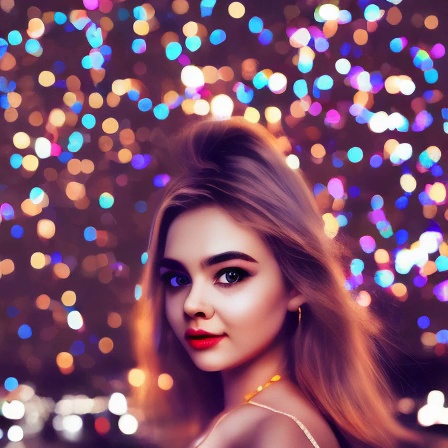}&
    \includegraphics[width=0.12\linewidth]{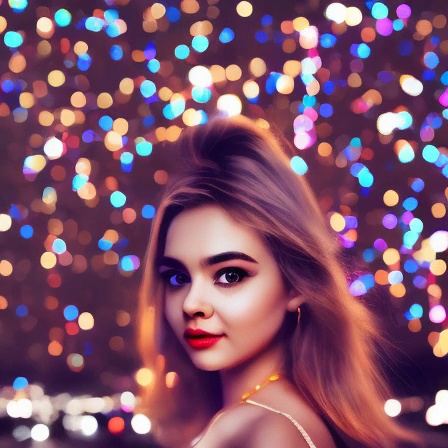}
    \\
    \rotatebox[origin=l]{90}{AnimateZero}&
    \includegraphics[width=0.12\linewidth]{gif/AZvsOthers/line2/az/015_frame_0.jpg}&
    \includegraphics[width=0.12\linewidth]{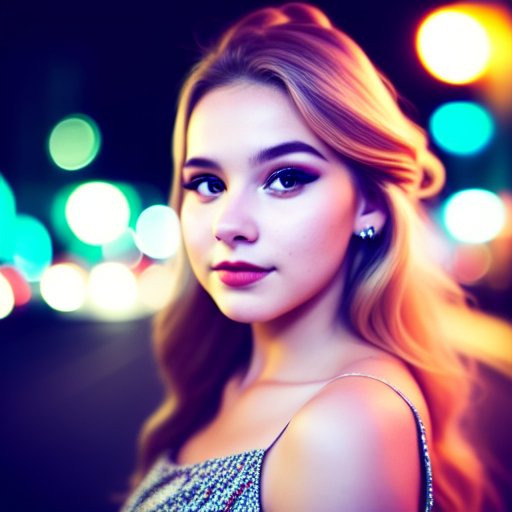}&
    \includegraphics[width=0.12\linewidth]{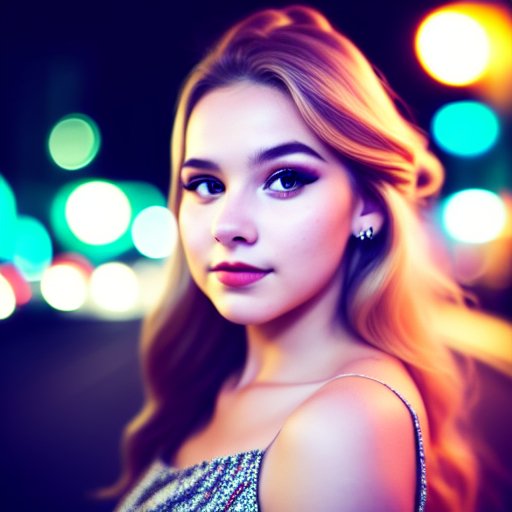}&
    \includegraphics[width=0.12\linewidth]{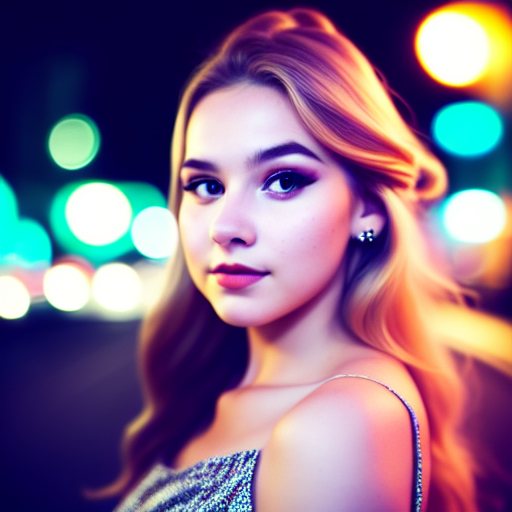}&
    \includegraphics[width=0.12\linewidth]{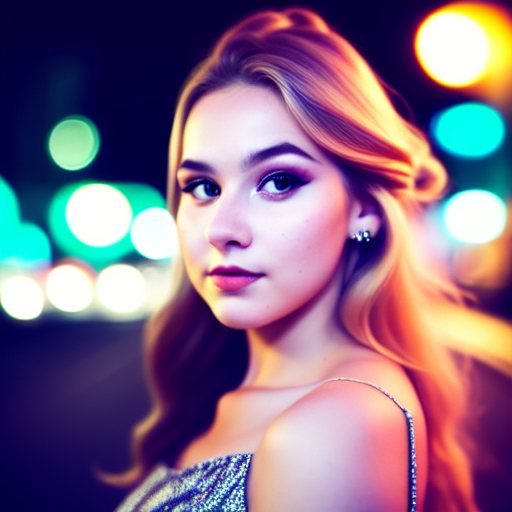}&
    \includegraphics[width=0.12\linewidth]{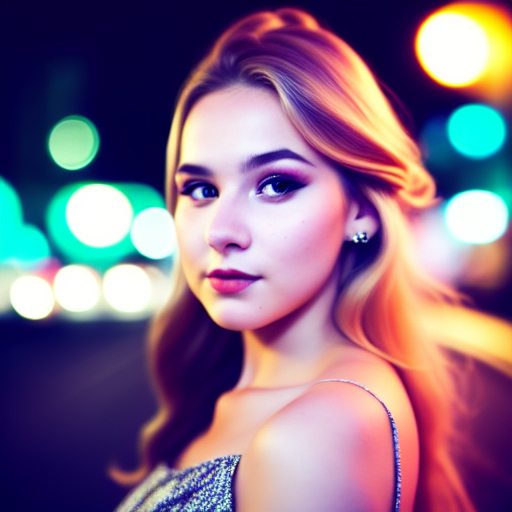}&
    \includegraphics[width=0.12\linewidth]{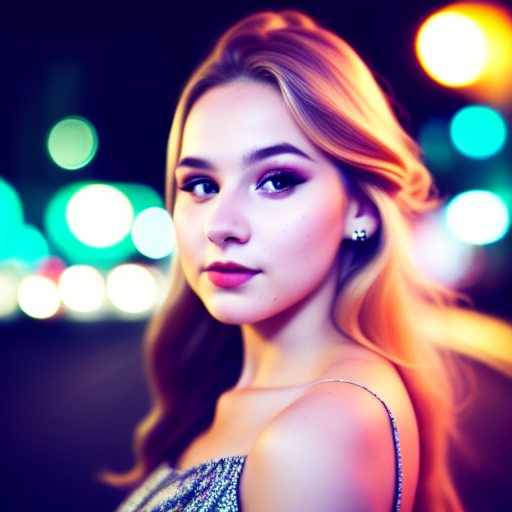}&
    \includegraphics[width=0.12\linewidth]{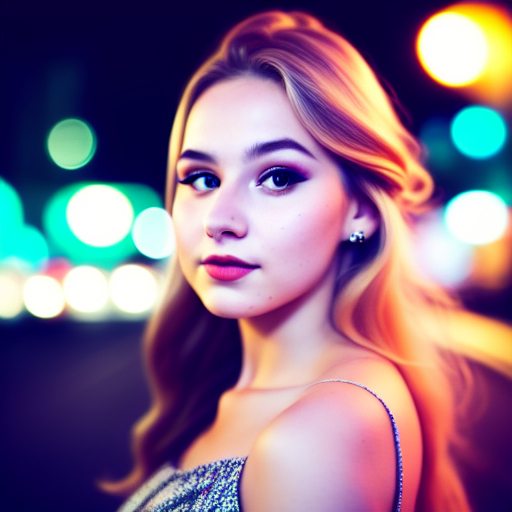}
    \\
    \multicolumn{9}{c}{\textit{``closeup face photo of 18 y.o swedish woman in dress, makeup, night city street, motion blur, ...''}}
    \end{tabular}
\vspace{-0.5em}
  \caption{Static frames sequences in Fig.~\ref{fig:ADvsOthers} (part 2).}
  \vspace{-1em}
\label{fig:static-fig5-2} 
\end{figure*}

%% file: append-fig-tex/static-fig7.tex
\begin{figure*}[ht]
  \centering
  \begin{tabular}{c@{\hspace{0.5em}}c@{\hspace{0.em}}c@{\hspace{0.em}}c@{\hspace{0.em}}c@{\hspace{0.0em}}c@{\hspace{0.em}}c@{\hspace{0.em}}c@{\hspace{0.em}}c} 
  \rotatebox[origin=l]{90}{\small AnimateDiff~\cite{guo2023animatediff}}&
    \includegraphics[width=0.12\linewidth]{gif/limitation/1/097_frame_0.jpg}&
    \includegraphics[width=0.12\linewidth]{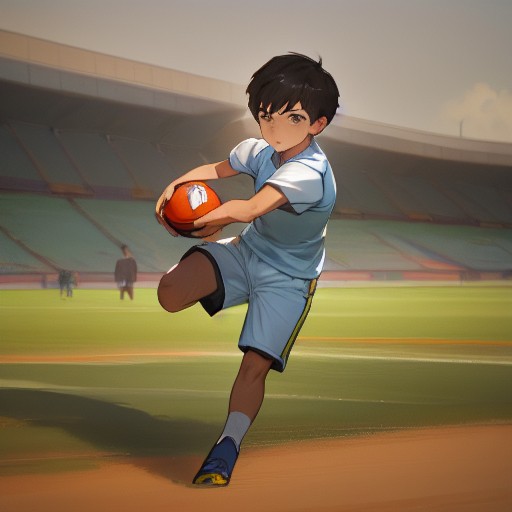}&
    \includegraphics[width=0.12\linewidth]{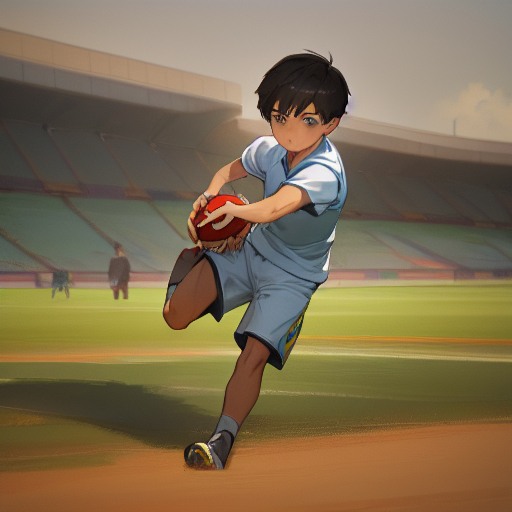}&
    \includegraphics[width=0.12\linewidth]{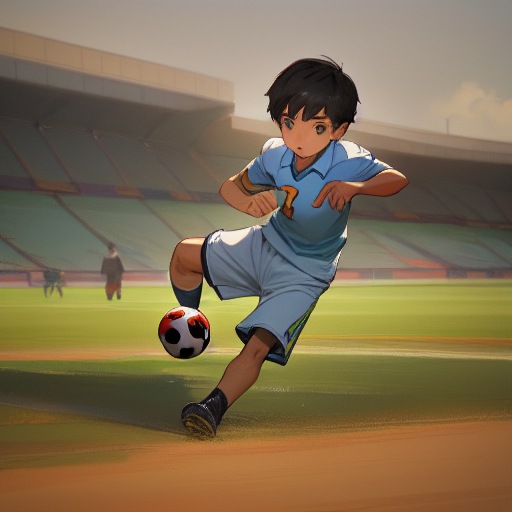}&
    \includegraphics[width=0.12\linewidth]{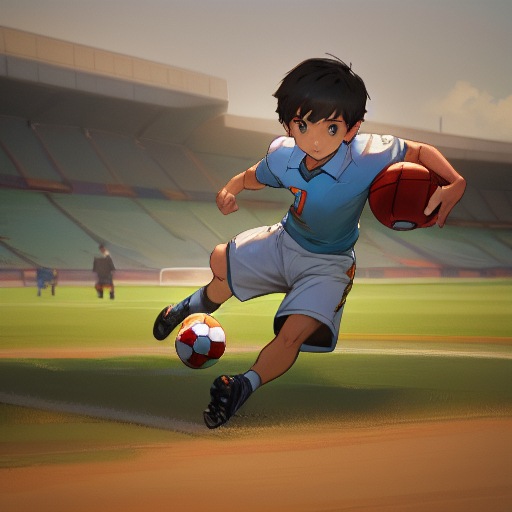}&
    \includegraphics[width=0.12\linewidth]{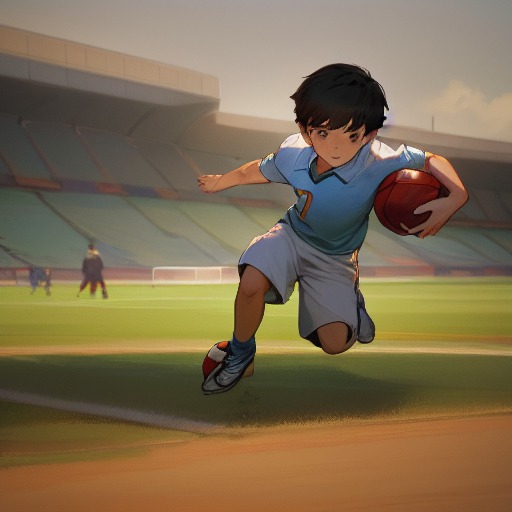}&
    \includegraphics[width=0.12\linewidth]{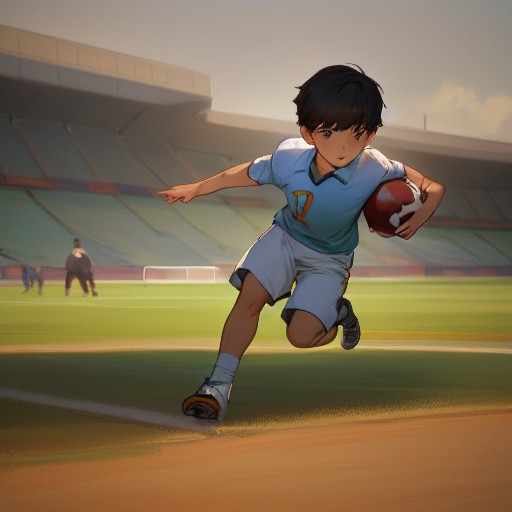}&
    \includegraphics[width=0.12\linewidth]{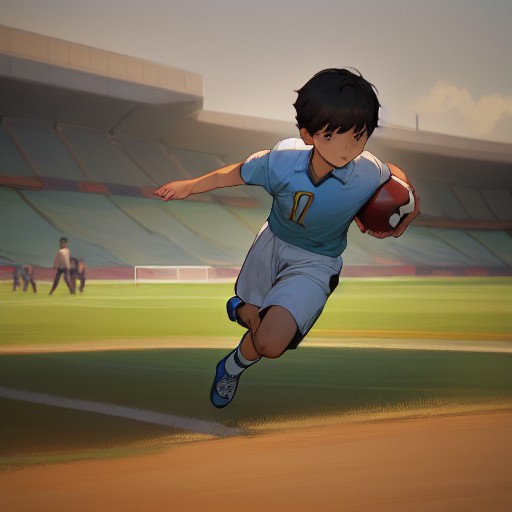}
    \\
    \rotatebox[origin=l]{90}{AnimateZero}&
    \includegraphics[width=0.12\linewidth]{gif/limitation/2/099_frame_0.jpg}&
    \includegraphics[width=0.12\linewidth]{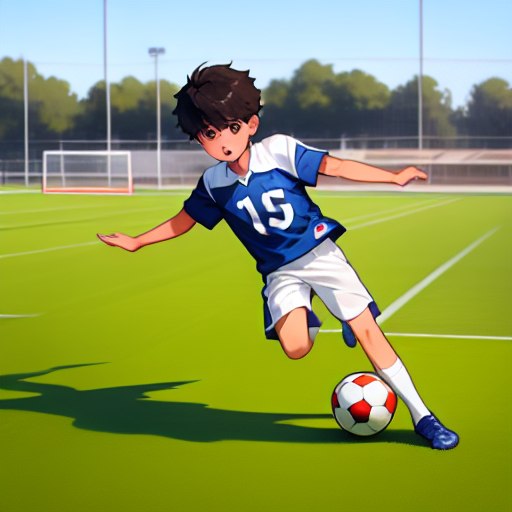}&
    \includegraphics[width=0.12\linewidth]{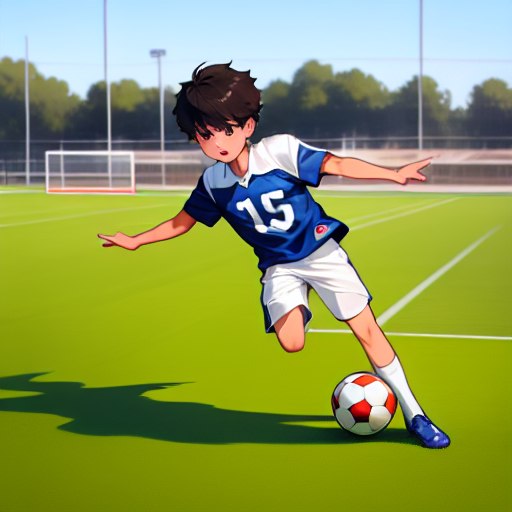}&
    \includegraphics[width=0.12\linewidth]{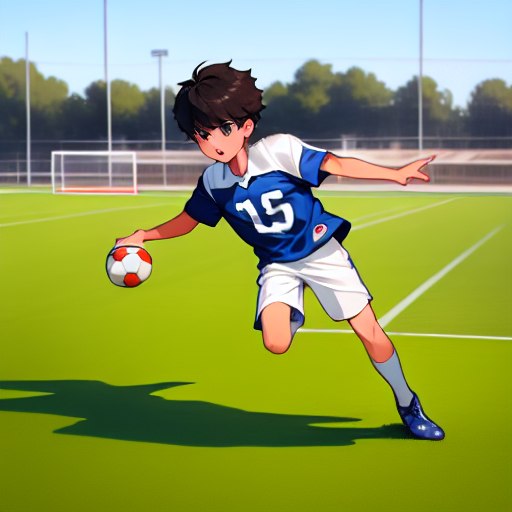}&
    \includegraphics[width=0.12\linewidth]{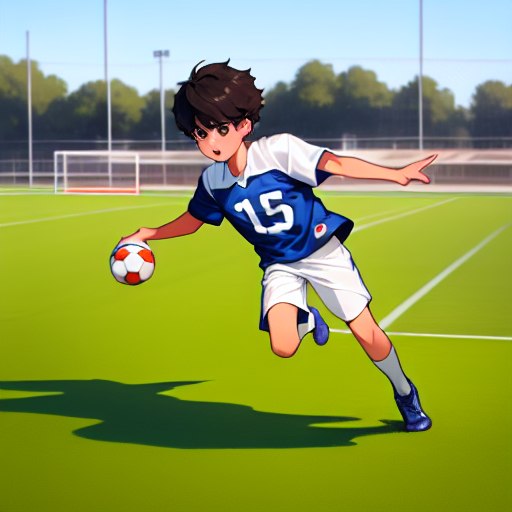}&
    \includegraphics[width=0.12\linewidth]{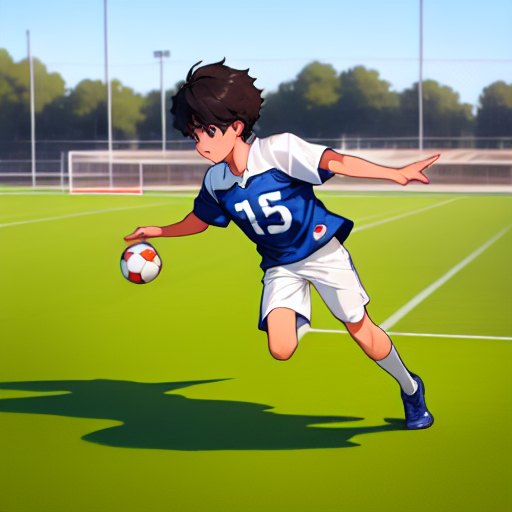}&
    \includegraphics[width=0.12\linewidth]{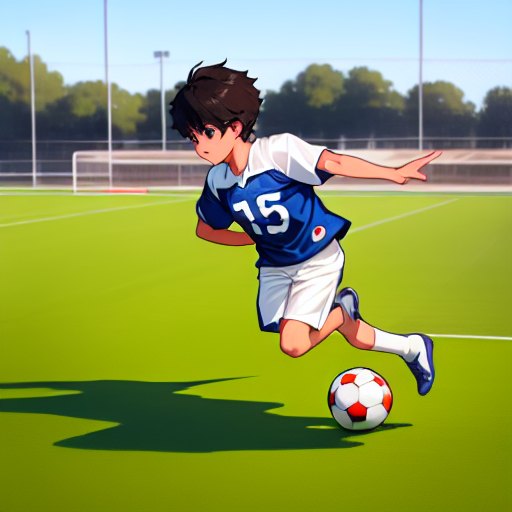}&
    \includegraphics[width=0.12\linewidth]{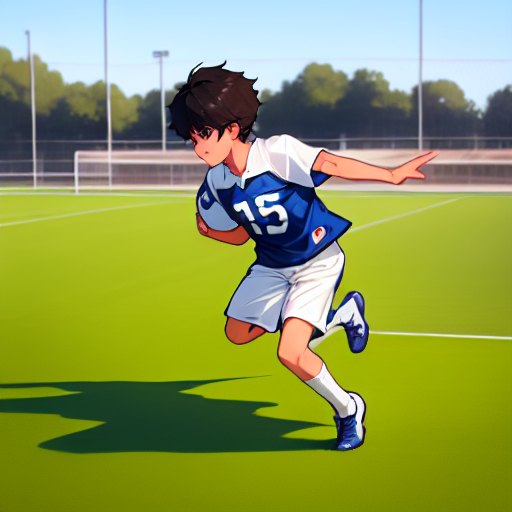}
    \\
    \multicolumn{9}{c}{\textit{``1boy, playing football, ...''}}\vspace{0.5em}
    \\
    \rotatebox[origin=l]{90}{\small AnimateDiff~\cite{guo2023animatediff}}&
    \includegraphics[width=0.12\linewidth]{gif/limitation/3/102_frame_0.jpg}&
    \includegraphics[width=0.12\linewidth]{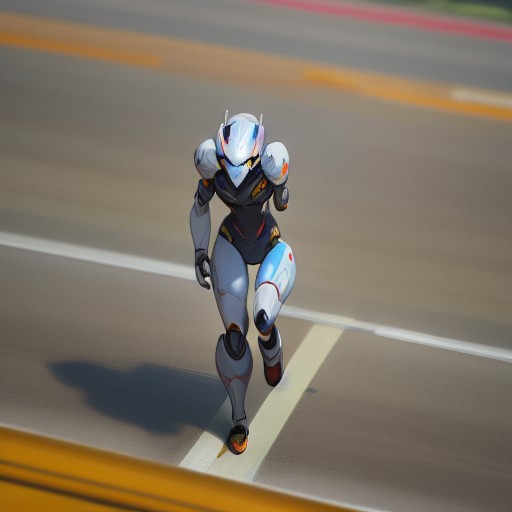}&
    \includegraphics[width=0.12\linewidth]{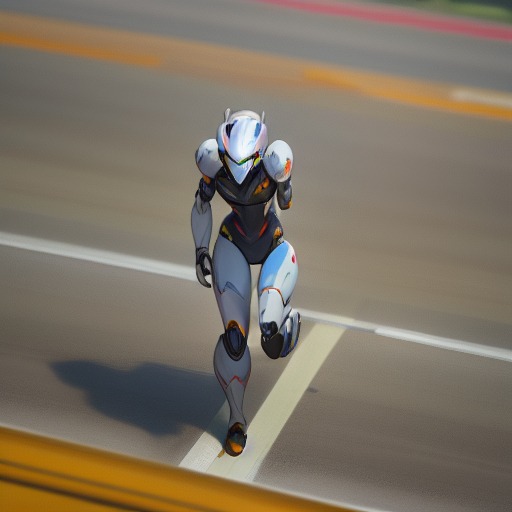}&
    \includegraphics[width=0.12\linewidth]{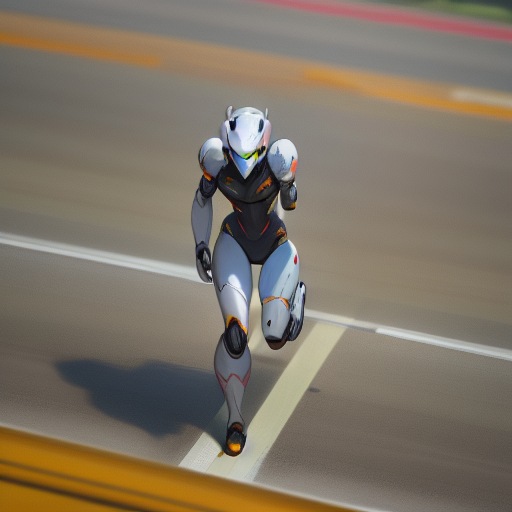}&
    \includegraphics[width=0.12\linewidth]{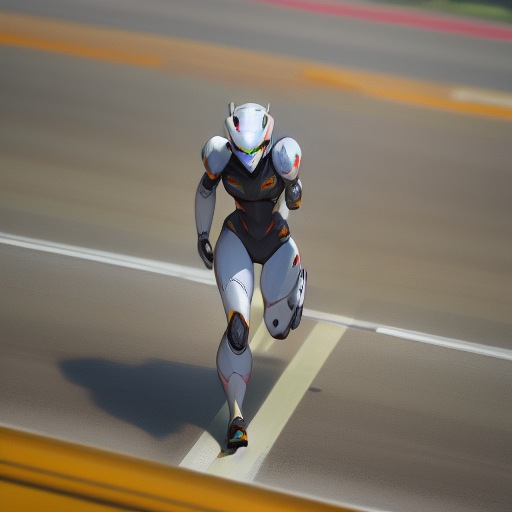}&
    \includegraphics[width=0.12\linewidth]{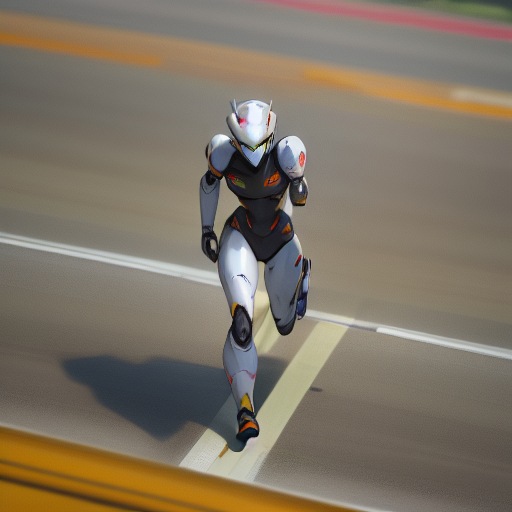}&
    \includegraphics[width=0.12\linewidth]{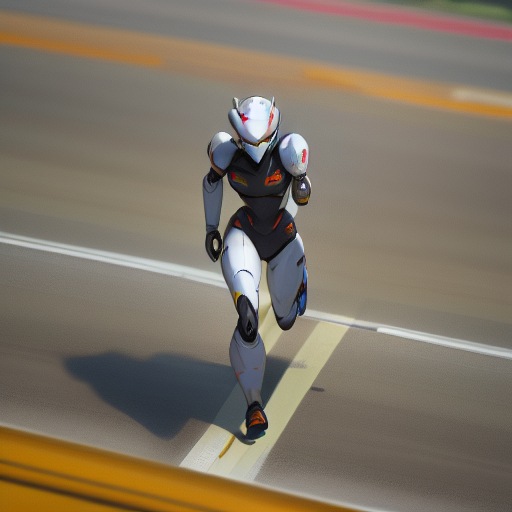}&
    \includegraphics[width=0.12\linewidth]{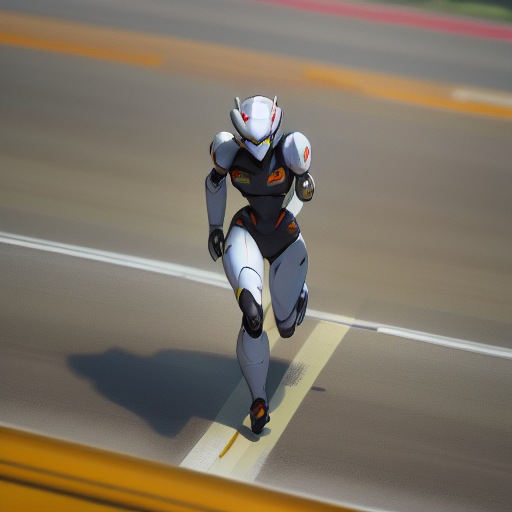}
    \\
    \rotatebox[origin=l]{90}{AnimateZero}&
    \includegraphics[width=0.12\linewidth]{gif/limitation/4/101_frame_0.jpg}&
    \includegraphics[width=0.12\linewidth]{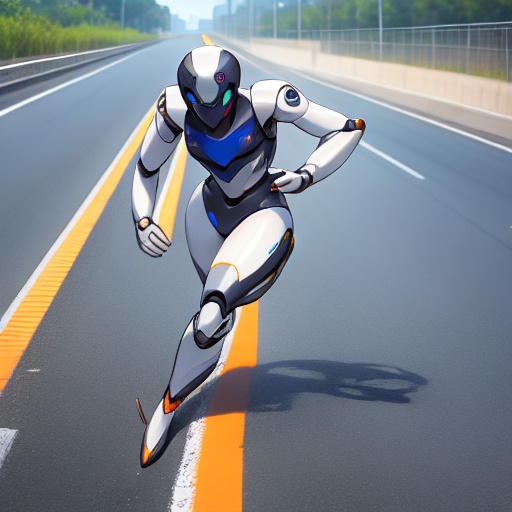}&
    \includegraphics[width=0.12\linewidth]{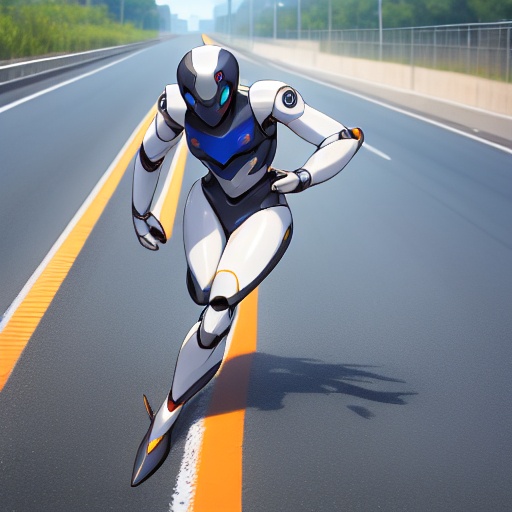}&
    \includegraphics[width=0.12\linewidth]{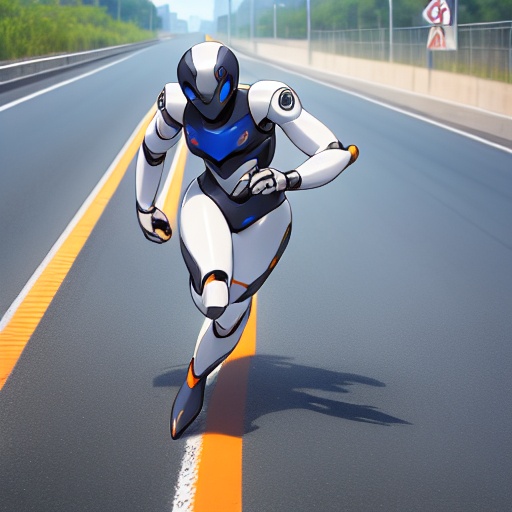}&
    \includegraphics[width=0.12\linewidth]{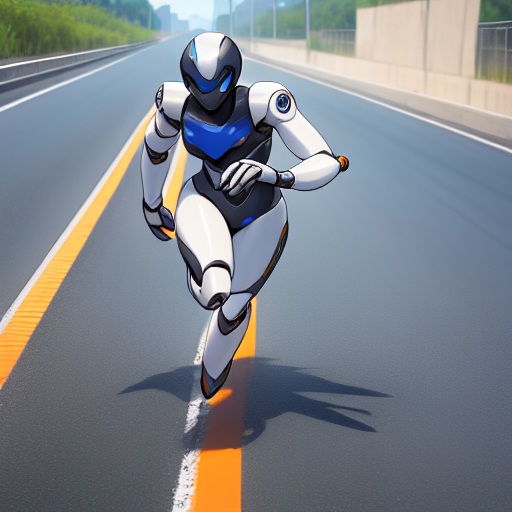}&
    \includegraphics[width=0.12\linewidth]{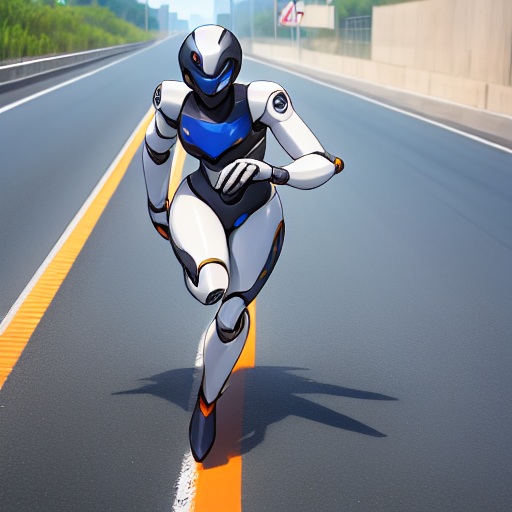}&
    \includegraphics[width=0.12\linewidth]{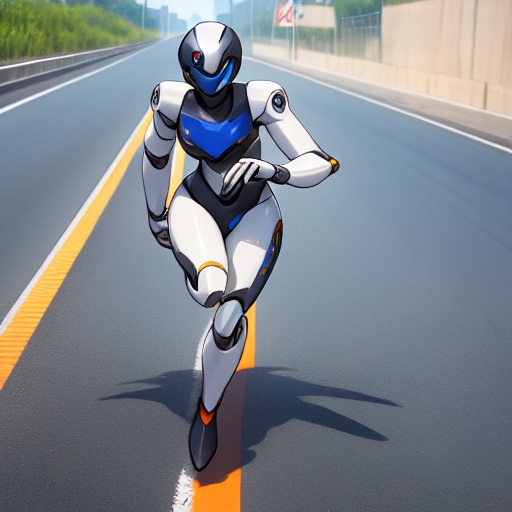}&
    \includegraphics[width=0.12\linewidth]{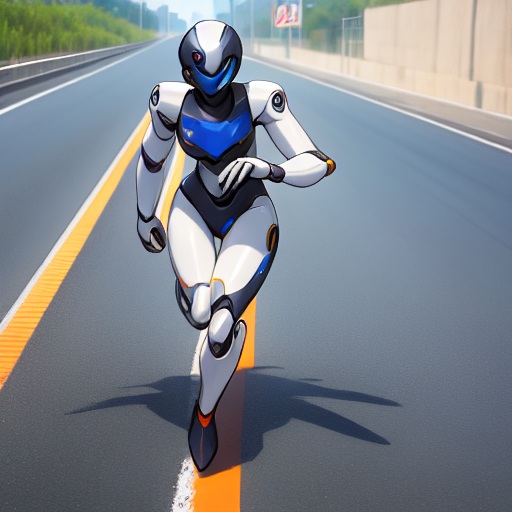}
    \\
    \multicolumn{9}{c}{\textit{``robot, running, ...''}}
    \end{tabular}
\vspace{-0.5em}
  \caption{Static frames sequences in Fig.~\ref{fig:limitation}.}
  \vspace{-1em}
\label{fig:static-fig7} 
\end{figure*}

%% file: append-fig-tex/static-figa.tex
\begin{figure*}[ht]
  \centering
  \begin{tabular}{c@{\hspace{0.5em}}c@{\hspace{0.em}}c@{\hspace{0.em}}c@{\hspace{0.em}}c@{\hspace{0.0em}}c@{\hspace{0.em}}c@{\hspace{0.em}}c@{\hspace{0.em}}c} 
  \rotatebox[origin=l]{90}{w/o time-travel}&
    \includegraphics[width=0.12\linewidth]{gif/time-travel/080/080_frame_0.jpg}&
    \includegraphics[width=0.12\linewidth]{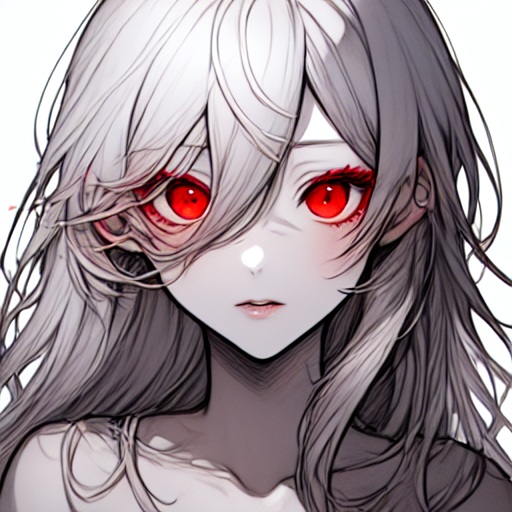}&
    \includegraphics[width=0.12\linewidth]{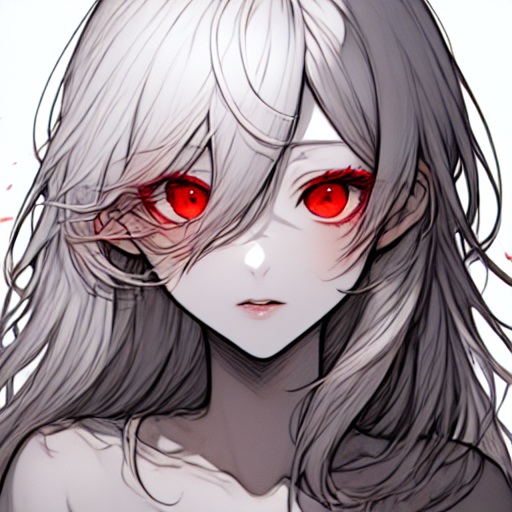}&
    \includegraphics[width=0.12\linewidth]{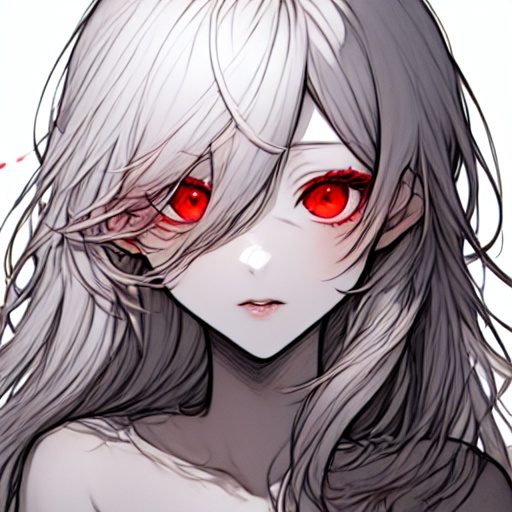}&
    \includegraphics[width=0.12\linewidth]{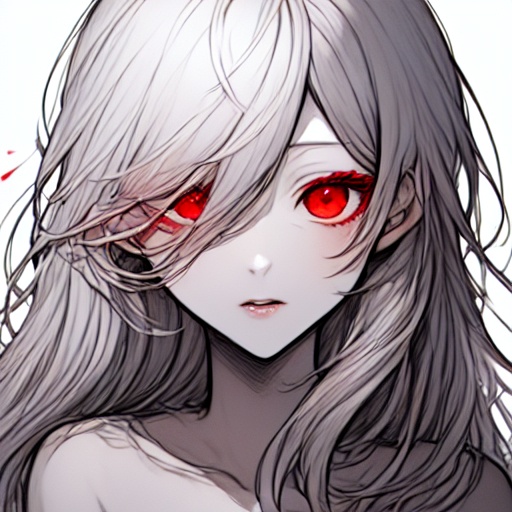}&
    \includegraphics[width=0.12\linewidth]{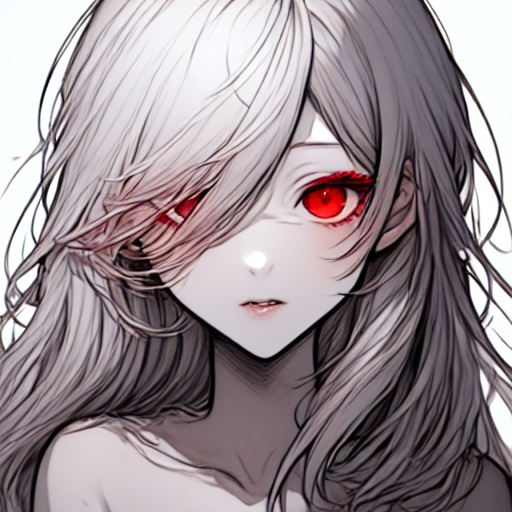}&
    \includegraphics[width=0.12\linewidth]{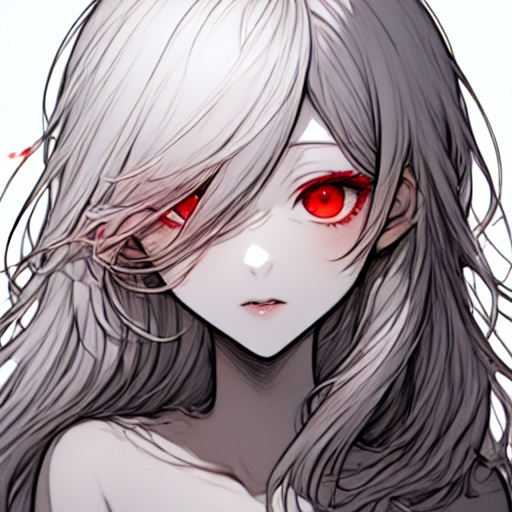}&
    \includegraphics[width=0.12\linewidth]{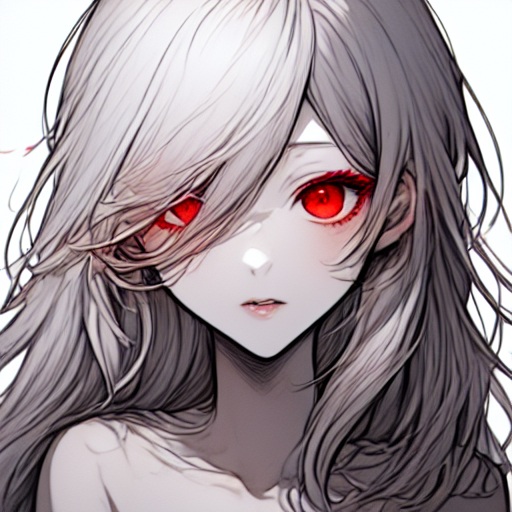}
    \\
    \rotatebox[origin=l]{90}{w/ time-travel}&
    \includegraphics[width=0.12\linewidth]{gif/time-travel/081/081_frame_0.jpg}&
    \includegraphics[width=0.12\linewidth]{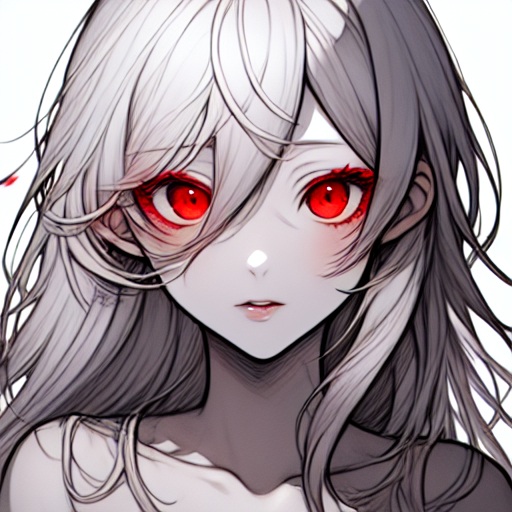}&
    \includegraphics[width=0.12\linewidth]{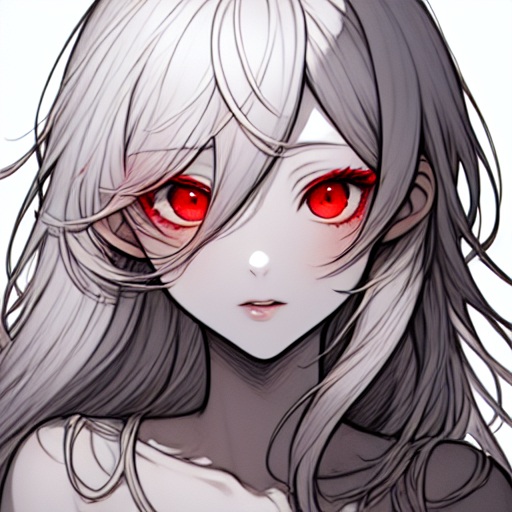}&
    \includegraphics[width=0.12\linewidth]{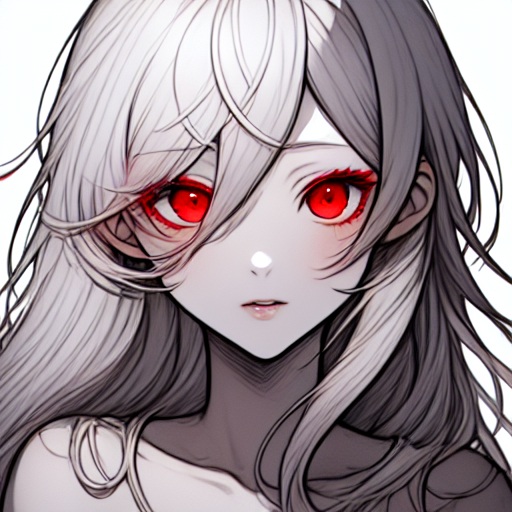}&
    \includegraphics[width=0.12\linewidth]{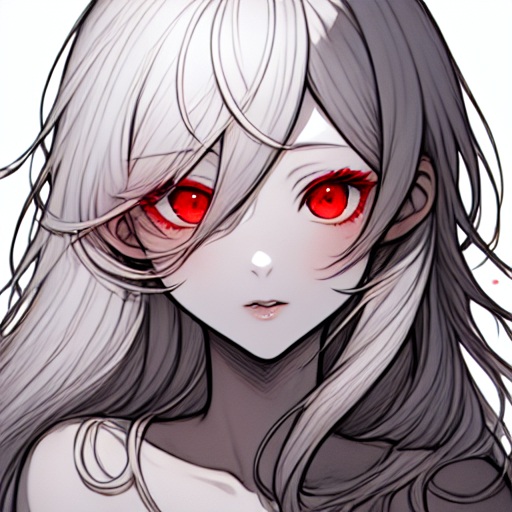}&
    \includegraphics[width=0.12\linewidth]{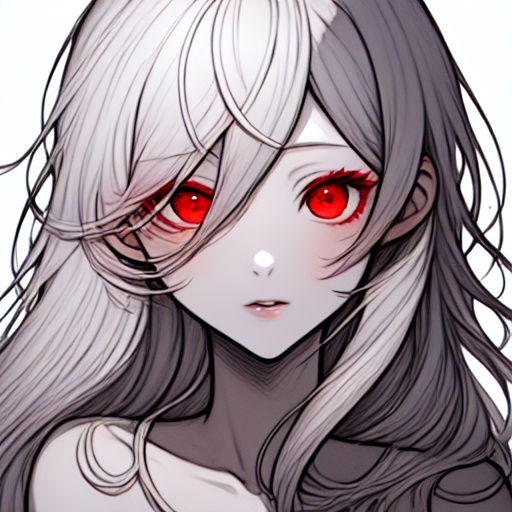}&
    \includegraphics[width=0.12\linewidth]{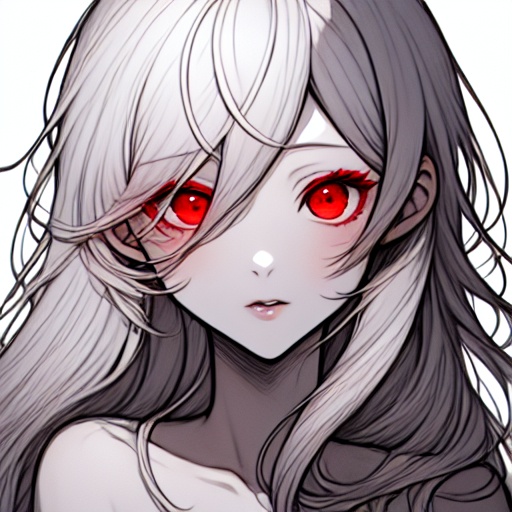}&
    \includegraphics[width=0.12\linewidth]{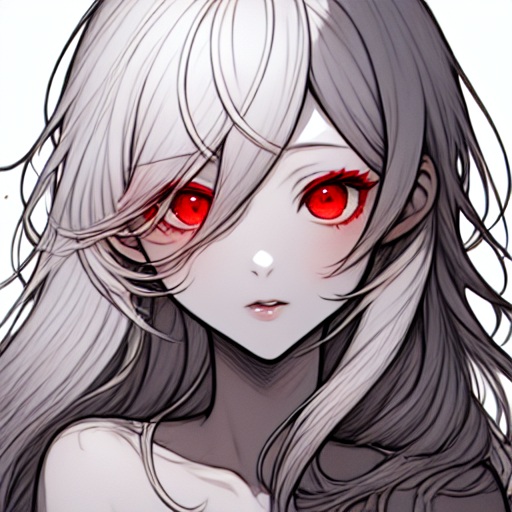}
    \\
    \multicolumn{9}{c}{\textit{``1girl, red eyes, silver hair, shiny skin, ...''}}\vspace{0.5em}
    \\
    \rotatebox[origin=l]{90}{w/o time-travel}&
    \includegraphics[width=0.12\linewidth]{gif/time-travel/084/084_frame_0.jpg}&
    \includegraphics[width=0.12\linewidth]{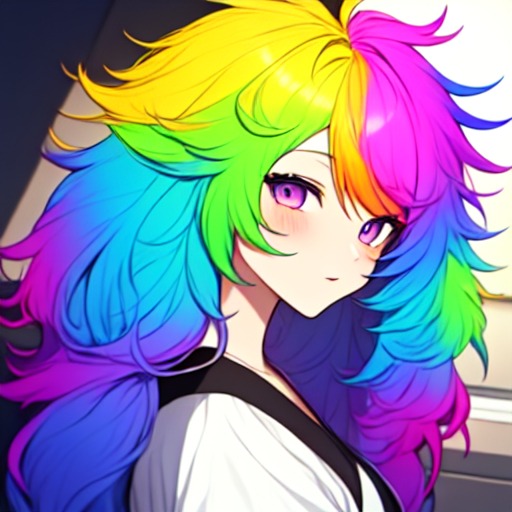}&
    \includegraphics[width=0.12\linewidth]{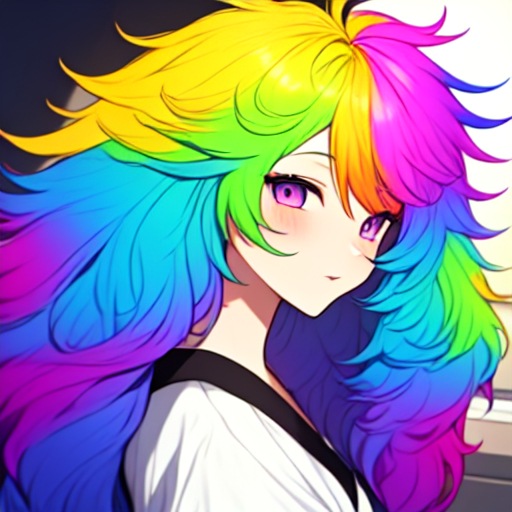}&
    \includegraphics[width=0.12\linewidth]{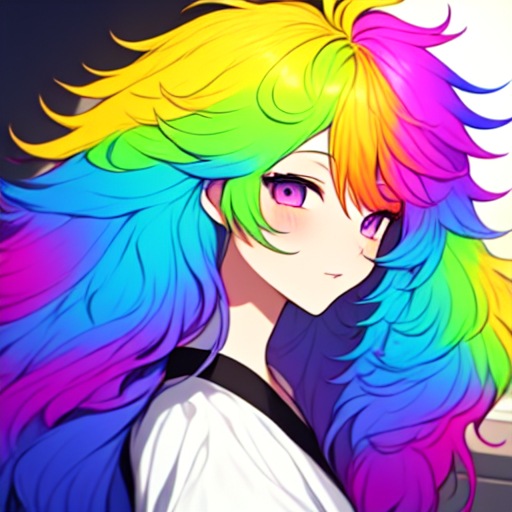}&
    \includegraphics[width=0.12\linewidth]{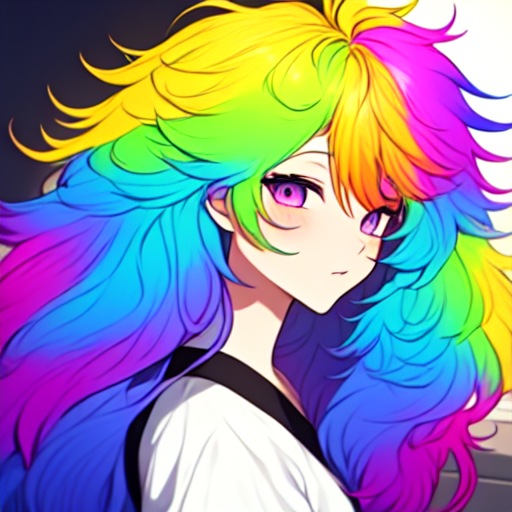}&
    \includegraphics[width=0.12\linewidth]{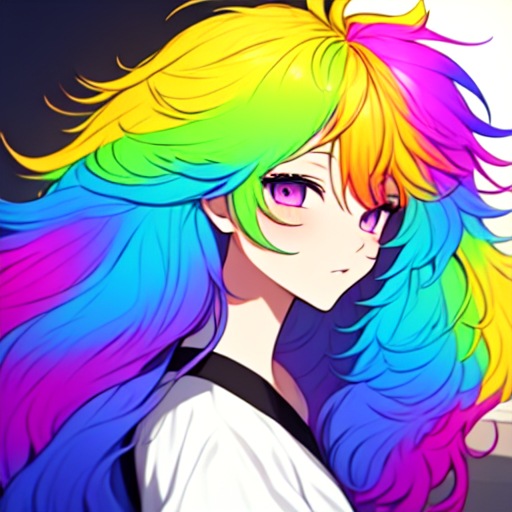}&
    \includegraphics[width=0.12\linewidth]{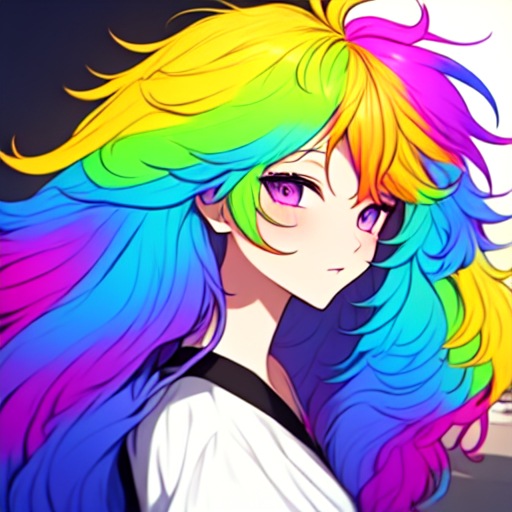}&
    \includegraphics[width=0.12\linewidth]{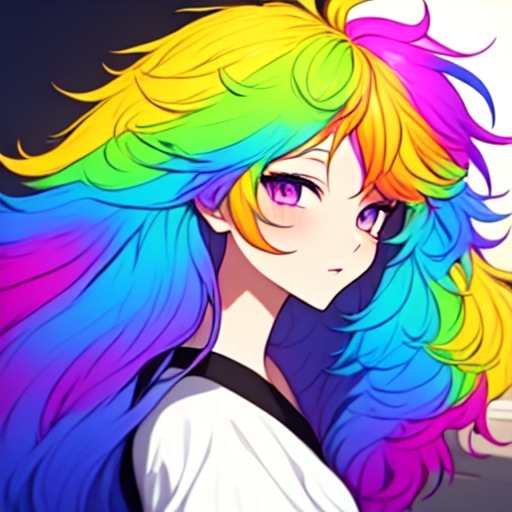}
    \\
    \rotatebox[origin=l]{90}{w/ time-travel}&
    \includegraphics[width=0.12\linewidth]{gif/time-travel/085/085_frame_0.jpg}&
    \includegraphics[width=0.12\linewidth]{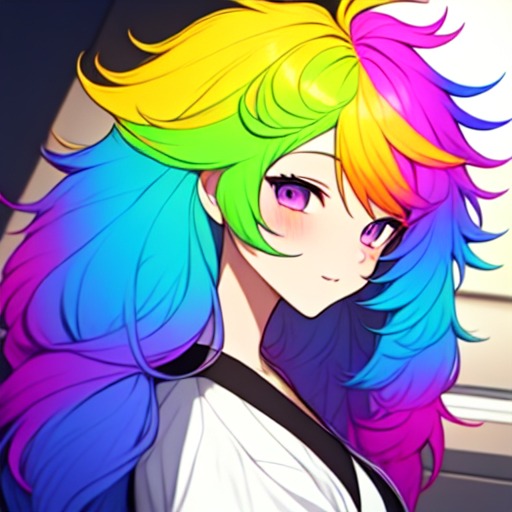}&
    \includegraphics[width=0.12\linewidth]{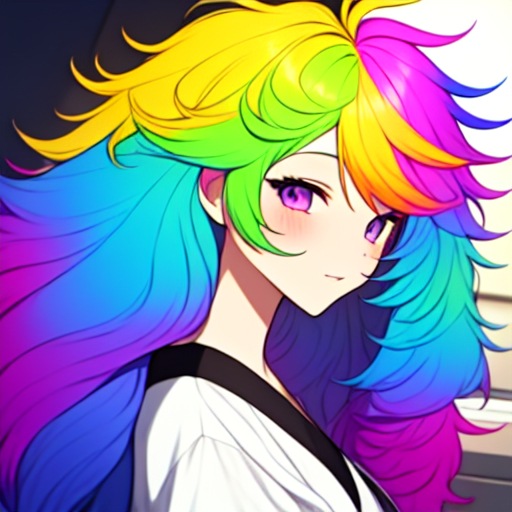}&
    \includegraphics[width=0.12\linewidth]{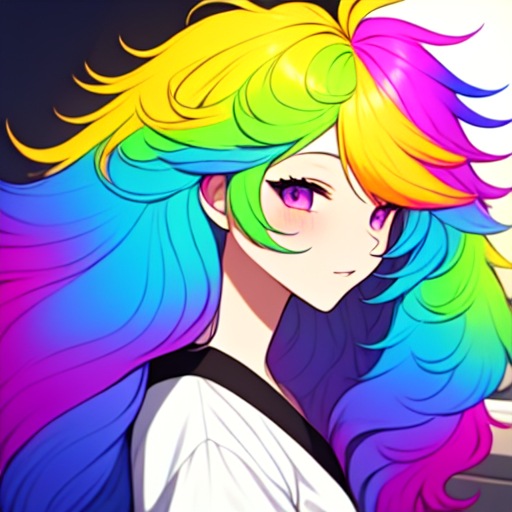}&
    \includegraphics[width=0.12\linewidth]{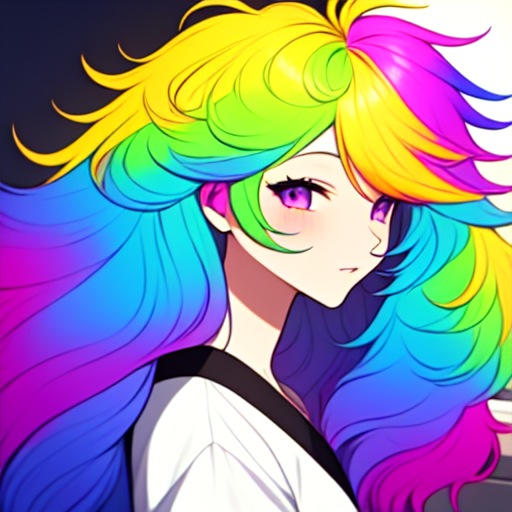}&
    \includegraphics[width=0.12\linewidth]{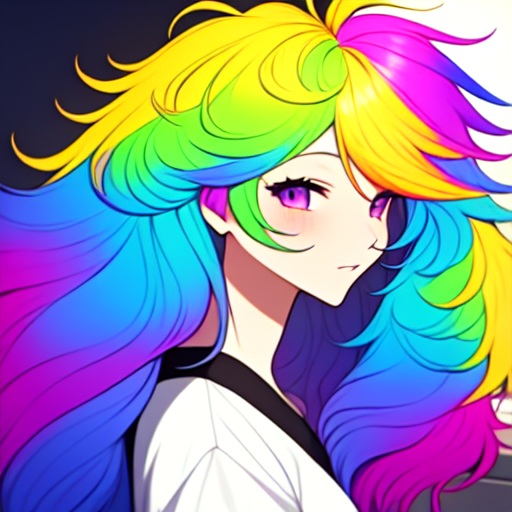}&
    \includegraphics[width=0.12\linewidth]{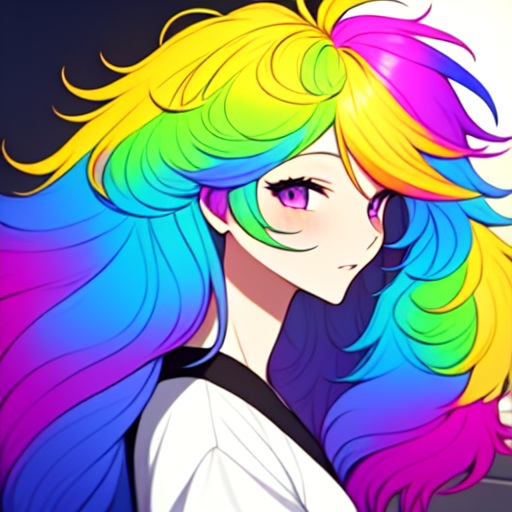}&
    \includegraphics[width=0.12\linewidth]{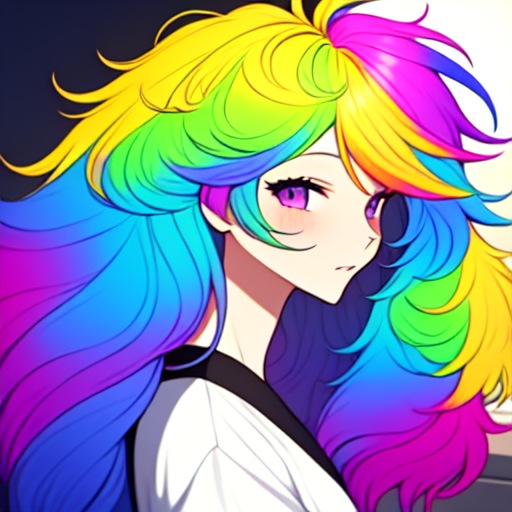}
    \\
    \multicolumn{9}{c}{\textit{``1girl with rainbow hair, really wild hair, ...''}}
    \end{tabular}
\vspace{-0.5em}
  \caption{Static frames sequences in Fig.~\ref{fig:time-travel}.}
  \vspace{-1em}
\label{fig:static-figa} 
\end{figure*}